%% file: pnas.tex
\documentclass[9pt,twocolumn,twoside]{pnas-new}

\usepackage{ifthen}
\newboolean{arxiv}
\setboolean{arxiv}{false}
\usepackage{pratik}
\templatetype{pnasresearcharticle}
\input{notation}

\graphicspath{{fig/}}

\makeatletter \renewcommand{\@dotsep}{10000} \makeatother
\usepackage[color=gray!30,textsize=tiny]{todonotes}

\newcommand{\change}[1]{{#1}}

\title{
The Training Process of Many Deep Networks Explores the Same Low-Dimensional Manifold
}

\author[a]{Jialin Mao}
\author[b]{Itay Griniasty}
\author[b]{Han Kheng Teoh}
\author[a]{Rahul Ramesh}
\author[a]{Rubing Yang}
\author[c]{Mark K. Transtrum}
\author[b]{James P. Sethna}
\author[a]{Pratik Chaudhari}

\affil[a]{University of Pennsylvania}
\affil[b]{Cornell University}
\affil[c]{Brigham Young University}

\leadauthor{Jialin Mao}
\significancestatement{
\input{significance}
}

\authorcontributions{All authors designed the research, analyzed the results, wrote and edited the manuscript. JM, IG, HKT, RR and RY conducted the numerical experiments.}
\authordeclaration{The authors declare no competing interests.}
\correspondingauthor{Corresponding Author: Pratik Chaudhari (\href{mailto:pratikac@seas.upenn.edu}{pratikac@seas.upenn.edu})}


\begin{abstract}
\input{abstract}
\end{abstract}

\doi{\url{https://doi.org/10.1073/pnas.2310002121}}

\begin{document}


\maketitle
\thispagestyle{firststyle}
\ifthenelse{\boolean{shortarticle}}{\ifthenelse{\boolean{singlecolumn}}{\abscontentformatted}{\abscontent}}{}

\input{intro}
\input{methods}
\input{results}

\input{discussion}
\input{ack}

\bibliography{bib/pratik,bib/main}

\newpage
\input{appendix}
\end{document}

%% file: notation.tex
\def \yvec {{\vec{y}}}
\def \yvecs {{\vec{y^*}}}

\def \d {\text{d}}
\def \dB {\text{d}_{\text{B}}}
\def \dG {\text{d}_{\text{G}}}
\def \dH {\text{d}_{\text{H}}}
\def \dsKL {\text{d}_{\text{sKL}}}
\def \dtraj {{\text{d}_{\text{traj}}}}

\def \tt {{\tilde{\t}}}

%% file: significance.tex


Training a deep neural network involves solving a high-dimensional, large-scale and non-convex optimization problem and should be prohibitively hard---but it is quite tractable in practice. To shed light upon this paradox, we develop new tools for the analysis and visualization of the prediction space of high-dimensional probabilistic models. Our experimental data shows that the training process explores a low-dimensional manifold in the prediction space. Networks with many different architectures, trained with different optimization procedures, and regularization techniques traverse the same manifold. This suggests that the optimization problem in deep learning is inherently low-dimensional.

%% file: abstract.tex
We develop information-geometric techniques to analyze the trajectories of the predictions of deep networks during training. By examining the underlying high-dimensional probabilistic models, we reveal that the training process explores an effectively low-dimensional manifold. Networks with a wide range of architectures, sizes, trained using different optimization methods, regularization techniques, data augmentation techniques, and weight initializations lie on the same manifold in the prediction space. We study the details of this manifold to find that networks with different architectures follow distinguishable trajectories but other factors have a minimal influence; larger networks train along a similar manifold as that of smaller networks, just faster; and networks initialized at very different parts of the prediction space converge to the solution along a similar manifold.

%% file: intro.tex


We show that training trajectories of multiple deep neural networks with different architectures, optimization algorithms, hyper-parameter settings, and regularization methods evolve on a remarkably low-dimensional manifold in the space of probability distributions. The key idea is to analyze the probabilistic model underlying a deep neural networks via their representation as probabilistic models as they are trained to classify images. Consider a dataset $\cbr{(x_n, y^*_n)}_{n=1}^N$ of $N$ samples, each of which consists of an input $x_n$ and its corresponding ground-truth label $y_n^* \in \cbr{1,\ldots,C}$ where $C$ is the number of classes. Let $\yvec = (y_1, \ldots, y_N) \in \cbr{1,\dots,C}^N$ denote any sequence of outputs. If samples in the dataset are independent and identically distributed, then the joint probability of the predictions can be modeled as
\beq{
    P_w(\yvec) = \prod_{n=1}^N p_w^n(y_n)
    \label{eq:def:Pw}
}
where $w$ are the parameters of the network and we have used the shorthand $p_w^n(y_n) \equiv p_w(y_n \mid x_n)$. \change{This is the joint likelihood of all the $N$ labels given the inputs and the parameters $w$; see~\cref{s:app:derivation} for details.} The probability distribution in~\cref{eq:def:Pw} is $N(C-1)$-dimensional object. Any network that makes predictions on the same set of samples---irrespective of its architecture, the optimization algorithm and regularization techniques that were used to train it---can be analyzed as a probabilistic model in this same $N(C-1)$-dimensional space; we will refer to this space as the ``prediction space''. We develop techniques to analyze such high-dimensional probabilistic models and embed these models into lower-dimensional spaces for visualization.

We first show, using experimental data (with $NC \sim 10^6-10^8$), that the training process explores an effectively low-dimensional manifold in the prediction space. The top three dimensions in our embedding explain 76\% of the ``stress'' (which is a quantity used to characterize how well the embedding preserves pairwise distances) between probability distributions of about 150,000 different models with many different architectures, sizes, optimization methods, regularization mechanisms, data augmentation techniques, and weight initializations. In spite of this huge diversity in configurations, the probabilistic models underlying these networks lie on the same manifold in the prediction space. This sheds new light upon a key open question in deep learning, namely how can training a deep network, with many millions of weights, on datasets with millions of samples, using a non-convex objective, be feasible.

We next study the details of the structure of this manifold. We find that networks with different architectures have distinguishable trajectories in the prediction space; in contrast, details of the optimization method and regularization technique do not change the trajectories in the prediction space much. We find that a larger network trains along a similar manifold as that of a smaller network with a similar architecture but it makes more progress for the same number of gradient updates. We find that models initialized at very different parts of the prediction space, e.g., by first fitting them to random labels, train along trajectories that merge quickly, approaching the true labels along the same manifold.



%% file: methods.tex

\section*{Methods\footnote{To aid the reader, \cref{s:app:notation} collects all the notation in one place.}}
\label{s:methods}

\paragraph{Measuring distances in the prediction space}
We first mark two special points in the prediction space that we will refer to frequently. The true probabilistic model of the data which corresponds to ground-truth labels is denoted by $P_* = \delta_{\yvecs}(\yvec)$ where $\yvecs$ are ground-truth labels and $\delta$ is the Kronecker delta function. We will call this the ``truth''. Similarly, we will mark a point called ``ignorance'': it is a probability distribution $P_0$  that predicts $p^n_0(c) = 1/C$ for all samples $n$ and classes $c$. Given two probabilistic models $P_u$ and $P_v$ with weights $u$ and $v$ respectively, the Bhattacharyya distance per sample between them is
\beq{
    \aed{
        \dB(P_u, P_v)  & =
        -N^{-1} \log \sum_\yvec \prod_{n=1}^N \sqrt{p_u^n(y_n)} \sqrt{p_v^n(y_n)}\\
        &\stackrel{(*)}{=} -N^{-1} \log \prod_{n=1}^N \sum_{c=1}^C \sqrt{p_u^n(c)} \sqrt{p_v^n(c)};\\
        &= -N^{-1} \sum_n \log \sum_c \sqrt{p_u^n(c)} \sqrt{p_v^n(c)};
    }
    \label{eq:dB}
}
here $(*)$ follows because samples are independent; \change{see~\cref{s:app:derivation} for more details}. In other words, the Bhattacharyya distance between two probabilistic models can be written as the average of the Bhattacharyya distances of their predictive distributions $p_u^n$ and $p_v^n$ on each input $x_n$. We can also use other distances to measure the discrepancy between $P_u$ and $P_v$, such as the symmetrized Kullback-Leibler divergence~\cite{teohVisualizingProbabilisticModels2020} (see~\cref{eq:isKL}), or the geodesic distance on the product space (see~\cref{eq:dG}). But many other distances (e.g., the Hellinger distance $\dH(P_w, P_*) = 2 \rbr{1- \prod_n \sum_c \sqrt{p^n_w(c)} \sqrt{p^n_*(c)}}$) saturate quickly as the number of dimensions of the probability distribution grows, obscuring the intrinsic low-dimensional structures we seek.  This is because two high-dimensional random vectors are orthogonal with high probability. When the number of samples $N$ is large, distances such as the Bhattacharyya distance are better behaved due to their logarithms.

\paragraph{Measuring distances between trajectories in the prediction space}

\begin{wrapfigure}{r}{0.5\linewidth}
\centering
\includegraphics[width=\linewidth]{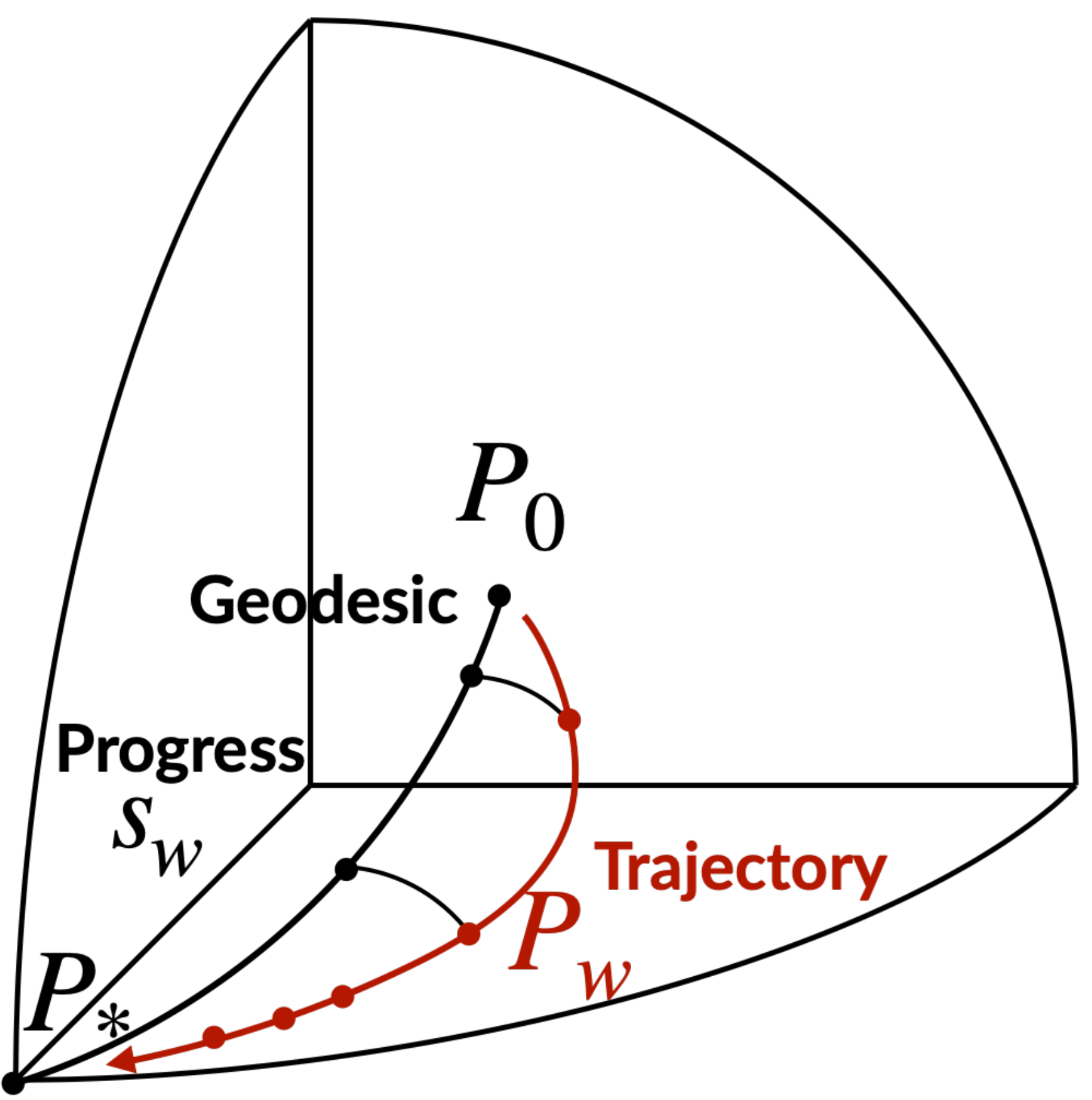}
\caption{A schematic of the procedure in~\cref{eq:tw} used to compute progress $s_w$ by projecting a model $P_w$ along a training trajectory onto the geodesic between ignorance $P_0$ and truth $P_*$.}
\label{fig:progress}
\end{wrapfigure}

Consider a trajectory $(u(k))_{k =0,\ldots,T}$ in the weight space that is initialized at $u(0)$ and records the weights after each update made by the optimization method during training. This corresponds to a trajectory $\tt_u = (P_{u(k)})_{k =0,\ldots,T}$ in the prediction space. We are interested in distances between trajectories in the prediction space. Different networks (depending upon the initialization, architecture, and the training procedure) train at different speeds and make different amounts of progress towards $P^*$ after each epoch. This makes it problematic to simply use a distance like $\sum_k \dB(P_{u(k)}, P_{v(k)})$ which sums up the distances between models at each instant $k$. To see why, observe that such a distance between $\tt_u$ and $\tt_v := (u(0), u(2), u(4), \ldots, u(2k), u(2k+2), \ldots)$ which progresses twice as fast as $\tt_u$, is non-zero even if the two trajectories are intrinsically the same.

To better compare trajectories, we need a notion of time that allows us to index any trajectory in prediction space. We shall measure progress along the trajectory by the projection onto the geodesic between ignorance and truth. Geodesics are locally length-minimizing curves in a metric space. Our trajectories evolve on the product manifold of the individual probability distributions in~\cref{eq:def:Pw}. Geodesics in this space using the Fisher Information Metric (FIM)~\cite{amariInformationGeometryIts2016} are a good candidate for constructing our index. The FIM is realized by a simple embedding.  For each $n$, consider a vector consisting of the square-root of the probabilities $(\sqrt{p^n_u(c)})_{c=1,\ldots,C}$ as a point on a $(C-1)$-dimensional sphere. Therefore the geodesic connecting two probability distributions $P_u$ and $P_v$ is the great circle on the sphere. A point along it with interpolation parameter $\a \in [0,1]$ denoted by $P^\a_{u,v}(\vec y) = \prod_ n p^{n,\a}_{u,v}(y_n)$ satisfies~\cite[Eq.~47]{ito2020stochastic}
\beq{
    \sqrt{p^{n,\a}_{u,v}} = \f{ \sin\rbr{(1 - \a)\dG^n}}{\sin \rbr{\dG^n}} \sqrt{p_u^n} + \f{\sin\rbr{\a \dG^n}}{\sin \rbr{\dG^n}} \sqrt{p_v^n};
    \label{eq:P_lambda}
}
where $\dG^n = \cos^{-1} \rbr{\sum_c \sqrt{p_u^n(c)} \sqrt{p_v^n(c)}}$ is one half of the great circle distance between $p^n_u(\cdot)$ and $p^n_v(\cdot)$. Any point $P_w$ along a trajectory can be reindexed using ``progress'' that is defined as
\beq{
    s_w = \argmin_{\a \in [0,1]} \dG(P_w, P^\a_{0,*}),
    \label{eq:tw}
}
where
\[
    \textstyle \dG(P_u, P_v) = N^{-1} \sum_n \cos^{-1} \sum_c \sqrt{p_u^n(c)} \sqrt{p_v^n(c)}
\]
is the geodesic distance on the product manifold. Note that progress $s_w \in [0,1]$ and it intuitively quantifies the motion along the trajectory by projecting onto the geodesic connecting ignorance and truth as in~\cref{fig:progress}.  We discuss the relationship between progress and error in~\cref{s:app:progress_vs_error}. To find a point's progress we solve \cref{eq:tw} using a bisection search~\cite{brent1971algorithm}.

We would now like to convert each trajectory $\tt_u = (P_{u(k)})_{k=0,\ldots,T}$ into a continuous curve $\t_u = (P_{u(s)})_{s \in [0,1]}$ and uniformly sample them for values of $s$ between $[0,1]$. To do this, we first calculate the progress $s_{u(k)}$ of all checkpoints along the trajectory $\tt_u$ using~\cref{eq:tw}. For any $s \in [s_{u(k)}, s_{u(k+1)}]$, we can now define $\a = (s-s_{u(k)})/(s_{u(k+1)}-s_{u(k)})$ and calculate (using~\cref{eq:P_lambda}) the geodesically-interpolated probability distribution $P^\a_{u(k),u(k+1)}$ that corresponds to this progress $s$ on the trajectory of interest $\tt_u$. Finally, we define the distance between trajectories $\t_u$ and $\t_v$ as
\beq{
    \dtraj(\t_u, \t_v) = \int_0^1 \dB(P_{u(s)}, P_{v(s)}) \dd{s},
    \label{eq:dtraj}
}
which compares points on the trajectories at equal progress.

\paragraph{Embedding predictions into a lower-dimensional space for visualization}
We use a technique called intensive principal component analysis (InPCA)~\cite{quinn2019visualizing,teohVisualizingProbabilisticModels2020} which is closely related to multi-dimensional scaling (MDS~\cite{cox2008multidimensional}) to project the predictions of the network into a lower-dimensional space to visually inspect their training trajectories. For $m$ probability distributions, consider a matrix $D \in \reals^{m \times m}$ with entries $D_{uv} =  \dB(P_u, P_v)$ and
\beq{
        W = -L D L/2
        \label{eq:w}
}
where $L_{uv} = \delta_{uv} - 1/m$, and \(W\) is the centered version of $D$. An eigen-decomposition of $W = U \L U^\top$ where the eigenvalues are sorted in descending order of their magnitudes $\abs{\L_{00}} \geq \abs{\L_{11}} \geq \ldots$ allows us to compute the  embedding of the $m$ probability distributions into an $m$-dimensional Minkowski space with metric signature \((p,m-p)\) derived from the \(p\) positive eigenvalues of \(W\) as $\reals^{p, m-p} \ni X = U \sqrt{\lvert\L\rvert}$\footnote{In special relativity, the axes corresponding to negative eigenvalues are often referred to as imaginary coordinates, and the metric signature is replaced by \( (x, i t)\cdot (x, i t)=x^2+i^2t^2=x^2-t^2\). However, this is not the inner product \(\norm{(x,i t)} = x^2+t^2\) over the complex numbers. We define a space where the distance between \(``(1,i)"\) and the origin vanishes and therefore its embedding  is \(\reals^{p,m-p}\) and not \(\complex^m\). }. In standard PCA, the embedding is always Euclidean since the eigenvalues of $W$ are guaranteed to be non-negative. However, InPCA can have both positive and negative eigenvalues. Coordinates corresponding to positive eigenvalues are analogous to ``space-like'' components in special relativity that have a positive-squared contribution to the distance between two points. Coordinates corresponding to negative eigenvalues are ``time-like'' components in that they have a negative contribution to the distance between two points. One can think of the coordinates with negative eigenvalues as being imaginary axes in the embedding.  Space-like and time-like coordinates can give rise to ``light-like'' directions along which the distance between two visually different points is zero.

The key property of InPCA that we exploit in this paper is that its embedding is isometric, i.e.,
\beq{
    \norm{X_u - X_v}^2 = \dB(P_u, P_v) \geq 0
    \label{eq:isometric}
}
for embeddings $X_u, X_v \in \reals^{p,m-p}$ of two probability distributions $P_u$ and $P_v$ and the norm in Minkowski space is
\[
    \norm{X_u - X_v}^2 = \sum_{k=1}^m \text{sign}(\L_{kk}) \abs{X_{uk}-X_{vk}}^2;
\]
see~\cref{s:app:isometry} for a proof. Like PCA, InPCA generates an optimal embedding of a geometrical object with a fixed number of points, preserving long distance structures. Such an isometric embedding is different from the one created by methods like t-SNE~\cite{van2008visualizing} or UMAP~\cite{mcinnes2018umap} which approximately preserve local pairwise distances but distort the global geometry. All the analysis in this paper is conducted using the full pairwise Bhattacharyya distance matrix $D$. In contrast with t-SNE or UMAP, the isometric embedding in InPCA ensures that the visualization is consistent with our conclusions (up to the fact that we only visualize the top few dimensions). For a $d < m$ dimensional InPCA embedding, the fraction of the centered pairwise distance matrix $W$ that is preserved is
\beq{
    \textstyle  1 - \sqrt{\f{\sum_{ij} \rbr{W_{ij} - \sum_{k=1}^d \sqrt{\L_{k k}} U_{i k} \sqrt{\L_{k k}} U_{k j}}^2}{\sum_{ij} W_{ij}^2}}
    = 1 - \sqrt{\f{\sum_{k=d+1}^m \L_{k k}^2}{\sum_{i} \L_{ii}^2}};
    \label{eq:explained_stress}
}
which is similar to the explained variance for standard PCA. Following the MDS literature, we call this quantity ``explained stress''. In this paper, we embed predictions of $m$ \textasciitilde\ $10^3$--$10^5$ models with $NC$ \textasciitilde\ $10^6$--$10^8$ using InPCA. This is very challenging computationally. Implementing InPCA---or even PCA---for such large matrices requires a large amount of memory. We reduced the severity of this issue using Numpy's memmap functionality. Note that calculating only the top few eigenvectors of~\cref{eq:w} by magnitude suffices for the purpose of visualization. \cref{s:app:iskl} discusses embeddings using other methods.

\paragraph{Adding new networks into an existing embedding}
Given the embedding of predictions of $m$ networks we can project the prediction of a new network into the same space. Observe that we can rewrite~\cref{eq:w} to be
\beq{
    \aed{
    \textstyle & W_{uv} = -\f{\dB(P_u, P_v)}{2} +\\
    &\textstyle \f{1}{2 m} \sum_{u'}\rbr{ \dB(P_u, P_{u'}) + \dB(P_v, P_{u'}) - \f{1}{m} \sum_{v'} \dB(P_{u'}, P_{v'})};
    }
    \label{eq:w_expanded}
}
where $u',v' \in \{1,\dots,m\}$. The embedding of a new probability distribution $P_w$ into this space is
\(
    X_w = \sum_{u=1}^{n} W_{w,u} U_{u} \lvert\L_{uu}\rvert^{-1/2};
\)
where $U_u$ denotes the $u^{\text{th}}$ column of $U$. This is equivalent to a triangulation of the position of the added points, such that distances and the overall geometry are preserved. We discuss a generalization of this approach in~\cref{s:app:weighted_embedding}. Although we do not do so in this paper, this procedure can also be used to embed a large set of points by computing the eigen-decomposition for only a subset, e.g., as done in~\cite{de2004sparse}. 

\paragraph{Computing averages in the prediction space}
For our analysis, we will need to compute averages of the predictions of probabilistic models, e.g., of the same architecture but trained from different initializations. Depending upon what distance we use in the prediction space, there can be different ways to compute such an average. The most natural candidate is the Bhattacharyya centroid of a set of $m$ probability distributions $\{P_i\}_{i=1}^m$ given by $\argmin_{P_w} m^{-1} \sum_i \dB(P_i, P_w)$~\cite{nielsen2011burbea}. In this paper, we will need to compute such averages thousands of times. For computational convenience, we will instead use the arithmetic mean of the probabilities $m^{-1} \sum_u p^n_u(c)$ for all $n,c$ as our average, which we have found to produce similar results in preliminary experiments (see~\cref{fig:allcnn_mean},
which discusses the effect of different kinds of averaging). We have found that the harmonic mean of an ensemble of probabilistic models performs slightly better on the test data in comparison to their arithmetic mean, which is commonly used in machine learning.

%% file: results.tex

\section*{Results}
\label{s:results}

\begin{table}
\centering
\caption{
Median (and 25--75 percentile on the second row) train and test error (\%) of different architectures (with number of parameters in the brackets) used in our analysis, averaged over different optimization methods, regularization techniques and weight initializations.\protect\footnotemark}
\label{tab:test_error}
\renewcommand{\arraystretch}{1.25}
\resizebox{\linewidth}{!}{
\LARGE
\begin{tabular}{lrrrrrr}
\toprule
& \multicolumn{6}{c}{\textbf{CIFAR-10}}\\
& Fully-Connected & AllCNN & Small ResNet & Large ResNet & ConvMixer & ViT\\
& (3.8M) & (0.4M) & (0.3M) & (43.9M) & (0.6M) & (9.5M)\\
\midrule
Train Error & 1.5 & 0.1 & 0.6 & 0.0 & 0.0 & 0.3\\
& (0.0, 4.4) & (0.0, 0.5) & (0.0, 2.3) & (0.0, 0.0)  & (0.0, 0.0) & (0.0, 18.6)\\
Test Error & 39.7 & 15.4 & 17.6 & 9.6 & 11.7 & 32.7\\
& (38.1, 41.9) & (11.7, 20.3) & (12.5, 21.5) & (6.5, 11.2) & (9.9, 16.8) & (21.7, 36.2)\\
\bottomrule\\
\end{tabular}
}
\vspace*{1em}
\resizebox{0.6\linewidth}{!}{
\Large
\begin{tabular}{lrrr}
\toprule
& \multicolumn{3}{c}{\textbf{ImageNet}}\\
& ResNet-18 & ResNet-50 & ViT-S\\
& (11.6M) & (25.6M) & (22M)\\
\midrule
Train Error & 22.7 & 15.8 & 16.6 \\
& (22.5, 22.7) & (15.8, 15.8) & (15.1, 16.9) \\
Test Error & 31.9 & 25.2 & 41.5 \\
& (31.8, 31.9) & (25.1, 25.3) & (41.3, 42.2) \\
\bottomrule
\end{tabular}
}
\vspace*{-2em}
\end{table}
\footnotetext{For CIFAR-10, some configurations had models that did not get to zero train error, and in very few cases, models had 90\% train error. For ImageNet, all networks were trained with standard data augmentation techniques and they do not reach zero training error.}

\subsection*{Experimental Data
\footnote{Data, pre-processing scripts, and code are available at \href{https://github.com/grasp-lyrl/low-dimensional-deepnets}{https://github.com/grasp-lyrl/low-dimensional-deepnets}}
}
We trained 2,296 different configurations on the CIFAR-10 dataset~\cite{krizhevsky2009learning} corresponding to networks \footnote{In the sequel, ``network'' denotes a particular configuration with a specific architecture, optimization method, regularization technique, hyper-parameter choice, data-augmentation, and weight initialization. ``Model'' denotes a probability distribution along the training trajectory of such a network.} with different (a) network architectures (fully-connected, convolutional: AllCNN~\cite{springenbergStrivingSimplicityAll2015}, residual: Wide ResNet~\cite{zagoruyko2016wide}, and ConvMixer~\cite{trockman2022patches}, self-attention-based: ViT~\cite{dosovitskiy2020image}), (b) network sizes (a small residual network and a large residual network), (c) optimization methods (SGD, SGD with Nesterov's acceleration and Adam~\cite{kingma2014adam}), (d) hyper-parameters (learning rate and batch-size), (e) regularization mechanisms (with and without weight-decay~\cite{ioffe2015batch}), (f) data augmentation (mean-standard deviation-based normalization, and another one where we add horizontal flips and random crops) and (g) random initializations of weights (using 10 different random seeds). We recorded the training trajectories at about 70 different points during training (more frequently at the beginning of training when the models train quickly). This gave us 151,407 different models, after removing some models that did not train correctly due to numerical overflows/underflows during gradient updates.

We also performed a smaller scale experiment on ImageNet using (a) three different architectures (a small residual network: ResNet-18~\cite{he2016deep}, a larger residual network ResNet-50, and a self-attention-based network: ViT), (b) different optimization algorithms (SGD with Nesterov's acceleration for the residual networks, and a variant of Adam for ViT~\cite{heo2021adamp}), (c) 5 random weight initializations for the residual networks and 3 for the ViT. We recorded each training trajectory at 61 different points to obtain a total of 792 different models for ImageNet.

\cref{tab:test_error} summarizes the train and test errors of models used in our analysis. \cref{s:app:details} gives more details of the training procedure. About 60,000 GPU hours were used to obtain and analyze the data in this paper.

\subsection*{The training process explores an effectively low-dimensional manifold in the prediction space}
\begin{figure}
\centering
\begin{subfigure}[b]{0.7\linewidth}
\centering
\includegraphics[width=\linewidth]{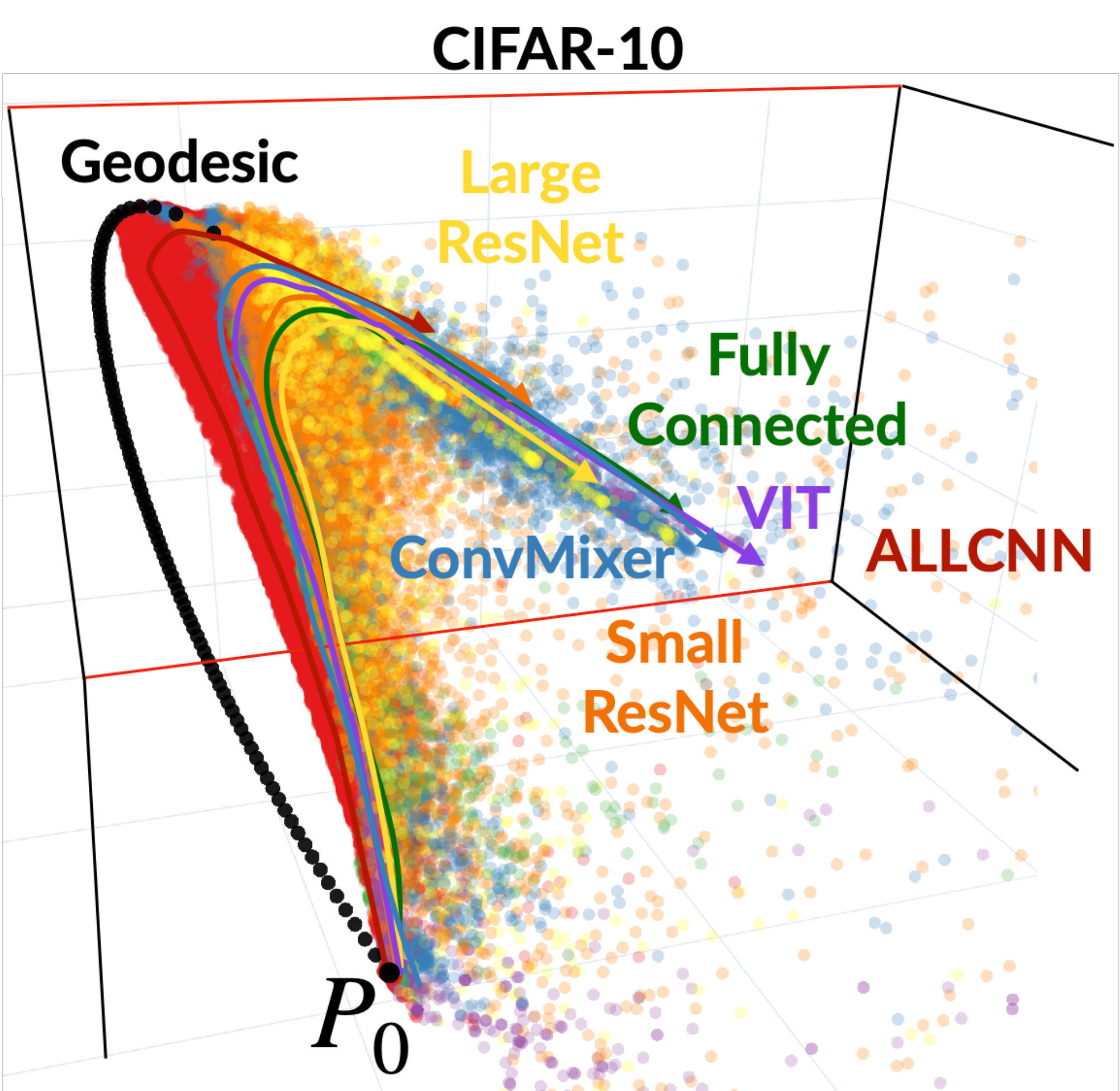}
\caption{}
\label{fig:all_models_train_3d}
\end{subfigure}%
\begin{subfigure}[b]{0.28\linewidth}
\centering
\includegraphics[width=\linewidth]{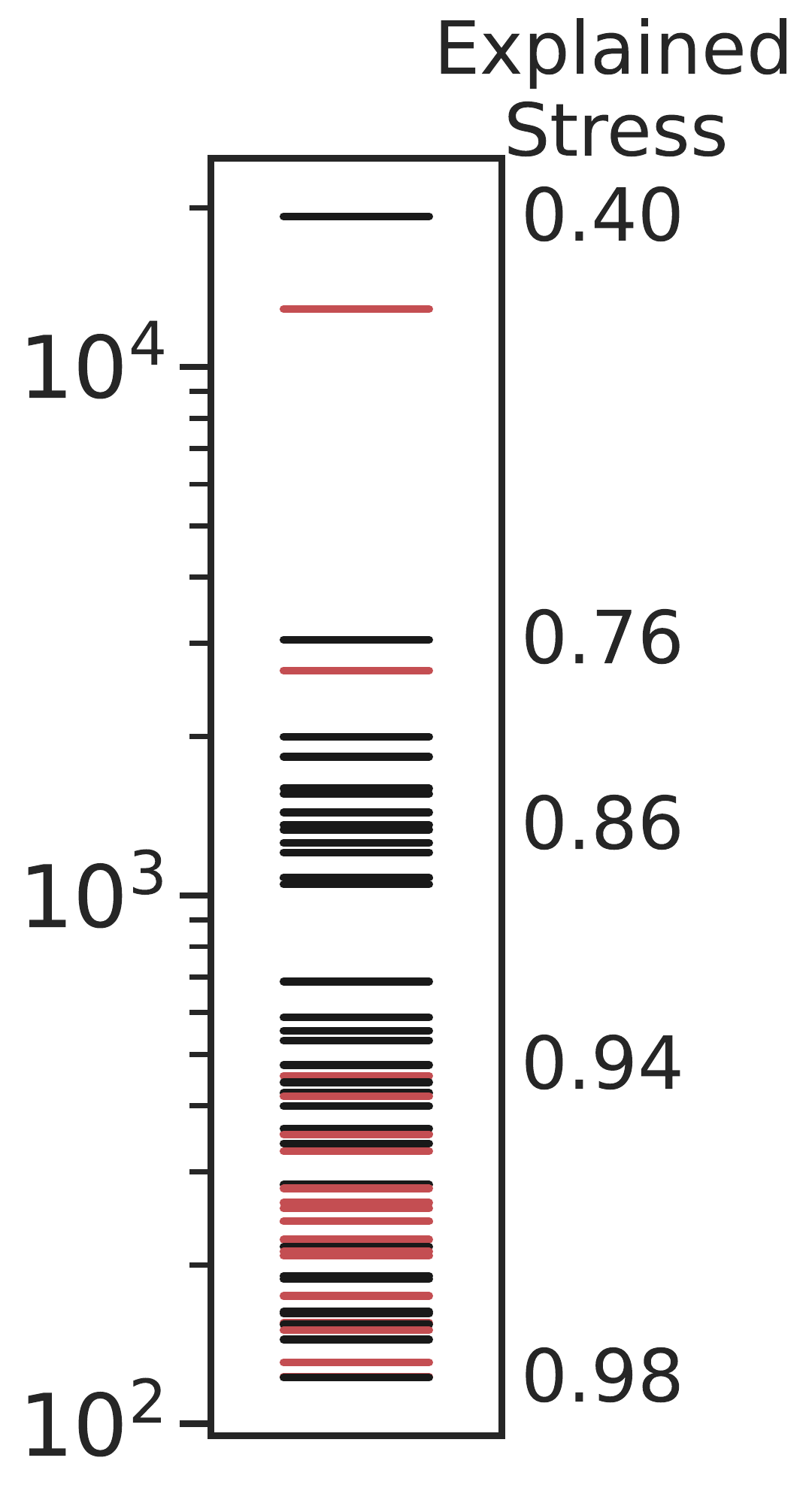}
\caption{}
\label{fig:all_models_train_eig_es}
\end{subfigure}
\begin{subfigure}[b]{0.75\linewidth}
\centering
\includegraphics[width=\linewidth]{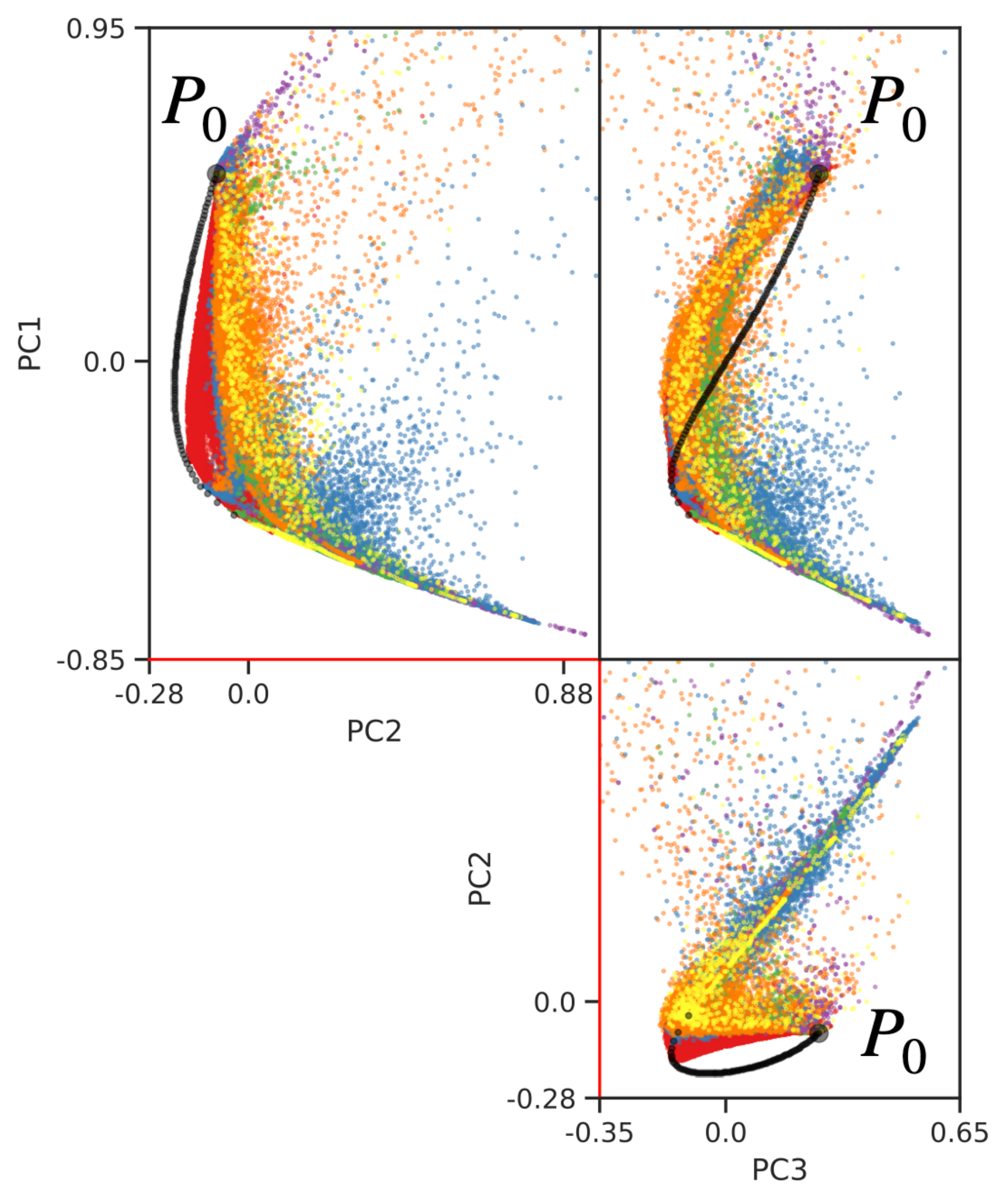}
\caption{}
\label{fig:all_models_train_2d}
\end{subfigure}
\begin{subfigure}[b]{0.55\linewidth}
\centering
\includegraphics[width=\linewidth]{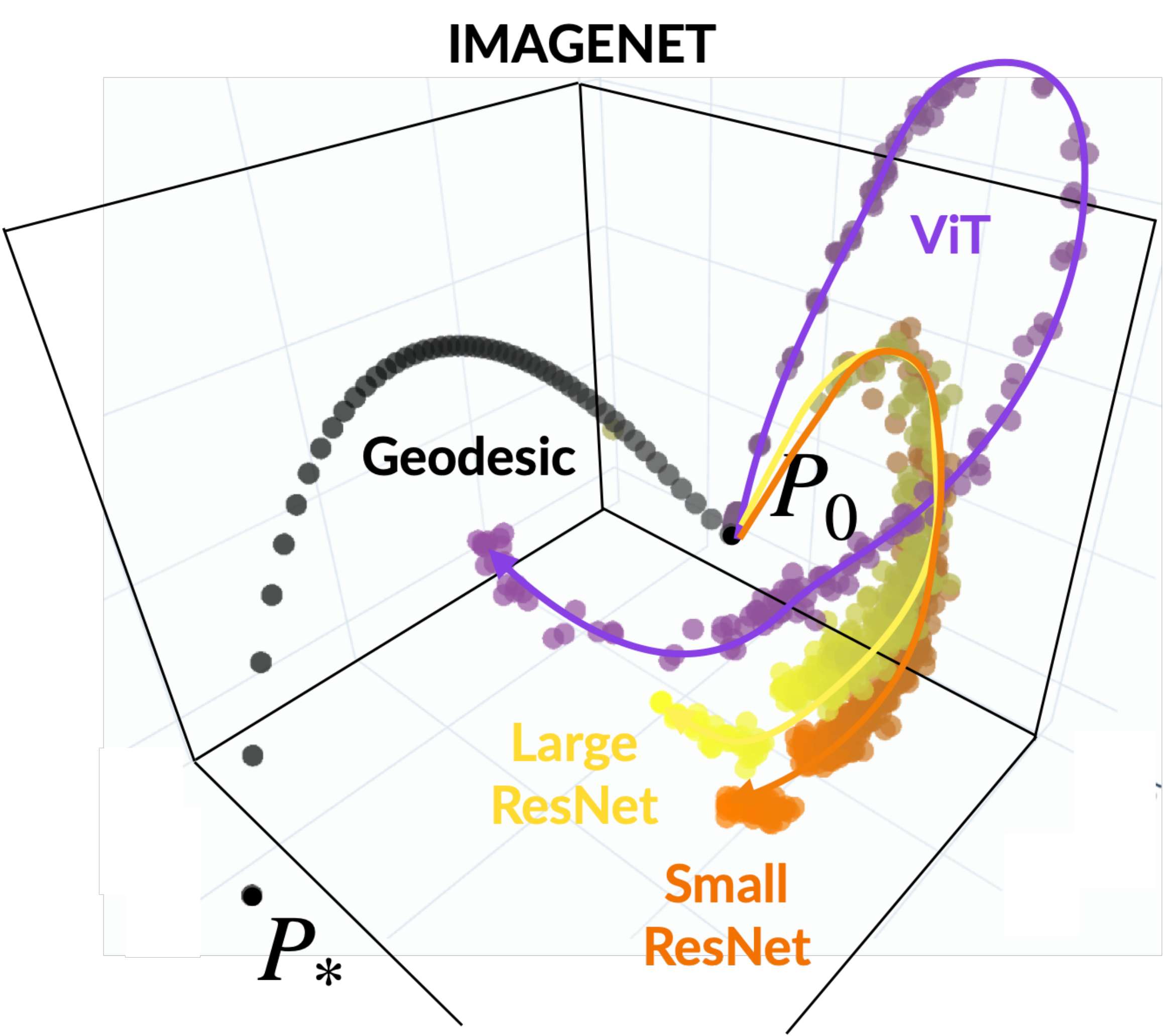}
\caption{}
\label{fig:imagenet_all_models_train_3d}
\end{subfigure}
\begin{subfigure}[b]{0.35\linewidth}
\centering
\includegraphics[width=\linewidth]{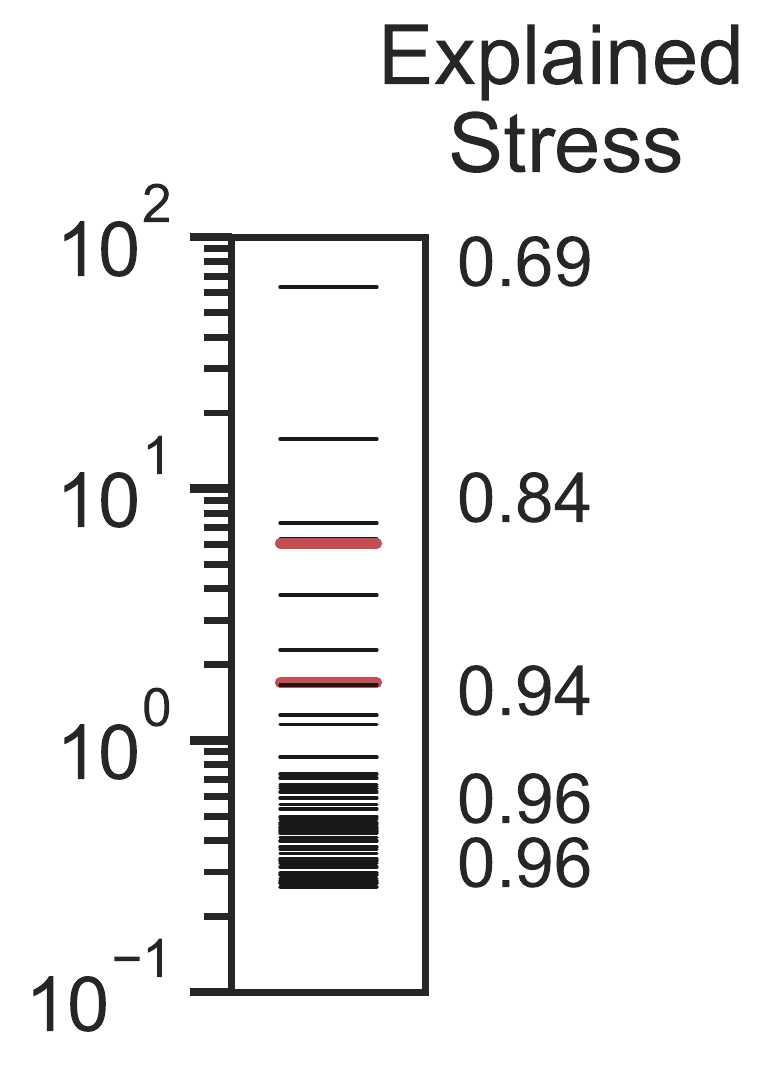}
\caption{}
\label{fig:imagenet_explained_stress_train}
\end{subfigure}
\caption{
The manifold of models along training trajectories of networks with different configurations (architectures denoted by different colors, optimization algorithms, hyper-parameters, and regularization mechanisms) is effectively low-dimensional for \textbf{(a)} CIFAR-10, and \textbf{(d)} ImageNet. Different configurations train along similar trajectories but are quite different from the geodesic between ignorance $P_0$ and truth $P_*$ (not seen here). The manifold is hyper-ribbon-like~\cite{transtrumGeometryNonlinearLeast2011}: eigenvalues of the InPCA distance matrix~\cref{eq:w} for CIFAR-10 \textbf{(b)} and ImageNet \textbf{(e)} are spread over a large range with the top few dimensions capturing a large fraction of the stress~\cref{eq:explained_stress} (numbers indicate explained stress in the top 1, 3, 10, 25 and 50 dimensions). Time-like coordinates corresponding to negative InPCA eigenvalues are red. \textbf{(c)}: a pairwise comparison for the first three principal components, note that PC2 is time-like (same data as~\textbf{(a)}). In \textbf{(a,d)}, we have drawn smooth curves denoting trajectories by hand to guide the reader.
}
\label{fig:all_models_train}
\end{figure}
\cref{fig:all_models_train_3d} shows the first three dimensions of the InPCA embedding of the probabilistic model in~\cref{eq:def:Pw} computed over samples in the training set. Each point corresponds to one model (i.e., one architecture, optimization algorithm, hyper-parameters, regularization, weight initialization and a particular checkpoint along the training trajectory) and is colored by the architecture. The explained stress~\cref{eq:explained_stress} of the first three dimensions is 76\% as shown in~\cref{fig:all_models_train_eig_es}; it increases to $98\%$ within the first 50 dimensions. The prediction space for CIFAR-10 has $4.5 \times 10^5$ dimensions ($N=5 \times 10^4$ and $C=10$); the rank of the distance matrix in InPCA is at most 151,407. For ImageNet, all networks are trained on the entire training set ($N=1.28 \times 10^6$) but we use a subset of the training samples ($N=50,000$) across $C=10^3$ classes  to calculate the embedding (i.e., the prediction space has $4.995 \times 10^7$ dimensions). For ImageNet, nearly 84\% of the explained stress is captured by the top three components of the InPCA embedding~\cref{fig:imagenet_all_models_train_3d}; this increases to 96\% in the top 50 dimensions. The fact that so few dimensions capture such a large fraction of the stress suggests that in spite of the huge diversity in the configurations of these networks, they all explore an effectively low-dimensional manifold in the prediction space during training.

\begin{figure}
\centering
\begin{subfigure}[b]{0.46\linewidth}
\centering
\includegraphics[width=\linewidth]{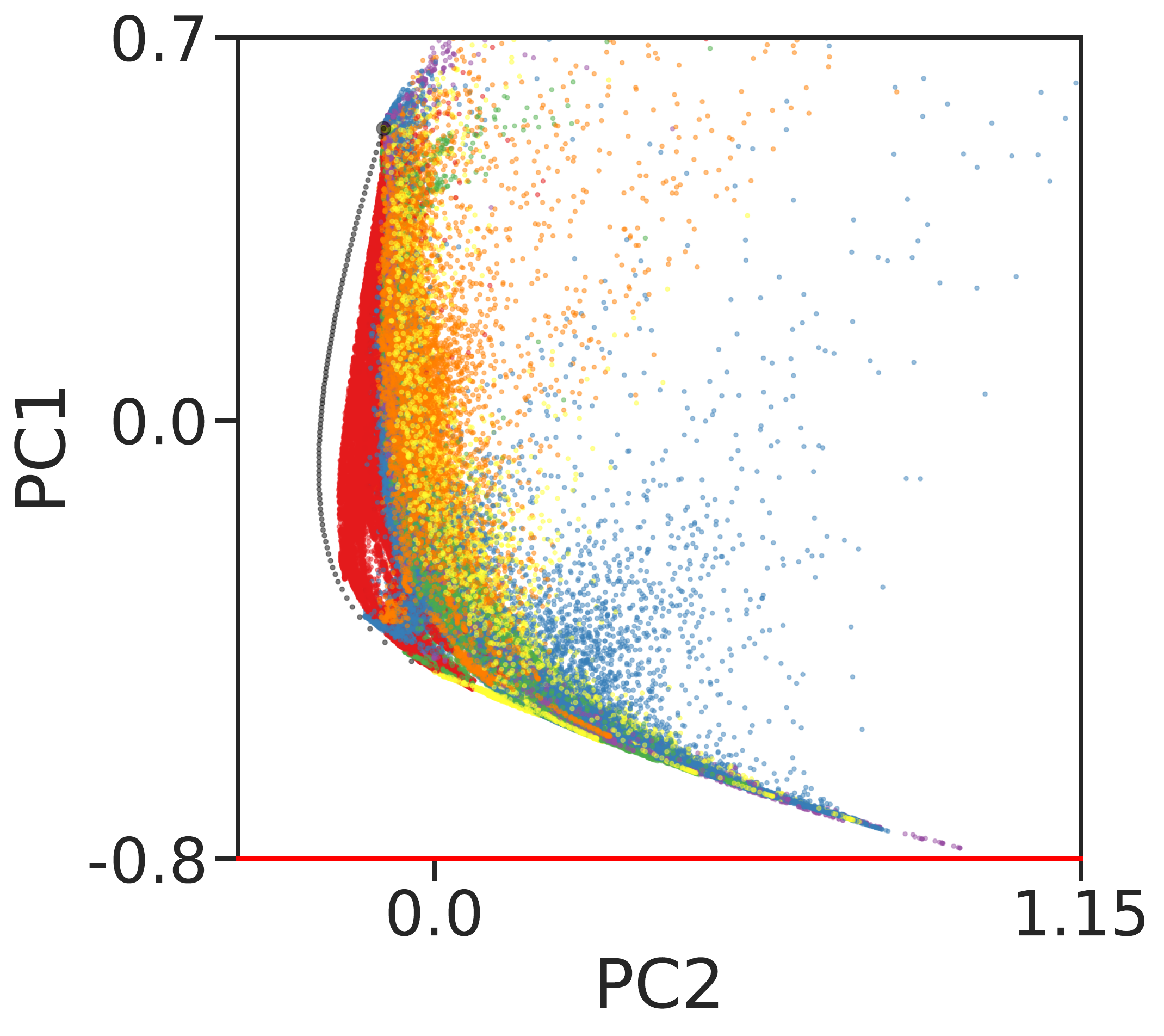}
\caption{}
\label{fig:all_models_train_2d_pc12}
\end{subfigure}
\begin{subfigure}[b]{0.49\linewidth}
\centering
\includegraphics[width=\linewidth]{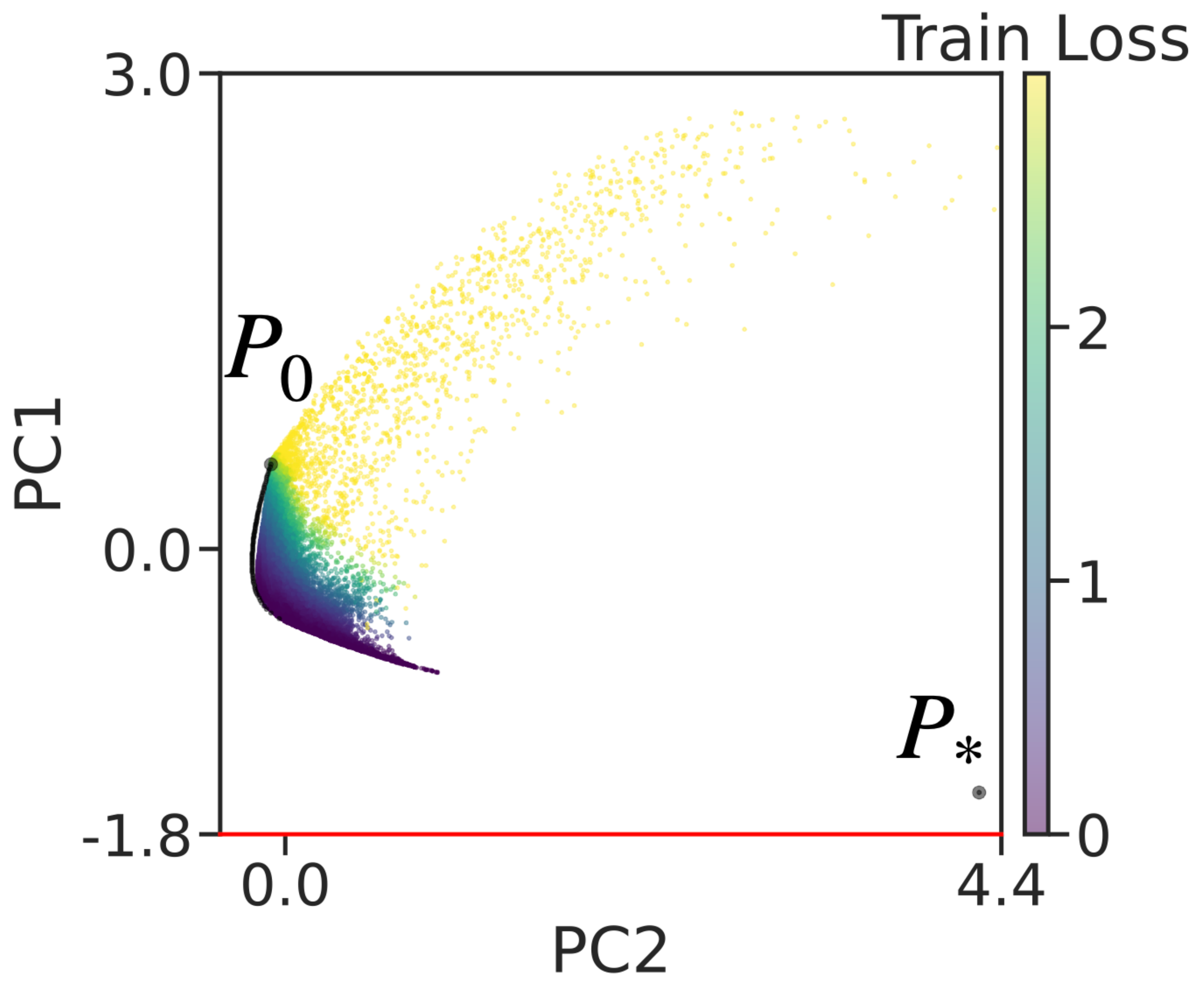}
\caption{}
\label{fig:all_models_train_2d_ps}
\end{subfigure}
\begin{subfigure}[b]{0.49\linewidth}
\centering
\includegraphics[width=\linewidth]{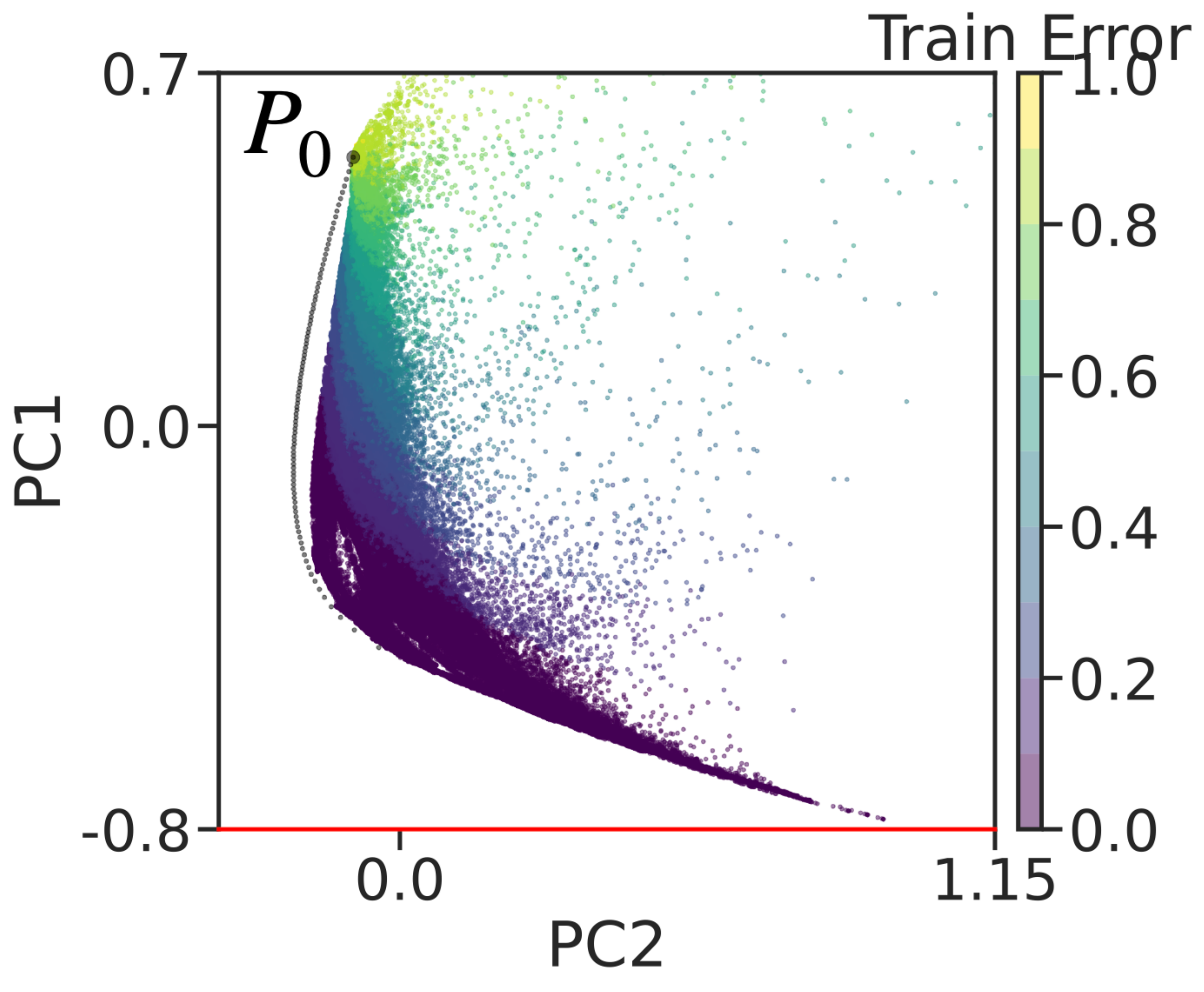}
\caption{}
\label{fig:all_models_train_2d_error}
\end{subfigure}
\begin{subfigure}[b]{0.49\linewidth}
\centering
\includegraphics[width=\linewidth]{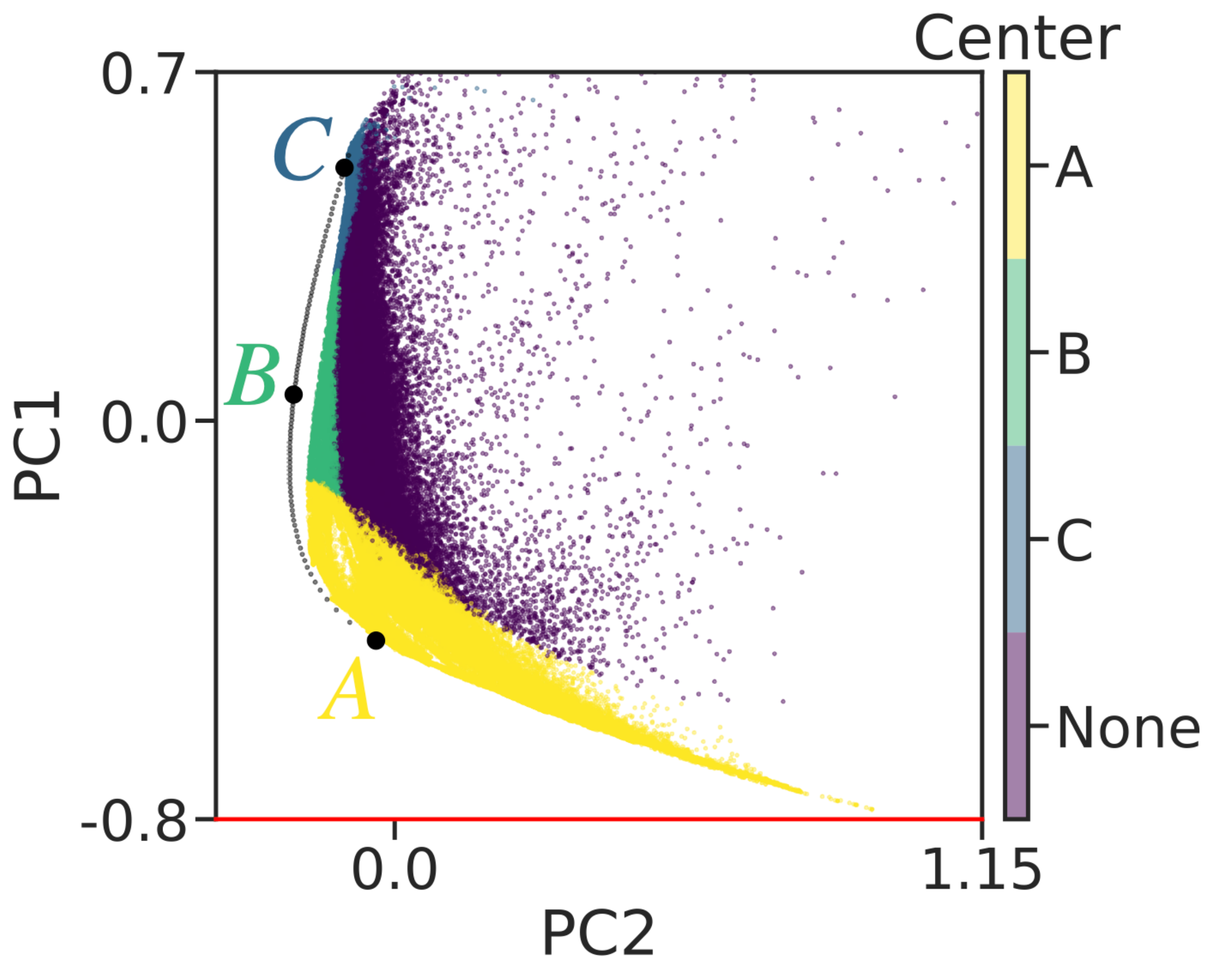}
\caption{}
\label{fig:all_models_train_2d_spread}
\end{subfigure}
\caption{Comparison of the top two principal components of an InPCA embedding of all models on CIFAR-10 colored by the architectures \textbf{(a)} (same as~\cref{fig:all_models_train_2d}), train loss \textbf{(b)}, which is two times the Bhattacharyya distance $\dB(P, P_*)$ for classification tasks like ours, train error in \textbf{(c)}, and by whether they are within a Bhattacharyya distance \textless\ 0.15 from models marked A, B, and C on the geodesic in \textbf{(d)}. These figures are discussed in the narrative and should be studied together with~\cref{fig:all_models_train_2d}.}
\label{fig:all_models_train_2d_details}
\end{figure}

Ignorance is marked by $P_0$. The truth $P_*$ is off the edge of the plot (see~\cref{fig:all_models_train_2d_ps}). The black curve denotes the embedding of the geodesic between $P_0$ and $P_*$ calculated using~\cref{eq:P_lambda}. 
Typical weight initialization schemes initialize models near $P_0$ irrespective of the configuration. Towards the end of training, models that trained well are close to the truth $P_*$ in terms of the Bhattacharyya distance. Note that if the truth $P_*$ has probabilities that are either zero or one (which is the case in our experiments), then the Bhattacharyya distance is one half of the cross-entropy loss used for classification. In this large prediction space, training trajectories of different configurations could be very diverse; on the contrary, not only do they all lie on an effectively low-dimensional manifold but trajectories of different configurations appear remarkably similar to each other. Sub-manifolds corresponding to each configuration seem to be rather similar; we will analyze this quantitatively in~\cref{fig:dendrogram_train_end}. For now, we note that probabilistic models learned by different architectures, training, and regularization methods, are very similar to each other---not only at the end of training when they fit the data but also along the entire training trajectory.

All trajectories seem to take a different path than the geodesic (shortest distance) path between $P_0$ and $P_*$. However, the geodesic is also largely captured by the top few dimensions of InPCA. Along the geodesic, all samples are trained towards the truth at the same rate, and so all models on it have zero training error. The deviation of paths away from the geodesic may reflect the learning of easy images early and confusing ones late, \change{perhaps due to first-order gradient-based methods}. We explore this further in~\cref{fig:all_models_train_d2geod,fig:all_models_test_d2geod}. \change{The geodesic corresponds to the trajectory of natural gradient descent~\cite{6790500}, which is not a first-order method. That the geodesic is faithfully represented in the low-dimensional embedding suggests that the low dimensionality observed in~\cref{fig:all_models_train_3d} is not a direct consequence of using gradient-based algorithms.}

All these observations also hold for networks trained on ImageNet. Note that in this case, the top three eigenvalues of InPCA are all positive; we have noticed this to be the case when the number of models embedded is small. The manifold of all trajectories is still effectively low-dimensional. Sub-manifolds spanned by ViTs and ResNets appear different from each other while sub-manifolds of the smaller and larger ResNet are quite similar; we will see in~\cref{fig:dendrogram_train_end} that architectures are the primary distinguishing factors of different training trajectories. In this case, all three architectures are quite different from the geodesic. Training trajectories do not end as close to truth $P_*$ as those of CIFAR-10; for ImageNet, the trajectories end at a progress~\cref{eq:tw} close to 0.9. This should not be surprising because typically networks trained on ImageNet do not achieve zero training error (zero training error can be achieved but they perform very poorly on the test data).

\paragraph{Characterizing the details of the train manifold}
\cref{fig:all_models_train_2d_pc12} shows a pairwise comparison for the first three principal components of InPCA (same data as that of~\cref{fig:all_models_train_3d}). Qualitatively, the first principal component, which is space-like, distinguishes models according to their distance to the truth $P_*$ (i.e., half of the cross-entropy loss). The second principal component, however, is time-like because the second eigenvalue of InPCA is negative; shown in red in~\cref{fig:all_models_train}. The third principal component is again space-like. All models that train well have small Bhattacharyya distances to the truth $P_*$ towards the end of training; they also have small errors (zero in almost all cases). But these probabilistic models are different from each other, and they are also different from the truth $P_*$. Our visualization technique is emphasizing these subtle differences using  all coordinates, including the imaginary coordinate corresponding to the negative eigenvalue. \cref{fig:all_models_train_2d_ps} shows the train loss of all models (colored by purple for small, yellow for large). Even if the truth looks far away from them visually (\textgreater\ 4 in a Euclidean sense), models colored purple in~\cref{fig:all_models_train_2d_ps} have small distances from the truth $\dB(P_w, P_*) < 0.2$; incidentally their Minkowski distance to the truth in the top three coordinates is negative.

\begin{figure}
\centering
\begin{subfigure}[b]{0.49\linewidth}
\includegraphics[width=\linewidth]{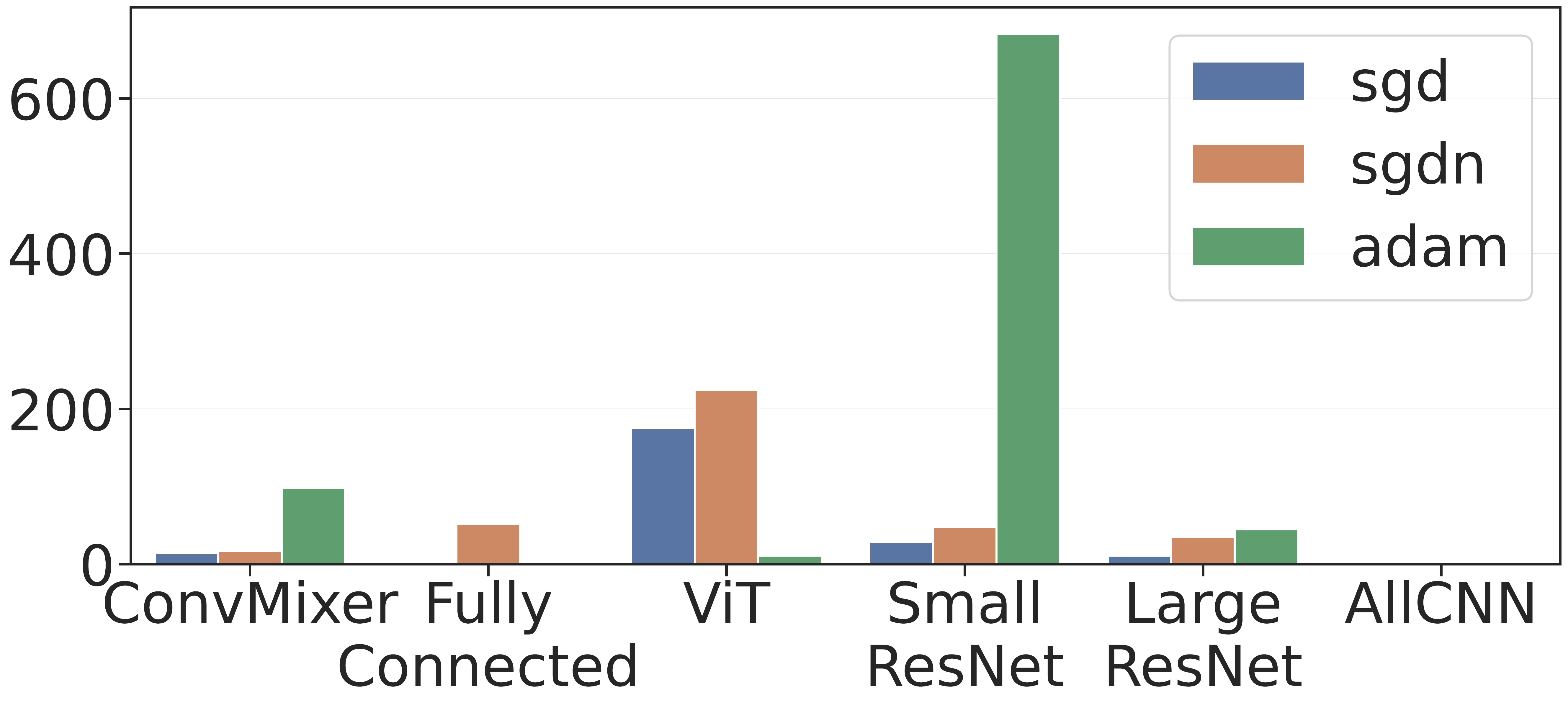}
\caption{}
\label{fig:all_models_train_by_model}
\end{subfigure}
\begin{subfigure}[b]{0.49\linewidth}
\includegraphics[width=\linewidth]{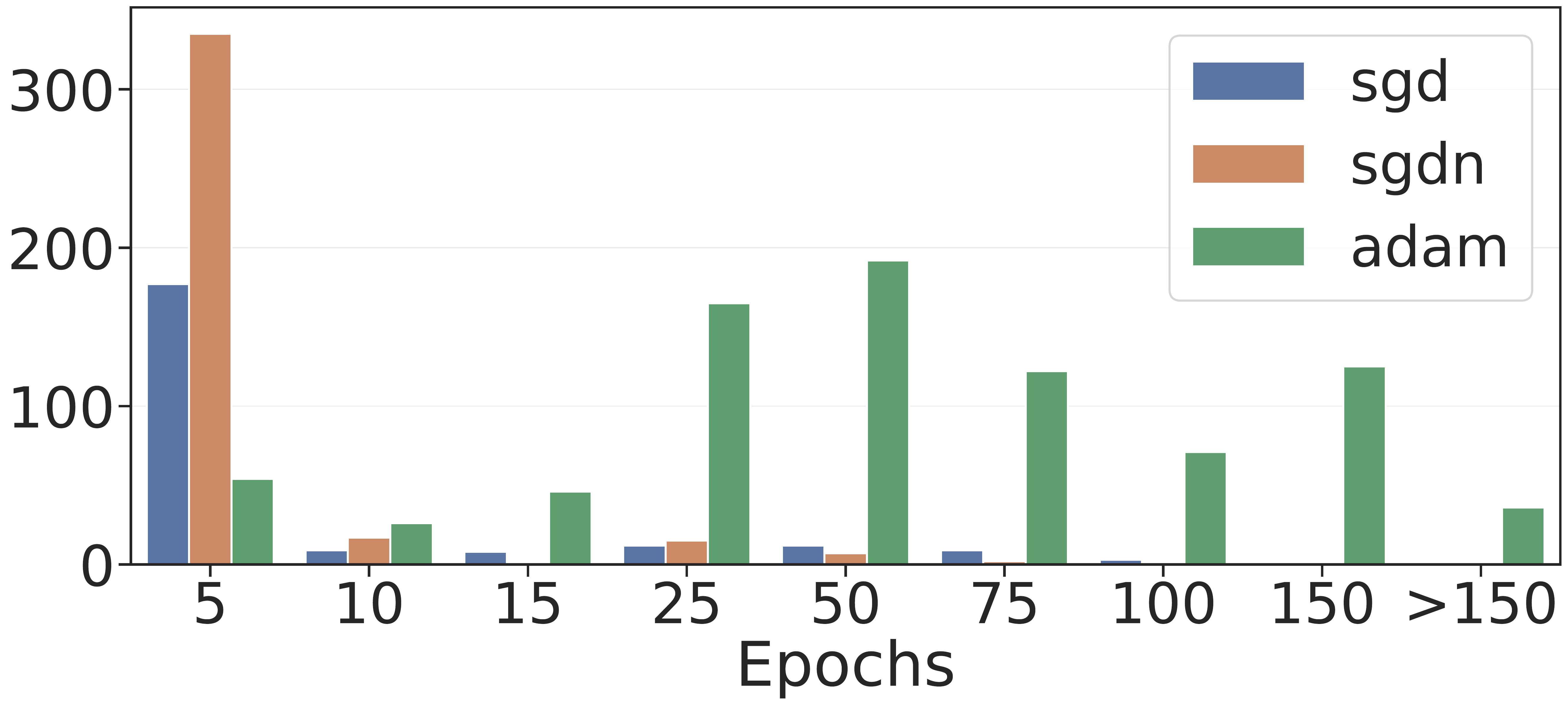}
\caption{}
\label{fig:all_models_train_by_nepochs}
\end{subfigure}
\caption{Number of models $P$ with $\dB(P, P_*) > 2$ (that are away from the main manifold) stratified by \textbf{(a)} architectures and \textbf{(b)} the number of epochs.}
\label{fig:all_models_train_adam}
\end{figure}

In~\cref{fig:all_models_train_2d_ps}, the spread of points (yellow) near $P_0$ consists of some models that have 90\% error (same as that of ignorance). There are 1500 such points, coming from 370 different trajectories (over 85\% of points are from 145 trajectories). Over half of these high error deviating networks (see~\cref{fig:all_models_train_by_nepochs}) eventually trained to zero error. These models have the same error as that of ignorance $P_0$ but the visualization method distinguishes them from ignorance because their probabilities are not uniform. The spread of the points in the visualization in this case is therefore coming from differences in the probabilities. These models can be brought back to the manifold of good training trajectories simply by training them further. Now notice the points colored purple in~\cref{fig:all_models_train_2d_spread}. These models have a large Bhattacharyya distance (\textgreater\ 0.15) from points marked $A, B$ or $C$ on the geodesic (which corresponds to progress of 0.01, 0.5 and 0.99 respectively). \cref{fig:all_models_train_2d_error} shows that these models also have very different errors from each other. This spread of points away from the manifold is therefore also coming from large differences in the probabilities.

Now notice the blue cluster of models (ConvMixer) in~\cref{fig:all_models_train_2d_pc12}; as~\cref{fig:all_models_train_2d_spread} shows, the distance of a bulk of these ConvMixer models to point A is small (\textless\ 0.1). And~\cref{fig:all_models_train_2d_error} suggests that these models have error \textless\  10\% (some also have larger errors). In this region, the spread of the points in the visualization is coming predominantly from the small differences in the probabilities.

\cref{fig:all_models_train_by_model} studies models that are away from the manifold, with $\dB(P, P_*) > 2$ (yellow in~\cref{fig:all_models_train_2d_ps}). For ConvMixer and the two residual networks, a majority of these models were trained by Adam. No AllCNN models were away from the manifold. \cref{fig:all_models_train_by_nepochs} stratifies these models by the optimization algorithm. In early stages of training, these are networks trained with SGD or SGD with Nesterov's acceleration with large batch-sizes (more than 500); this accounts for about 35\% of the models. Adam is primarily responsible for models that are away from the manifold at later stages of training (about 55\% of the points). We speculate that this could be related to poorer test errors of Adam than SGD for image classification tasks.

\subsection*{The manifold of predictions on the test data is also effectively low-dimensional, with more significant differences among architectures}

\begin{figure}
\centering
\begin{subfigure}[b]{0.7\linewidth}
\centering
\includegraphics[width=\linewidth]{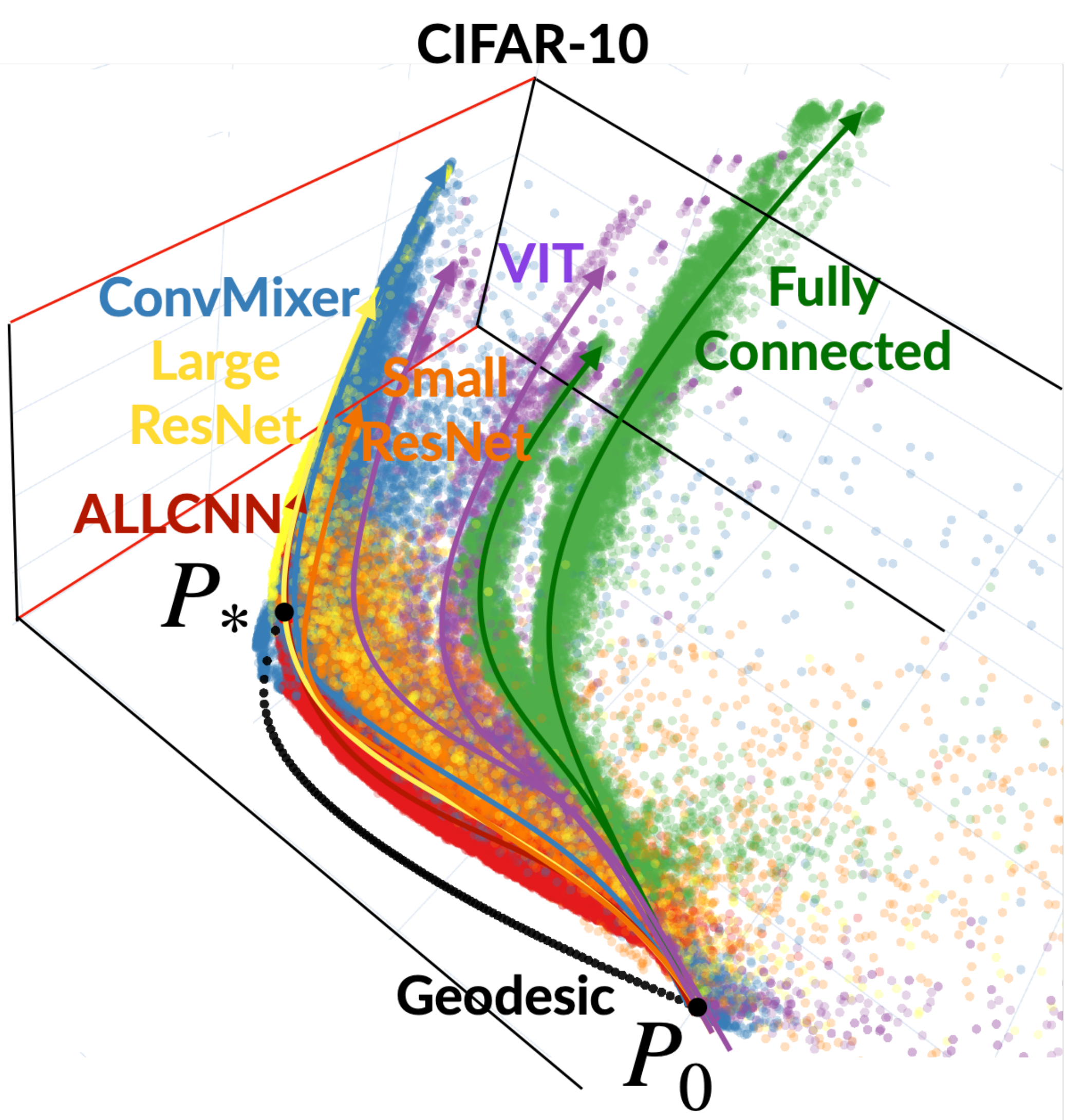}
\caption{}
\label{fig:all_models_test_3d}
\end{subfigure}%
\begin{subfigure}[b]{0.29\linewidth}
\centering
\includegraphics[width=\linewidth]{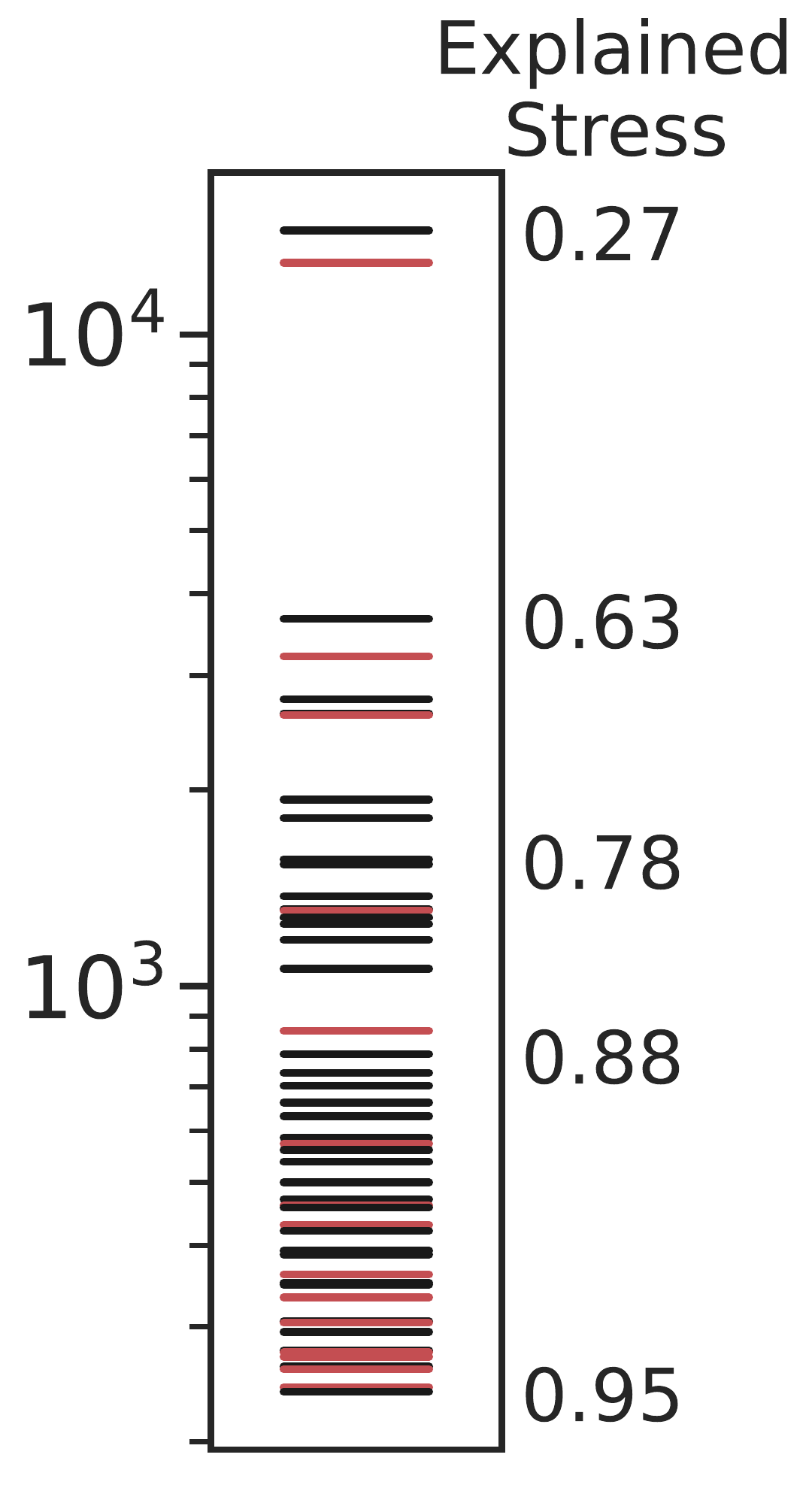}
\caption{}
\label{fig:all_models_test_eig_es}
\end{subfigure}
\begin{subfigure}[b]{0.75\linewidth}
\centering
\includegraphics[width=\linewidth]{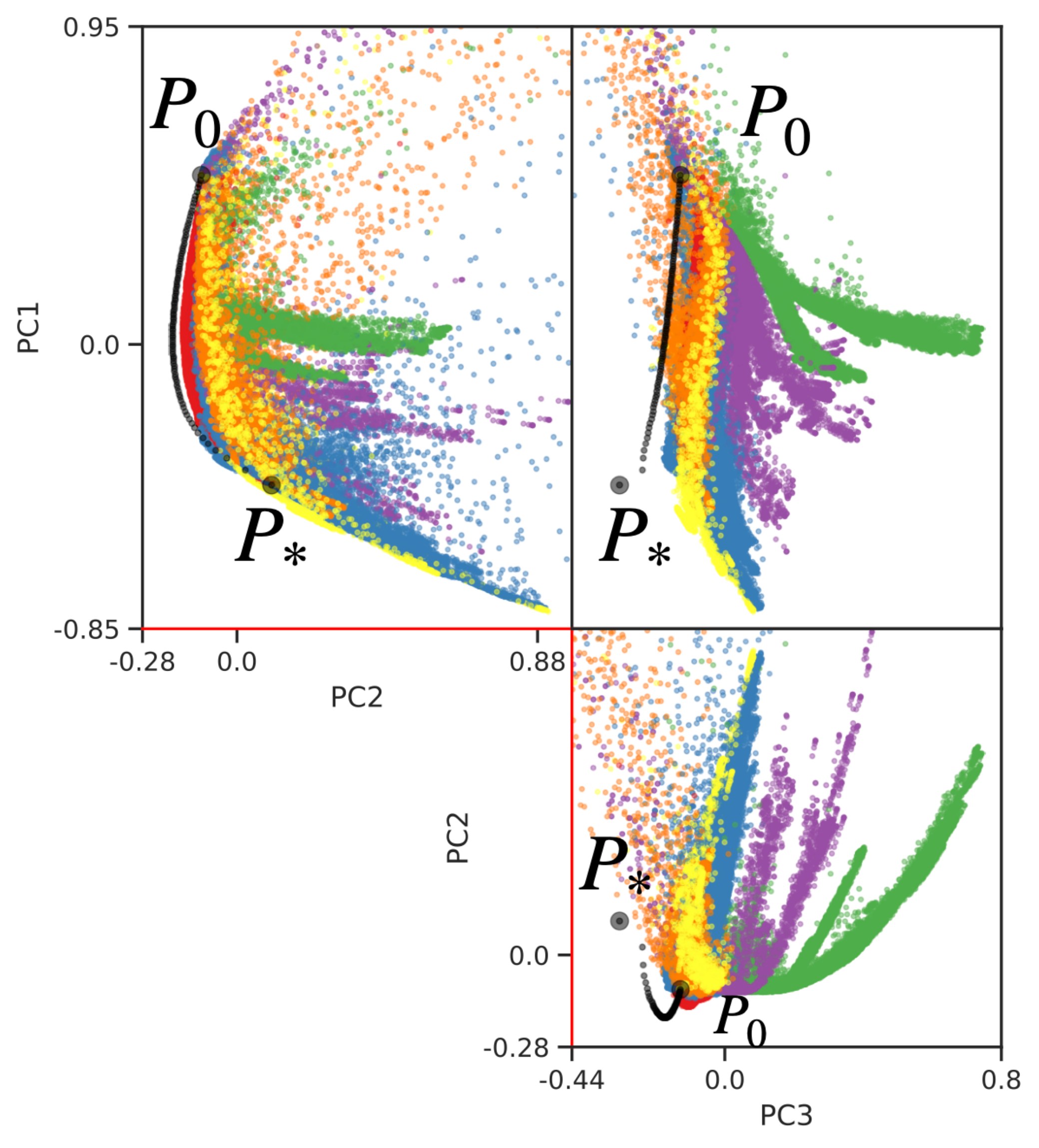}
\caption{}
\label{fig:all_models_test_2d}
\end{subfigure}
\begin{subfigure}[b]{0.55\linewidth}
\centering
\includegraphics[width=\linewidth]{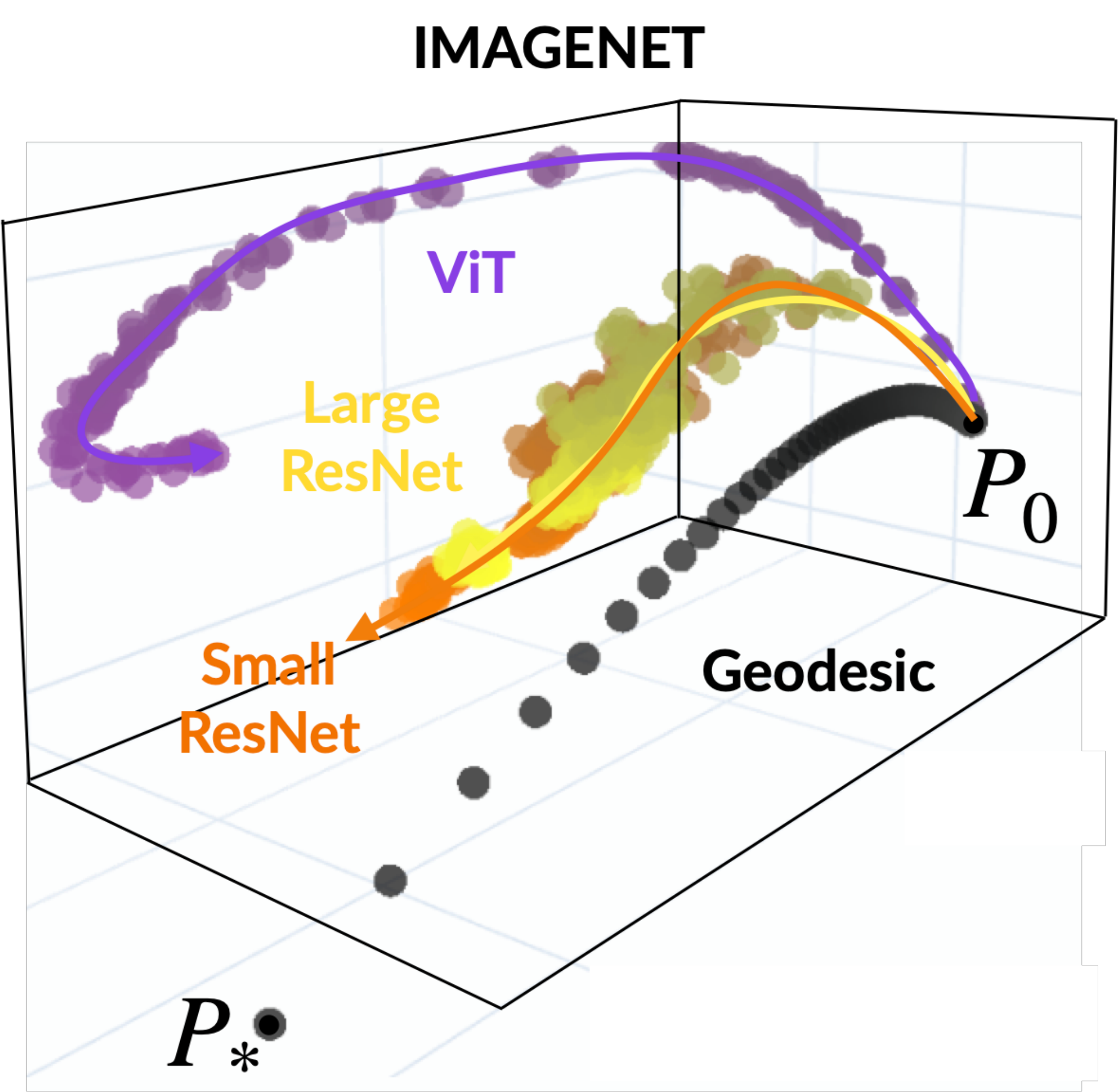}
\caption{}
\label{fig:imagenet_all_models_test_3d}
\end{subfigure}%
\begin{subfigure}[b]{0.35\linewidth}
\centering
\includegraphics[width=\linewidth]{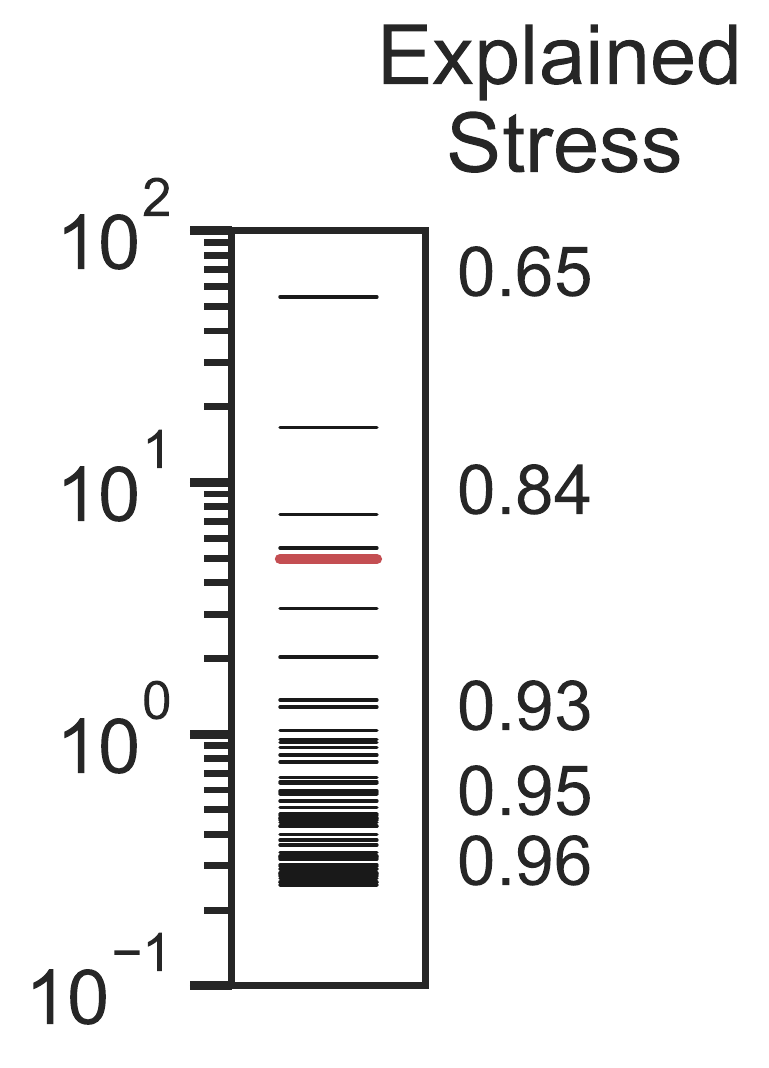}
\caption{}
\label{fig:imagenet_explained_stress_test}
\end{subfigure}
\caption{
Predictions on the test data of networks with different configurations (architectures denoted by different colors, different optimization algorithms and regularization mechanisms) on CIFAR-10 in \textbf{(a)} and on ImageNet in \textbf{(d)} is also effectively low-dimensional. Trajectories of different architectures are distinctive on the test data. Test manifold is also hyper-ribbon-like: eigenvalues of the InPCA distance matrix~\cref{eq:w} for CIFAR-10 \textbf{(b)}  and ImageNet \textbf{(e)}  are spread over a large range and the top few dimensions capture a large fraction of the stress~\cref{eq:explained_stress} (numbers indicate explained stress in the top 1, 3, 10, 25 and 50 dimensions. \textbf{(c)} shows a pairwise comparison for the first three principal components for CIFAR-10 models. PC1-PC2 of~\cref{fig:all_models_train_2d} look quite similar to those of \textbf{(c)}. In \textbf{(a,d)}, we have drawn smooth curves denoting trajectories by hand to guide the reader.
}
\label{fig:all_models_test}
\end{figure}

\cref{fig:all_models_test_3d} shows the first three dimensions of the InPCA embedding of predictions on the test data using the same networks as that of~\cref{fig:all_models_train_3d}. The explained stress of the first three dimensions is still high (63\%) and it increases to 95\% within the first 50 dimensions; these numbers are smaller than those for the training data. For CIFAR-10, the prediction space has $9\times 10^4$ dimensions ($N = 10^4$ and $C = 10$) and for ImageNet the prediction space has $4.995 \times 10^7$ dimensions ($N=50,000$ and $C=1000$). This suggests that in spite of the vast diversity in configurations of these networks, their trajectories in the prediction space of the test samples also lie on an effectively low-dimensional manifold.

The test manifold is broadly similar to the train manifold in~\cref{fig:all_models_train_3d}. Trajectories begin near ignorance ($\dB(P, P_0) < 0.6$ at the start of training) but they do not always end near $P_*$. This is expected because different architectures have different test loss/errors at the end of training. The Bhattacharyya distance to the truth is one half of the test cross-entropy loss; models with poor test loss should be farther from $P_*$ than those with a small test loss. Bhattacharyya distances of the end points of trajectories are as large as 0.58 for the test manifold compared to 0.02 for the train manifold after excluding models with train error \textgreater\ 10\%.

Trajectories of different configurations seem to be more dissimilar in~\cref{fig:all_models_test_3d} than those in~\cref{fig:all_models_train_3d}; networks of different architectures have more distinctive test trajectories. We have analyzed these differences quantitatively in~\cref{fig:dendrogram_test_end}. But it is remarkable that even if different architectures have quite different trajectories, different models with the same architecture predict similarly on the test data. In other words, all fully-connected networks make the same kind of mistakes, and all convolutional networks are correct on generally the same samples. For fully-connected networks and ViTs, we see two different test trajectories corresponding to the two kinds of data augmentation techniques. For convolutional architectures, there are minor differences in test trajectories due to augmentation. This could be because we used randomly cropped images for augmentation: convolutional networks are relatively insensitive to random crops because their features have translational equivariance.

\cref{s:app:analysis_test_trajectories} provides a detailed analysis of the test trajectories.

\paragraph{Embedding probabilistic models along train and test trajectories into the same space}

\begin{figure}
\centering
\begin{subfigure}[b]{0.55\linewidth}
\centering
\includegraphics[width=\linewidth]{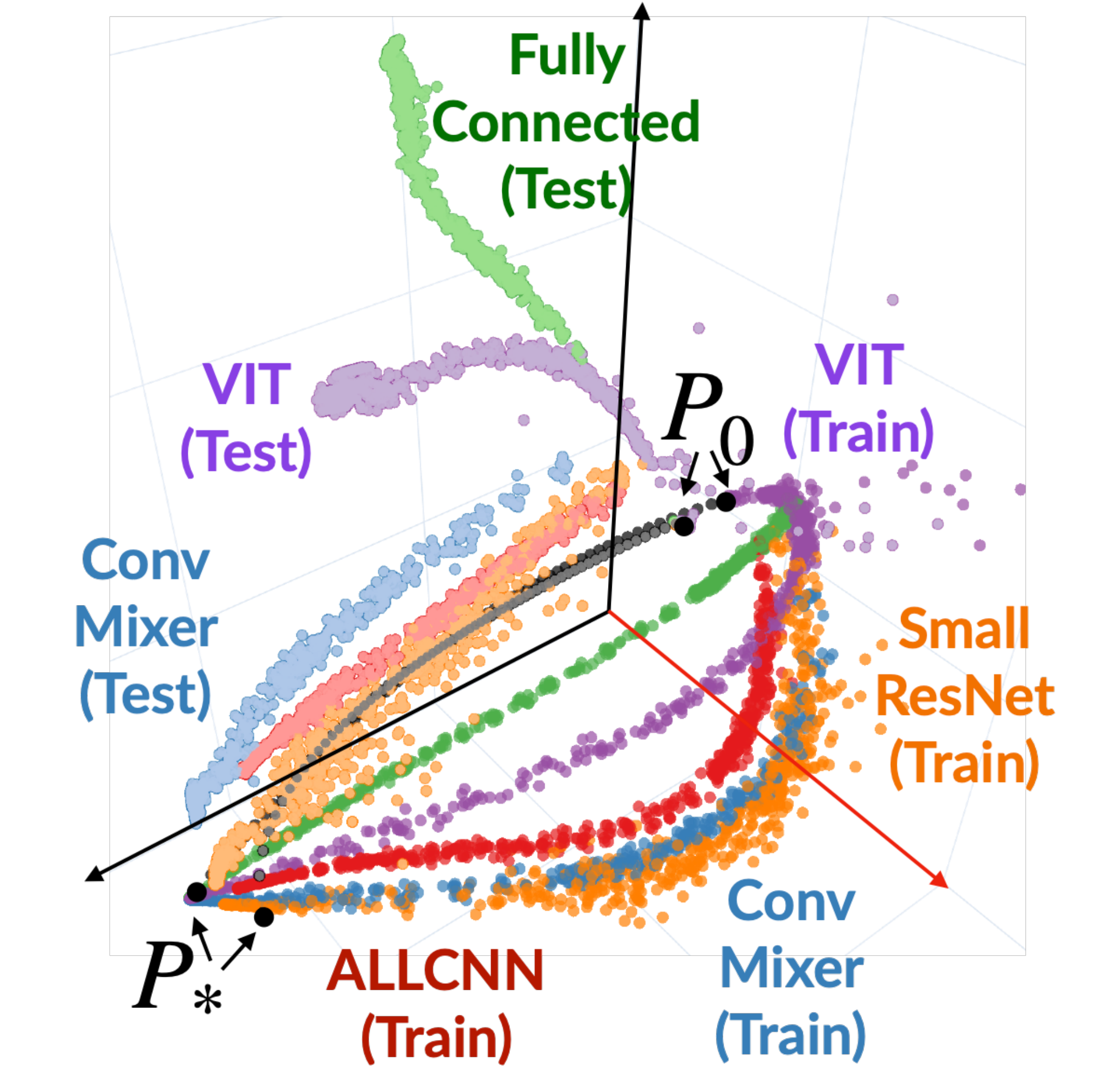}
\caption{}
\label{fig:train_test_inpca_3d}
\end{subfigure}
\begin{subfigure}[b]{0.43\linewidth}
\centering
\includegraphics[width=\linewidth]{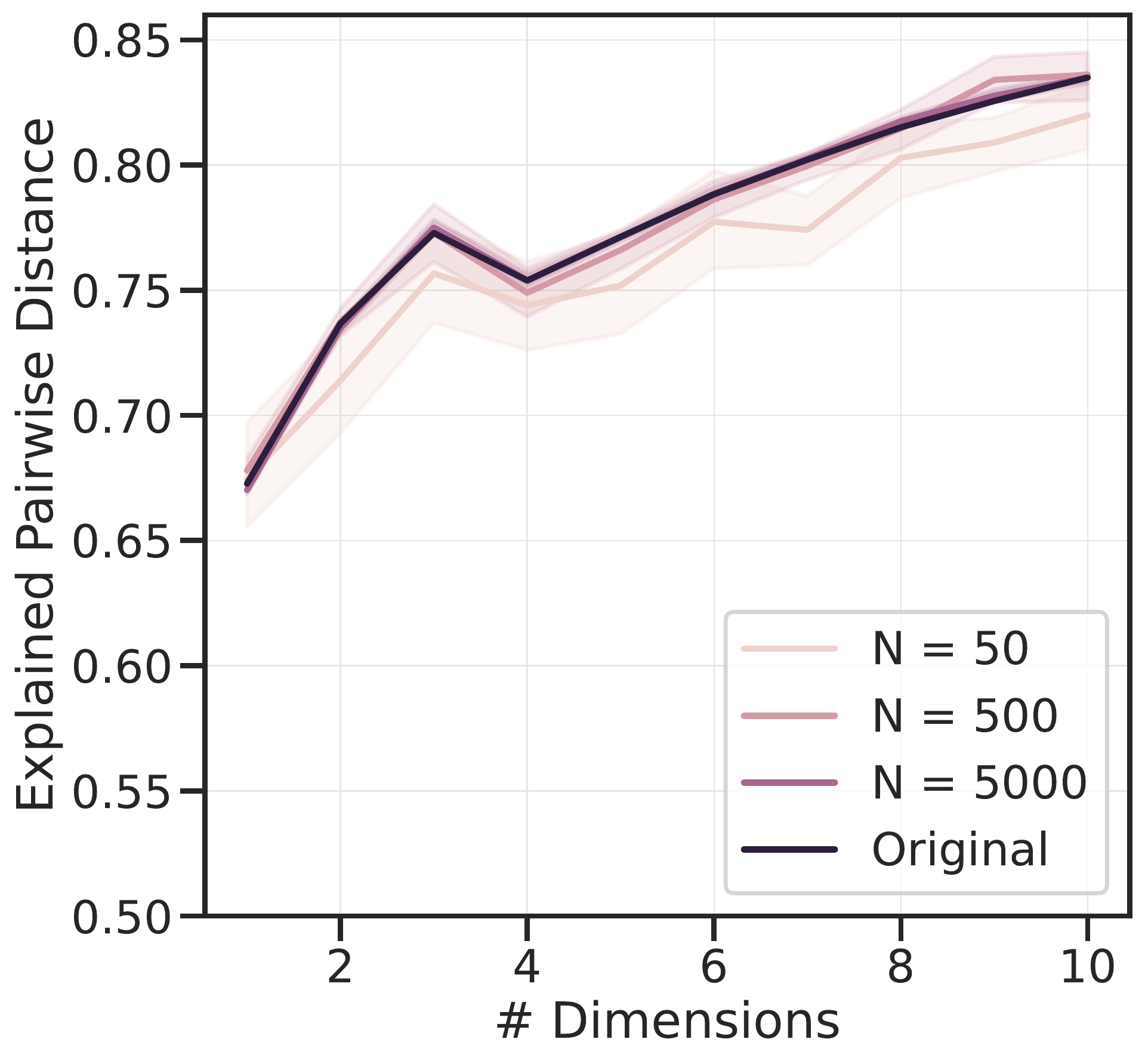}
\caption{}
\label{fig:explained_pairwise_distances_subsample_train}
\end{subfigure}
\caption{\textbf{(a)}: A joint embedding of a subset of networks on CIFAR-10 using their predictions on samples from both the train (bold) and test (translucent) sets. \textbf{(b)}: the explained pairwise Bhattacharyya distances computed using~\cref{eq:explained_pairwise_distances} is quite high for models on the train data after embedding them into an InPCA embedding computed using a small number of samples ($N=5000$, $N=500$, and $N=50$) in the train data. \cref{s:app:subset_inpca} discusses this further.}
\label{fig:train_test_together}
\end{figure}

So far, we have analyzed train and test manifolds independently of each other. Indeed, probabilistic models~\cref{eq:def:Pw} corresponding to train and test data belong to different sample spaces, even if the two were created from the same underlying weights. It is however useful to visualize the two manifolds in the same space to understand how progress towards the truth in the train space results in progress towards the truth in the test space.

We first computed InPCA coordinates using probabilistic models on train data, let us denote one such model with weights $u$ as $P_u$. We then used the procedure developed in~\cref{eq:w_expanded} to embed test models into these coordinates as follows. Let us denote by $P_u'$ the model on the test data for the same weights $u$. Calculate
\beq{
    \aed{
    \textstyle & W'_{uv} = -\f{\dB(P'_u, P'_v)}{2} +\\
    &\textstyle \f{1}{2 m} \rbr{\sum_{u'} \dB(P_u, P_{u'}) + \dB(P_v, P_{u'}) - \f{1}{m} \sum_{v'} \dB(P_{u'}, P_{v'})};
    }
    \label{eq:w_train_test}
}
for all models $P_u$ and $P_v$. The first term is the distance between two test models but the second term is computed using only train data and is the same as that of~\cref{eq:w_expanded}. The embedding of a test model $P_w'$ is set to be $X'_w = \sum_{u=1}^{n} W'_{w,u} U_{u} \lvert\L_{uu}\rvert^{-1/2}$ using the eigenvectors and eigenvalues of the train embedding. The procedure in~\cref{eq:w_expanded} was intended to embed new models of the same set of samples into an existing embedding. This present, somewhat peculiar, trick works when the number of train models and the number of test models are the same (which is the case for us), and when the second term in~\cref{eq:w_train_test} is close to its counterpart in~\cref{eq:w_expanded} (which is expected if there is self-averaging).

We first built an InPCA embedding using the train models and then used the procedure in~\cref{eq:w_train_test} to calculate the coordinates of the test models and obtained~\cref{fig:train_test_inpca_3d}. Observations drawn from this procedure are qualitatively the same as those from~\cref{fig:all_models_train,fig:all_models_test}, e.g., train and test trajectories of different architectures still lie on similar manifolds, test trajectories of AllCNN, ConvMixer and Small ResNet are close to each other, and test trajectories of Fully-Connected and ViT architectures are far from the others. The explained pairwise distances for the test models using the InPCA coordinates computed from the train models are also consistent with those obtained from embedding the test models independently like~\cref{fig:all_models_test_3d}; 0.52 versus 0.56 in the top 10 dimensions, respectively. This indicates that pairwise distances in the test data are well-preserved by the InPCA coordinates constructed using pairwise distances on the train data. When two models differ on the train data, they also differ in a similar way on the test data.

We also built a new InPCA embedding using pairwise Bhattacharyya distances in~\cref{eq:dB} calculated using only a subset of the samples. \cref{fig:explained_pairwise_distances_subsample_train,fig:explained_pairwise_distances_subsample,fig:subsampling} show the result of using the procedure in~\cref{eq:w_train_test} to project the original distance matrix into the coordinates of this new InPCA. The explained pairwise distance of the original checkpoints is consistently quite high, even when as few as $N=50$ or $N=10$ samples are used to calculate the embedding out of the 50,000 and 10,000 samples for train and test sets respectively. This suggests that our techniques for analysis of high-dimensional models can also be used on very large datasets. For ImageNet, where $C=1000$, we have also noticed that the InPCA embedding looks similar if we first project the output probabilities into a smaller space by multiplying by a random matrix (with columns that sum up to 1).

\subsection*{Architectures---not training or regularization schemes---primarily distinguish training trajectories in the prediction space}

\begin{figure}
\centering
\begin{subfigure}[b]{\linewidth}
\centering
\includegraphics[height=1.7\linewidth]{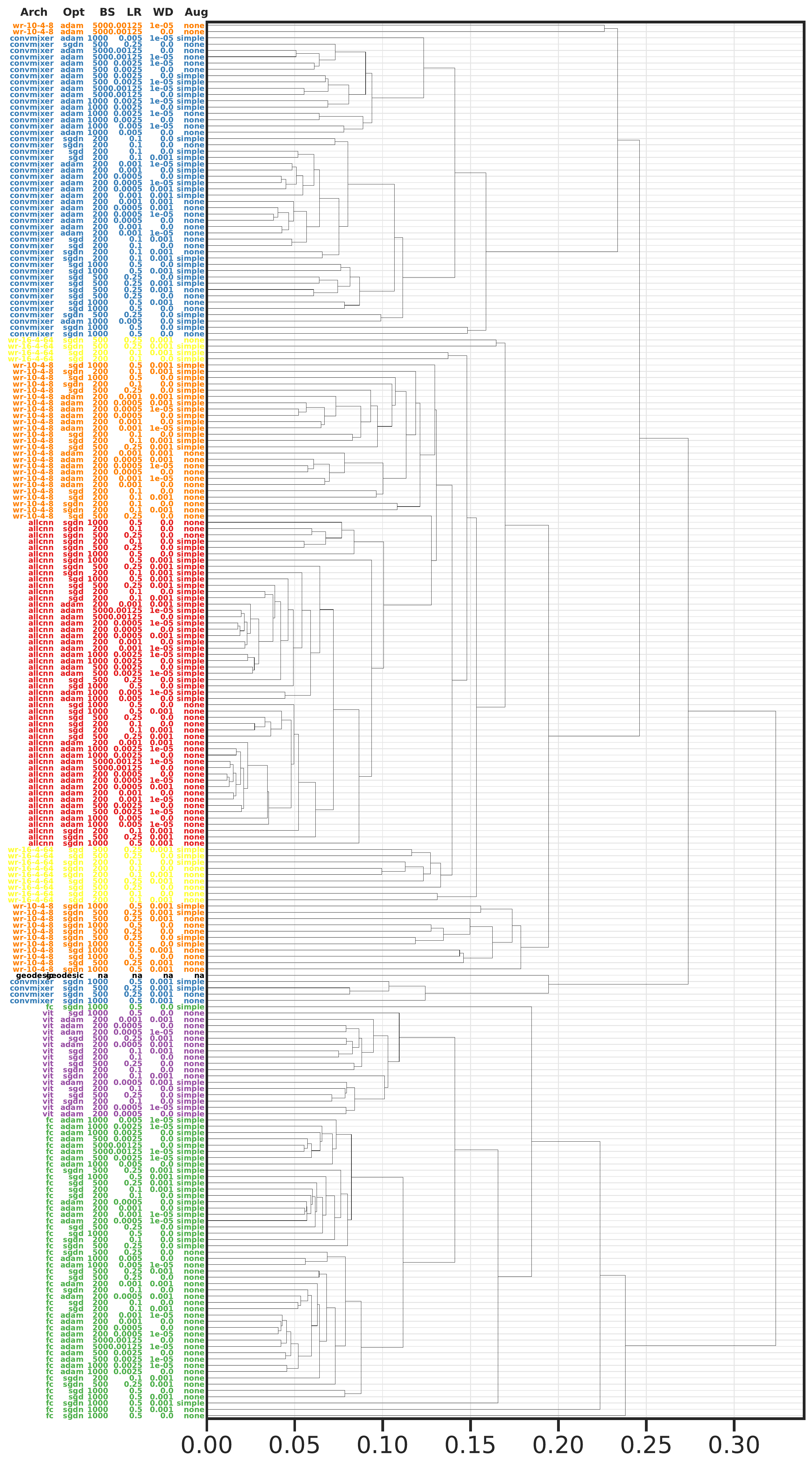}
\caption{}
\label{fig:dendrogram_train_end}
\end{subfigure}
\begin{subfigure}[b]{0.49\linewidth}
\centering
\includegraphics[width=\linewidth]{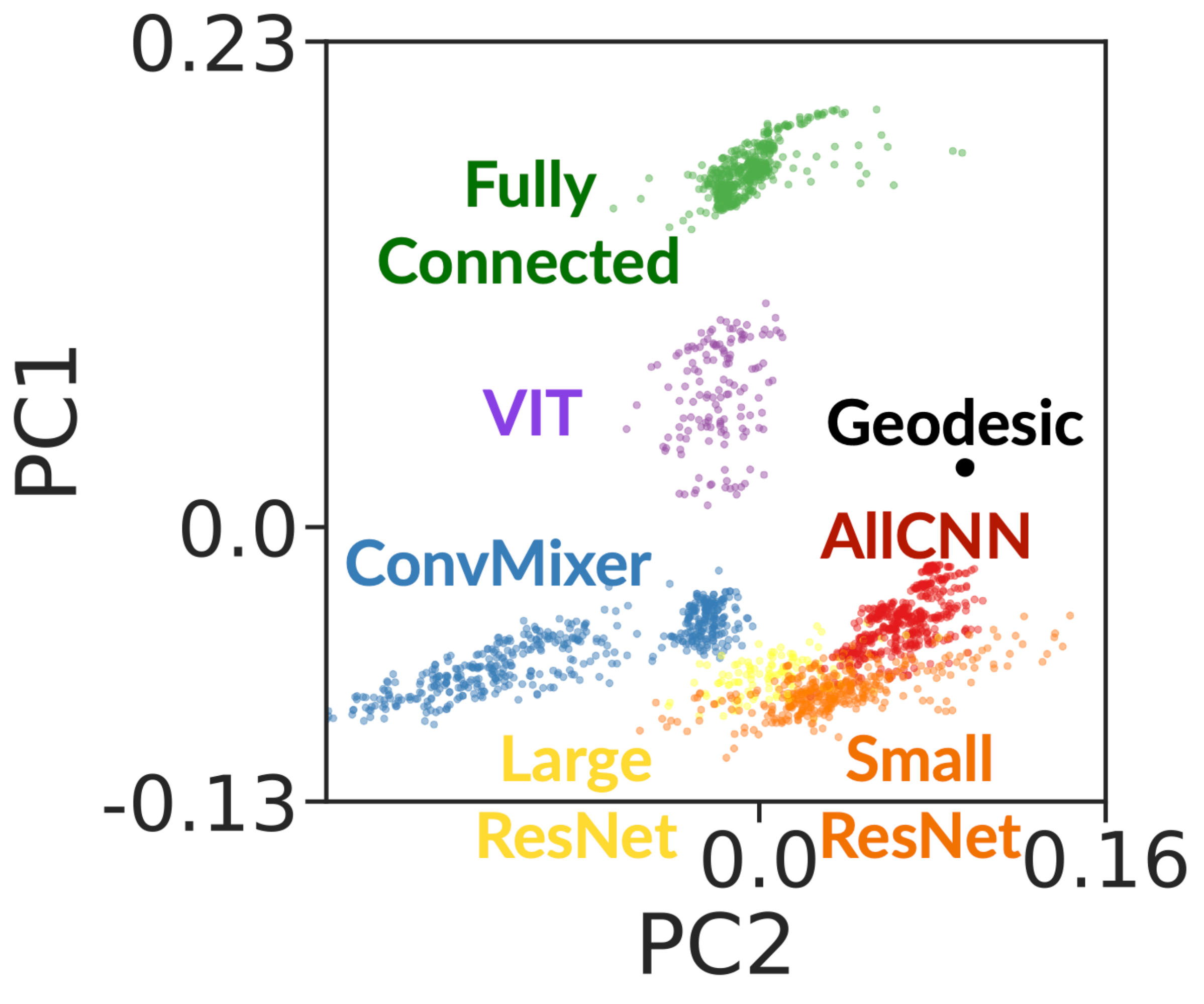}
\caption{}
\label{fig:train_trajectories_inpca}
\end{subfigure}
\begin{subfigure}[b]{0.49\linewidth}
\centering
\includegraphics[width=\linewidth]{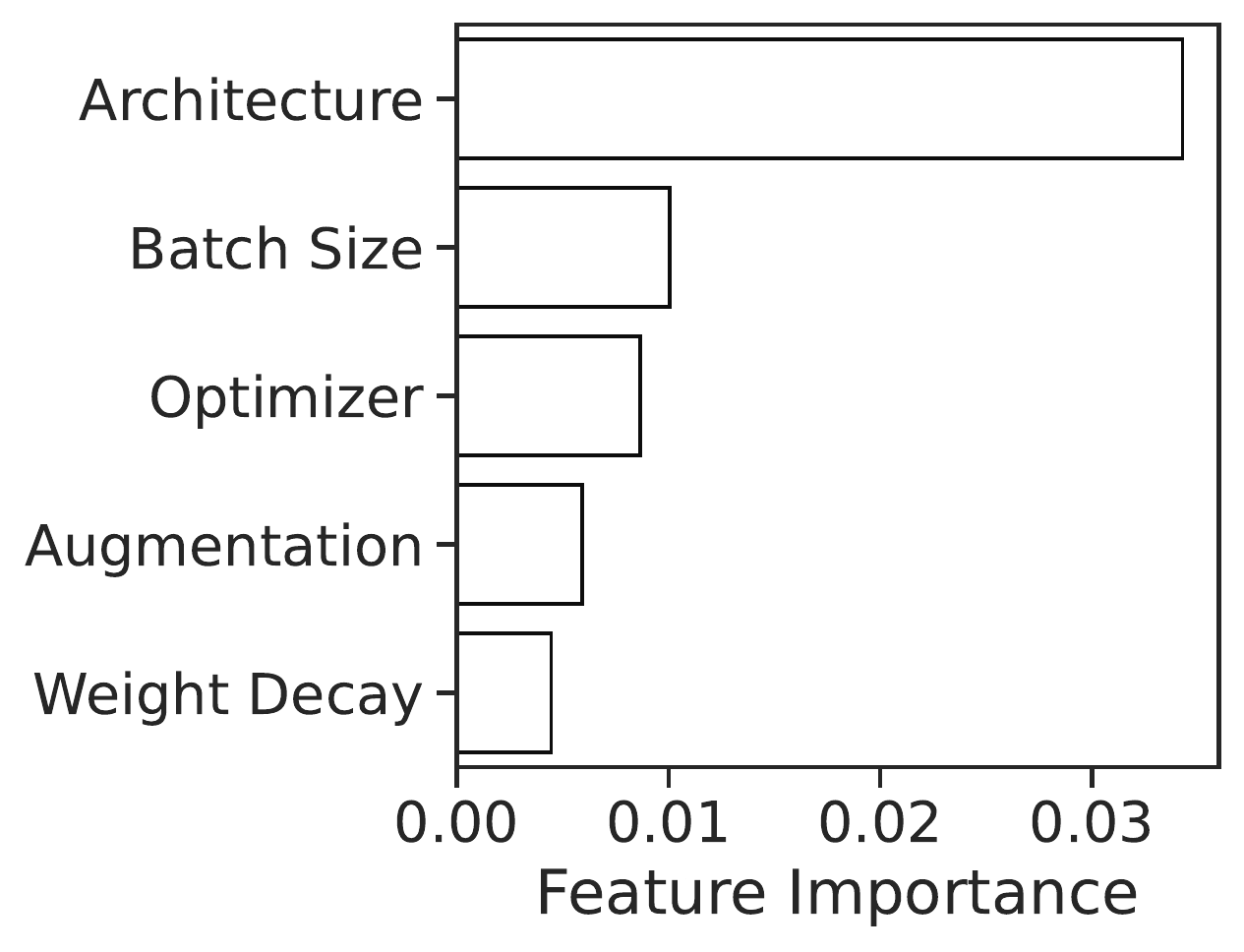}
\caption{}
\label{fig:train_trajectories_feature_importance}
\end{subfigure}
\caption{
\textbf{(a)}: dendrogram obtained from hierarchical clustering of pairwise distances (averaged over weight initializations) between training trajectories (computed using~\cref{eq:dtraj}) of networks with different configurations (X-labels correspond to architecture, optimization algorithm, batch-size, learning rate, weight-decay coefficient and augmentation strategy). There are strong similarities in how networks with different architectures, optimization algorithms and regularization mechanisms learn. \textbf{(b)}: the first two components of an InPCA embedding (without averaging over weight initializations) of train trajectories, each point is one trajectory; explained stress of top two dimensions is 63.6\%. \textbf{(c)}: variable importance from a permutation test ($p < 10^{-6}$) using a random forest to predict pairwise distances. These three plots suggest that architecture is the primary distinguishing factor of trajectories in the prediction space. Test trajectories exhibit similar patterns (see~\cref{fig:result2_test}).
}
\label{fig:result2_train}
\end{figure}

For all networks that trained to zero error, we interpolated the checkpoints from their trajectories to get models along the training trajectory that are equidistant in terms of their progress (\cref{eq:tw}) towards the truth $P_*$. Using these interpolations, we calculated the distance between trajectories corresponding to different configurations using~\cref{eq:dtraj}, averaged over the weight initializations. \cref{fig:dendrogram_train_end} shows a dendrogram obtained from a hierarchical clustering of these distances. Clusters identified from this analysis primarily correspond to different architectures (row colors match those in~\cref{fig:all_models_train_3d,fig:all_models_test_3d}). The cluster of trajectories of networks with convolutional architectures has a diameter that is about as large as the cluster of trajectories of fully-connected and self-attention-based networks (about 0.1 pairwise Bhattacharyya distance on average between models on these trajectories that have the same progress). This points to a strong similarity in how networks with different architectures, optimization algorithms, hyper-parameters, regularization and data augmentation techniques learn. Fully-connected and self-attention-based networks train along different trajectories than networks with convolutional architectures. The geodesic is far from all trajectories.

Within a cluster, say fully-connected networks (green), there are only marginal differences between different configurations, e.g., different optimization methods, different batch-sizes, weight-decay vs.\@ no weight decay, augmentation vs.\@ no augmentation. The dendrogram is created using distances between entire trajectories. So this analysis suggests that training trajectories of most fully-connected networks are similar. This pattern largely holds for the other architectures also. Small vs.\@ large residual networks (orange vs.\@ yellow respectively) have similar training trajectories; \cref{fig:small_vs_large_wrn} shows that the larger network progresses faster towards $P_*$.

Optimization (i.e., the algorithm and the batch-size) is the second prominent distinguishing factor. Within clusters of different architectures, networks trained with the same optimization algorithm have similar trajectories. In particular, for convolutional architectures, trajectories of Adam are more similar to each other than those of SGD or SGD with Nesterov's acceleration. We do not see such a separation for non-convolutional architectures where different optimization algorithms lead to similar trajectories (for them, differences come from data augmentation techniques). The details of different optimization algorithms matter little, e.g., trajectories of networks trained with different learning rate and batch-sizes are quite similar to each other. In general, networks that use weight-decay and networks that do not use weight-decay have similar trajectories. In general, for all architectures, networks trained with augmentation and without augmentation have only marginally different trajectories in the prediction space.

In~\cref{fig:train_trajectories_inpca}, we computed an InPCA embedding of the pairwise distances between trajectories corresponding to different configurations (without averaging across weight initializations). This gives a qualitative understanding of the dendrogram: clusters of InPCA are consistent with the clusters in the dendrogram. While an InPCA embedding of the pairwise distances between models in~\cref{fig:all_models_train_2d} depicts a low-dimensional manifold, \cref{fig:train_trajectories_inpca} illustrates differences in how different configurations train, in particular architectures. This is also evidence that our techniques can also be used to understand entire trajectories in the prediction space. We built a random forest-based predictor of the distance between trajectories of two configurations using their distance to the geodesic (real-valued covariate) and their configuration (categorical covariate) as inputs. A permutation-test performed using the random forest to estimate variable importance in~\cref{fig:train_trajectories_feature_importance} confirms our discussion above: architecture is the most important distinguishing factor of these trajectories and optimization (batch-size, training algorithm) is the next important factor.

\cref{s:app:analysis_train_trajectories} provides a more detailed analysis of the train trajectories. For all architectures, optimization algorithms and regularization mechanisms, networks with different weight initializations train along very similar trajectories in the prediction space. We quantify this phenomenon using ``tube widths'' which capture the differences between models corresponding to different weight initializations at the same progress. Train trajectories are close to the geodesic at early (because they begin near $P_0$) and late parts (because they end near $P_*$) of the training process. While test trajectories also begin near ignorance $P_0$, their distance to the geodesic is larger, and towards the end of training all test models are quite far from truth. As~\cref{s:app:analysis_test_trajectories,fig:dendrogram_test_end} show, test trajectories exhibit largely consistent patterns.

\subsection*{A larger network trains along a similar manifold as that of a smaller network with a similar architecture but makes more progress towards the truth for the same number of gradient updates}

\begin{figure}
\centering
\includegraphics[width=0.49\linewidth]{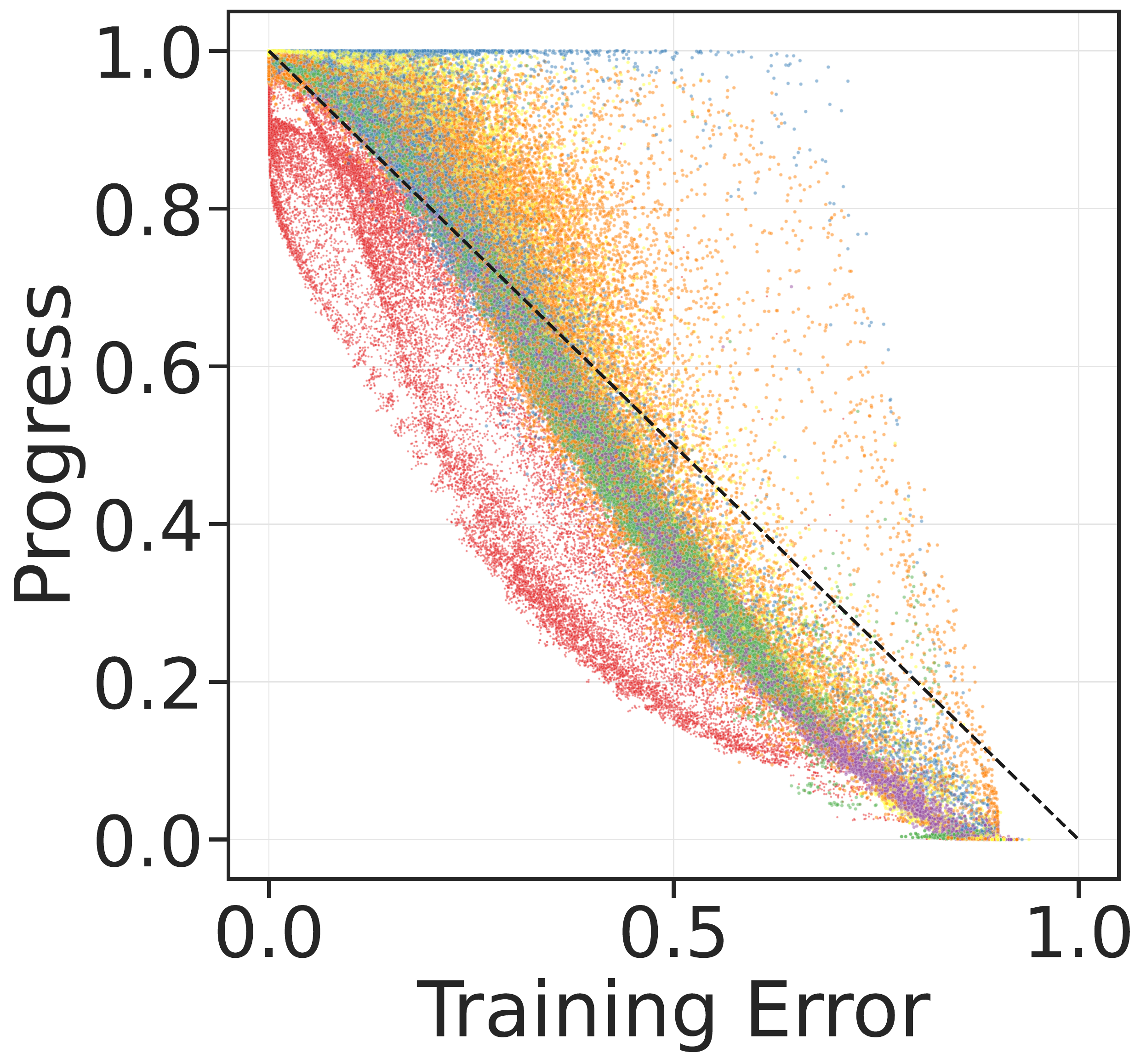}
\includegraphics[width=0.49\linewidth]{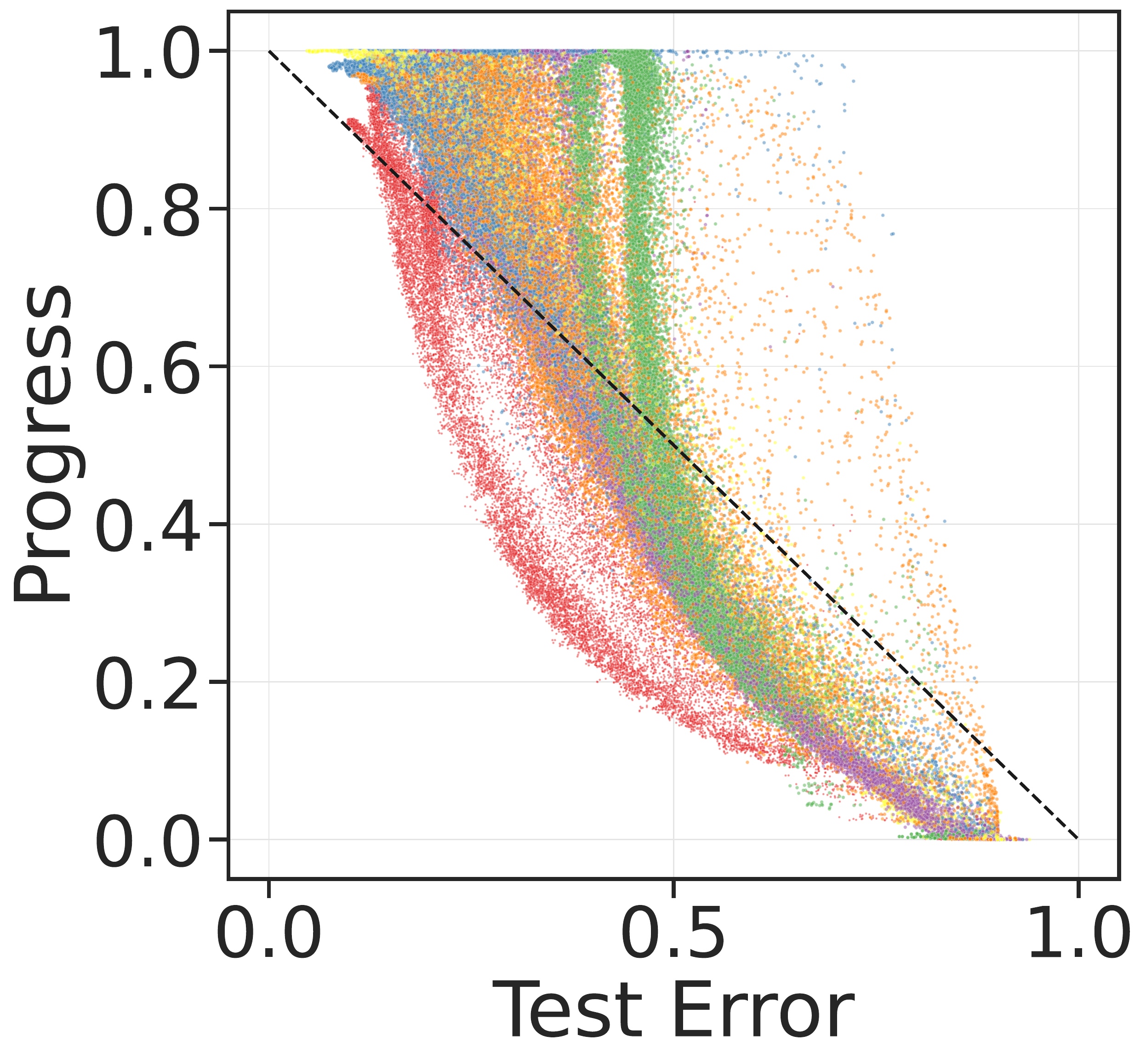}
\caption{Progress of models with different configurations (color scheme is same as that of~\cref{fig:all_models_train_3d}) is strongly correlated with \textbf{(a)} train error ($R^2=0.95$), and \textbf{(b)} test error ($R^2=0.88$).}
\label{fig:progress_vs_error}
\end{figure}

Networks with different configurations make progress towards the truth $P_*$ at different rates. As~\cref{fig:progress_vs_error} shows, progress is strongly correlated with both train error ($R^2=0.95$) and test error ($R^2=0.88$). Progress towards the train truth and towards the test truth are also highly correlated with each other ($R^2 = 0.99$). This suggests that progress, which can be calculated easily using~\cref{eq:tw}, is a good way to judge how close models are to both train and test truths. Note that models may not have a progress of 1 even if they have zero training error (AllCNN trained with Adam in our case). In our work, we have used progress, which is a geometrically natural quantity in probability space, to measure and interpolate trajectories. \cref{fig:progress_vs_error} also suggests that we could have used training error to interpolate checkpoints and would have obtained similar conclusions.

On both train and test manifold, at low error, AllCNN in red and Large ResNet in yellow have markedly different progress than other architectures (too low and too high respectively). Recall from~\cref{fig:all_models_train_3d} and~\cref{fig:all_models_train_d2geod} that trajectories of AllCNNs are also closest to the geodesic and those of Large ResNet are farthest. At high errors, which are typically seen at early training times, all architectures exhibit similar progress. Different weight initializations do not result in different rates of progress. For the same batch-size, SGD with Nesterov's acceleration makes faster progress than SGD or Adam at very early training times but this difference vanishes at later stages of training. In general, models trained with weight decay achieve a lower final progress on both train and test manifolds.

\begin{figure}[htpb]
\centering
\begin{subfigure}[c]{0.49\linewidth}
\includegraphics[width=\linewidth]{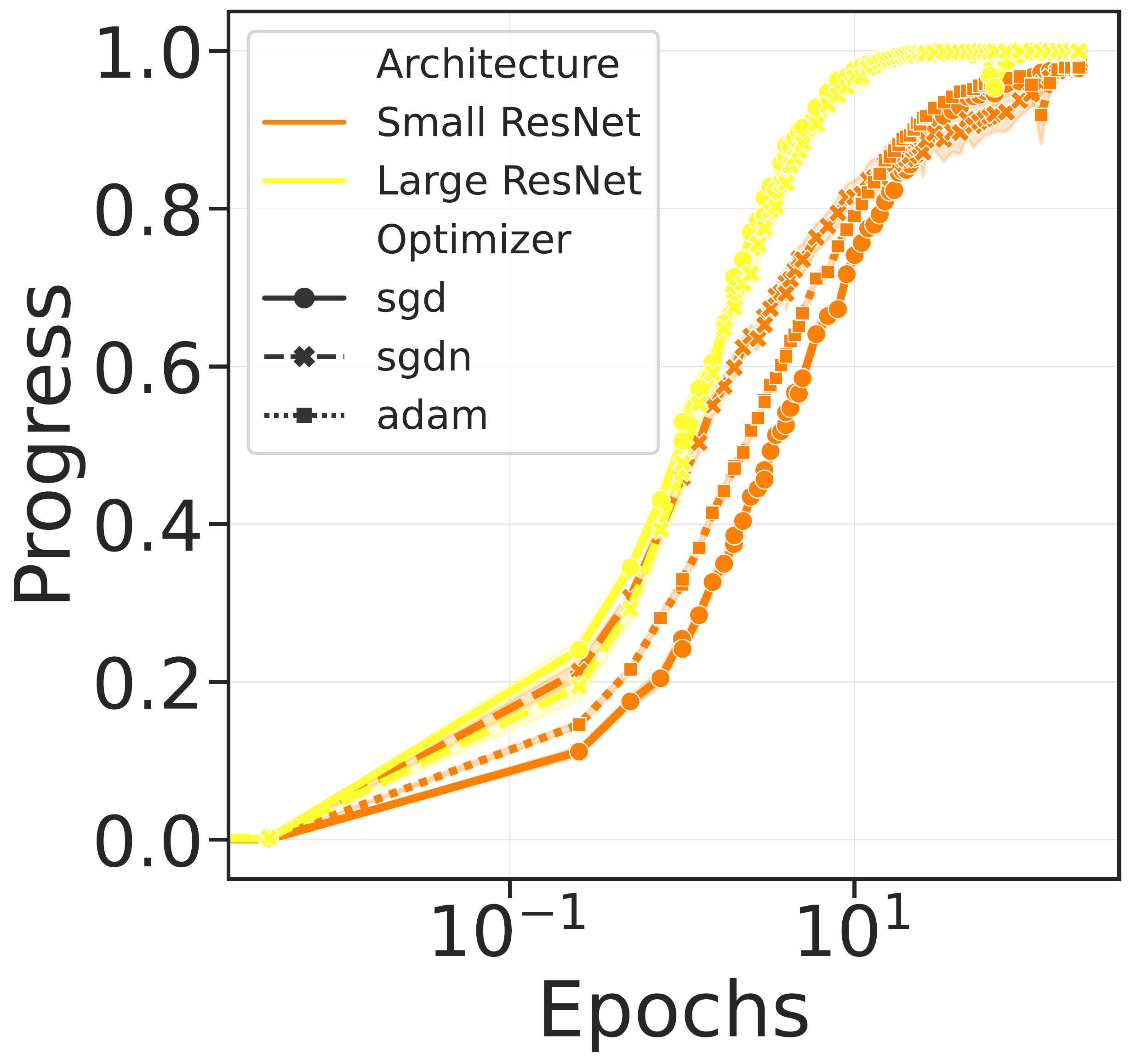}
\caption{}
\label{fig:small_vs_large_wrn_progress}
\end{subfigure}
\begin{subfigure}[c]{0.49\linewidth}
\centering
\includegraphics[width=\linewidth]{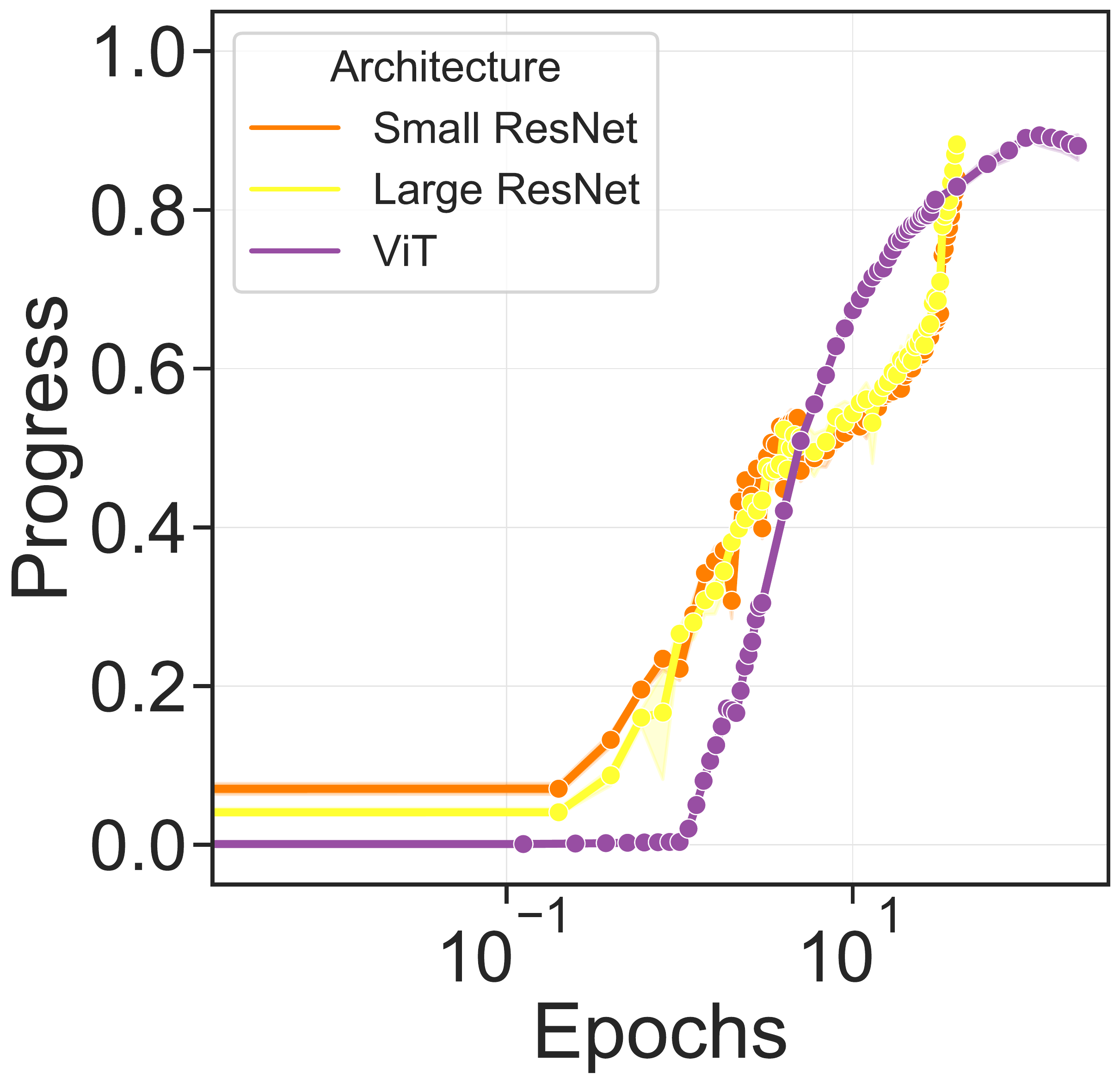}
\caption{}
\label{fig:imagenet_progress_train}
\end{subfigure}
\caption{A Large ResNet makes more progress towards the truth than a Small ResNet for the same number of gradient updates on CIFAR-10 \textbf{(a)} and ImageNet \textbf{(b)}, irrespective of the optimization algorithms. Since the manifold of train and test trajectories for the two architectures are very similar (see~\cref{fig:all_models_train_3d,fig:dendrogram_train_end}), this suggests that larger networks and smaller networks make the same kind of predictions but the larger ones simply learn faster.}
\label{fig:small_vs_large_wrn}
\end{figure}

We saw in~\cref{fig:dendrogram_train_end} that trajectories of the Large ResNet lie on the same sub-manifold as that of the Small ResNet; see~\cref{fig:all_models_tube_width} for the tube widths. The trend for the test manifold in~\cref{fig:dendrogram_test_end} is similar. After the same number of gradient updates, the Large ResNet makes more progress towards the truth than the Small ResNet on CIFAR-10 (\cref{fig:small_vs_large_wrn_progress}). \cref{fig:imagenet_progress_train} shows the training progress against epochs averaged over different weight initializations for models trained on ImageNet. Again, the larger network (ResNet-50) makes more progress compared to the smaller network (ResNet-18) when trained using an identical optimization algorithm, learning rate schedule, batch-size and data augmentation.

%% file: discussion.tex

\section*{Discussion}
\label{s:discussion}

\paragraph{\change{A new insight into optimization in deep learning}}
The central challenge in understanding why we can train deep networks effectively stems from the fact that the likelihood  $p_w(y \mid x)$ of an output $y$ given an input $x$ is a complicated function of the parameters $w$. There is a large body of work that tackles this issue, e.g., optimization and generalization in function spaces for simpler architectures~\cite{Baldi:1989:NNP:70359.70362,liang2020just} or analytical models~\cite{mei2019mean,chizat2020implicit,jacot2018neural}, analyzing representations of different layers~\cite{shwartz-zivOpeningBlackBox2017,achille2018emergence}, properties of stochastic optimization methods~\cite{chaudhari2017stochastic} etc. This has led to some successes, e.g., a characterization of the training dynamics and generalization for two-layer neural networks. But there is a vast diversity of different architectures, optimization methods and regularization mechanisms in deep learning, and it is difficult to draw general conclusions from these analyses.

We have taken a different approach in this experimental paper. We studied many different network configurations to discover surprising phenomena that are not predicted by existing theory. We give two examples here. First, the optimization process explores an effectively low-dimensional manifold in the space of predictions on the train and test data, in spite of the enormous dimensionality of both the embedding space and the weight space. This suggests that the optimization problem in deep learning might have a much smaller computational complexity than what is suggested by existing theory. Second, there is overwhelming empirical evidence that large networks with more parameters generalize better than smaller networks with fewer parameters~\cite{brown2020language,vaswani2017attention,dosovitskiy2021an}. A large body of work has sought to analyze this phenomenon~\cite{belkin2021fit,belkin2019reconciling,bartlett_montanari_rakhlin_2021} and it has also been argued that we need to rethink our understanding of generalization in machine learning~\cite{zhang2016understanding}. We have found that a Large ResNet trains along the same manifold as that of a Small ResNet. \change{It proceeds further towards the truth in the later parts of the trajectory.} In view of the effectiveness of pruning and knowledge distillation~\cite{frankleLotteryTicketHypothesis2019,hinton2015distilling}, this could mean that the superior test error of large networks could be matched by smaller networks using better training methods.

\change{There is some previous work that has argued that weight configurations along a particular training trajectory lie on low-dimensional manifolds, e.g., using PCA~\cite{Feng_Tu_2021}, or by arguing that the mini-batch gradient has a large overlap with the subspace spanned by the top few eigenvectors of the Hessian during training for networks without batch-normalization~\cite{Gur-Ari_Roberts_Dyer_2018,ghorbani19b,Sagun_Bottou_LeCun_2017}. These analyses that study the low-dimensionality of trajectories in the weight space provide important insights into the dynamics of training and foreshadow our work. But their findings are not related to the ones we discussed here. To wit, weights of different architectures lie in totally different vector spaces. We also checked that weights along trajectories of the same network configuration but different weight initialization cannot be explained using few principal components, i.e., they do not lie in a low-dimensional linear subspace, and in fact the explained variance of the top few dimensions decreases proportionally with the number of distinct weight initializations. The mapping between the weight space and the prediction space is quite complicated, and phenomena that occur in the former do not imply that they occur in the latter space in general. Even if the set of models explored by the training process were to lie in a low-dimensional linear subspace, the set of predictions of these models need not lie in a low-dimensional linear subspace. This is because the singular vectors of the Jacobian between the prediction space and the weight space can rotate. Conversely, if the predictions of a set of models lie on low-dimensional manifolds, this does not imply that weights do so as well, because, for instance, there are symmetries in the the parameterization of deep networks.
}


\paragraph{Computational Information Geometry}
Information Geometry~\cite{amariInformationGeometryIts2016} is a rich body of sophisticated ideas, but it has been difficult to wield it computationally, especially for high-dimensional probabilistic models like deep networks. The construction in~\cref{eq:def:Pw} is a finite-dimensional probability distribution, in contrast to the standard object in information geometry which is an infinite-dimensional probability distribution defined over the entire domain of input data. It is this construction fundamentally that enables us to perform complicated computations such as, embeddings of high-dimensional models, geodesics in these spaces, projections of a model onto the geodesic, distances between trajectories in the prediction space, etc. Analysis of high-dimensional probabilistic models is challenging due to the curse of dimensionality: most points are orthogonal to each other in such spaces~\cite{antognini2018pca}. Our visualization techniques, that build upon InPCA and IsKL~\cite{quinn2019visualizing,teohVisualizingProbabilisticModels2020}, work around this issue using multi-dimensional scaling~\cite{cox2008multidimensional,saxe2019mathematical} and distances between probability distributions that violate the triangle inequality, e.g., the Bhattacharya distance. This has some mysterious benefits, e.g., our visualization technique can distinguish between small differences in high-dimensional probability distributions as they approach the truth in Minkowski space~\cite{laub2004feature}. Together with these visualization techniques, the theory developed in this paper gives new tools for the analysis of high-dimensional probabilistic models.

\paragraph{Interpretation of the top three principal coordinates}
It is surprising that just three-dimensions can capture 76\% of the stress (for CIFAR-10) of such a large set of diverse training trajectories in~\cref{fig:all_models_train_3d}. We next offer an interpretation of this phenomenon. Our probabilistic models are an $N$-product of probability distributions corresponding to points $(\sqrt{p_u^n(1)},\dots,\sqrt{p_u^n(C)})$ which lie on a $(C-1)$-dimensional sphere. Training trajectories begin near ignorance $P_0$ and end near $P_*$, so let us consider the straight line that joins ignorance and truth as one basis. Tangents to a training trajectory at ignorance (e.g., when networks are presumably learning ``easy'' images) and at truth (e.g., when networks are learning the most challenging images) can be two more basis vectors. This defines a three-dimensional subspace of the 450,000-dimensional prediction space. To represent this three-dimensional space, we can choose four probability distributions: $P_0$, $P_*$, and $P_{s_1}, P_{s_2}$ computed by weighted averages of models with progress close to $s_1$ and $s_2$, respectively. The latter two are stand-ins for the tangents to the trajectories at $P_0$ and $P_*$ and they are calculated using
\beq{
    P_s = \frac{1}{Z} \sum_{P'} \exp \rbr{ \f{-(s_{P'}-s)^2}{2\s^2}} P',
    \label{eq:P_s}
}
where $Z = \sum_{P'} \exp \rbr{-(s_{P'}-s)^2/(2\s^2)}$ is the normalizing factor and $s_P'$ is the progress of the model $P'$. We choose $\sigma = 0.05$ for all the experiments and experiment with different choices of $s_1$ and $s_2$. We can now build an InPCA embedding using these 4 models, and using the procedure in~\cref{eq:w_expanded} (which is equivalent to weighted-InPCA discussed in~\cref{s:app:weighted_embedding}) we can add our original models in~\cref{fig:all_models_train_3d} into this new InPCA embedding.

\begin{wrapfigure}{r}{0.5\linewidth}
    \centering
    \includegraphics[width=\linewidth]{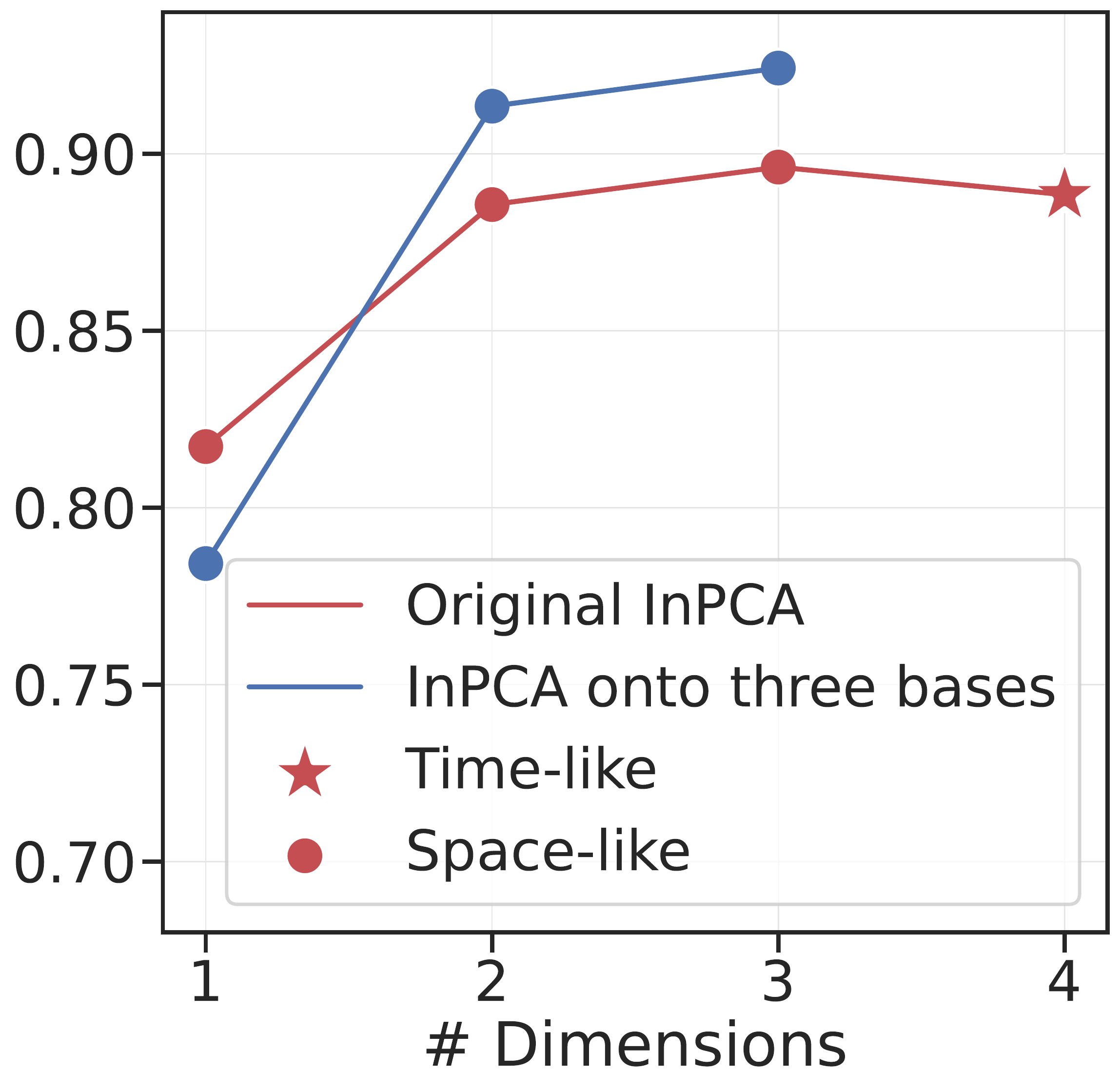}
    \caption{The procedure in~\cref{eq:w_expanded} was used to add original models used for~\cref{fig:all_models_train_3d} into an InPCA embedding created using 4 points corresponding to three ``bases'' (straight line from ignorance to truth, and tangents to the training trajectories at ignorance and truth) for three configurations, all with AllCNN architecture. This new embedding preserves pairwise Bhattacharyya distances between the original models to a similar degree as that of the original InPCA embedding. The two embeddings also assign the same signs to the top few eigenvalues; for the embedding using 4 points, only the first 3 dimensions are non-trivial.}
    \vspace*{-1em}
    \label{fig:allcnn_new_inpca_pairwise_distances}
\end{wrapfigure}

\cref{fig:allcnn_new_inpca_pairwise_distances} shows how well these new coordinates explain pairwise Bhattacharyya distances in $D \in \reals^{m \times m}$ for models of three configurations (AllCNN architectures trained with SGD, SGD with Nesterov's acceleration and Adam) for ten different weight initializations by calculating
\beq{
    \textstyle 1 - \f{\sum_{ij} \abr{D_{ij} - \norm{X_i -X_j}^2}}{\sum_{ij} D_{ij}}
    \label{eq:explained_pairwise_distances}
}
where $X_i \in \reals^{q,d-q}$ are the $d$-dimensional coordinates of the embedded points; we can calculate this quantity that we call ``explained pairwise distances'' using both these new and the original InPCA coordinates. Explained pairwise distances using the original InPCA embedding (which was created using all models) and this new InPCA embedding (which was created using only the 4 points: $P_0, P_*$ and $P_{s_1}, P_{s_2}$ for $s_1 = 1-s_2 = 0.3$) are both quite large---and similar to each other. The two embeddings are also consistent as to which coordinates are time-like (dimensions in~\cref{fig:allcnn_new_inpca_pairwise_distances} are ordered by the magnitude of eigenvalues).

\begin{figure*}
\centering
\begin{subfigure}[b]{0.3\linewidth}
\includegraphics[width=\linewidth]{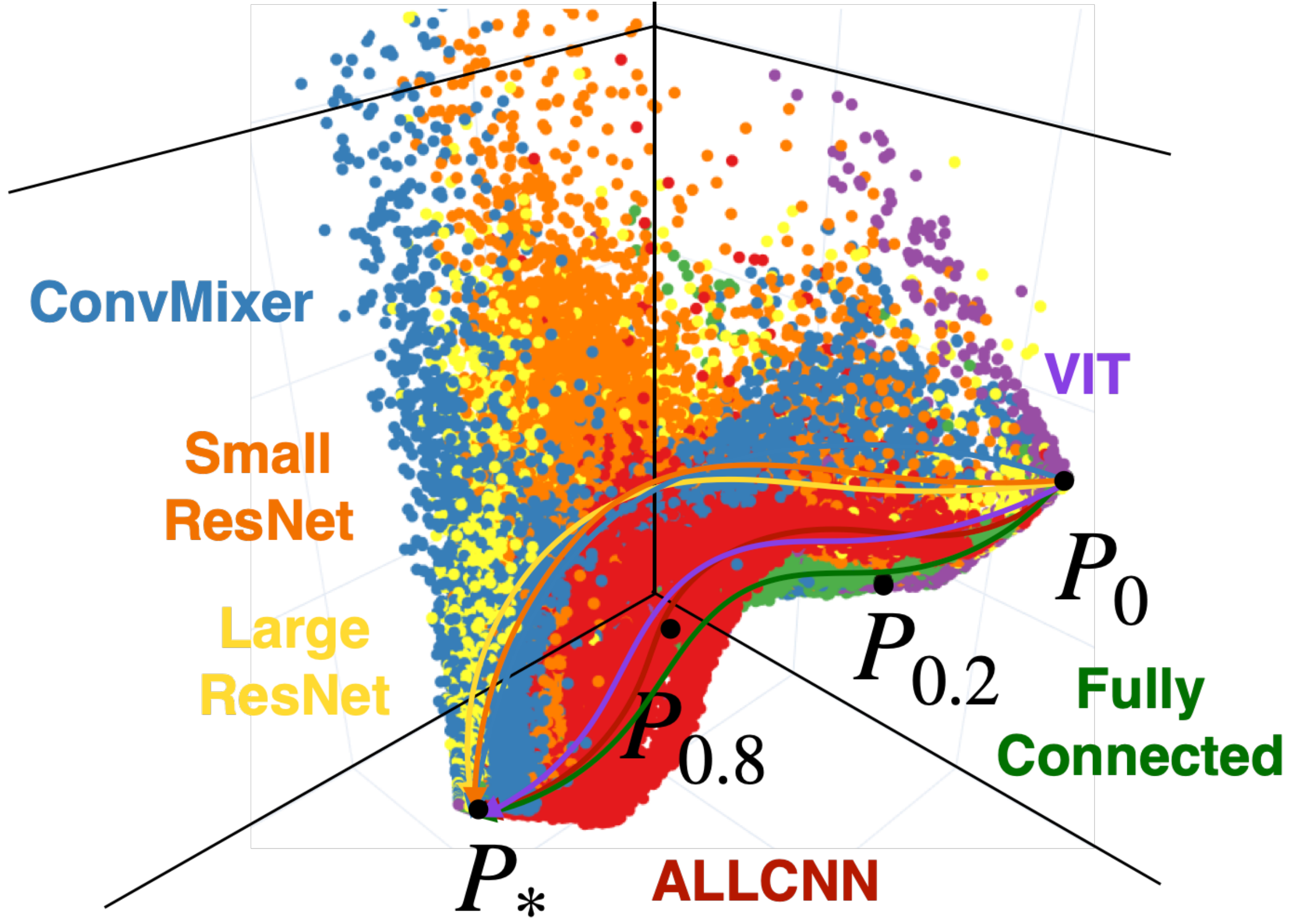}
\caption{InPCA using 4 points}
\label{fig:all_models_new_inpca_3d}
\end{subfigure}
\hspace*{2ex}
\begin{subfigure}[b]{0.32\linewidth}
\includegraphics[width=\linewidth]{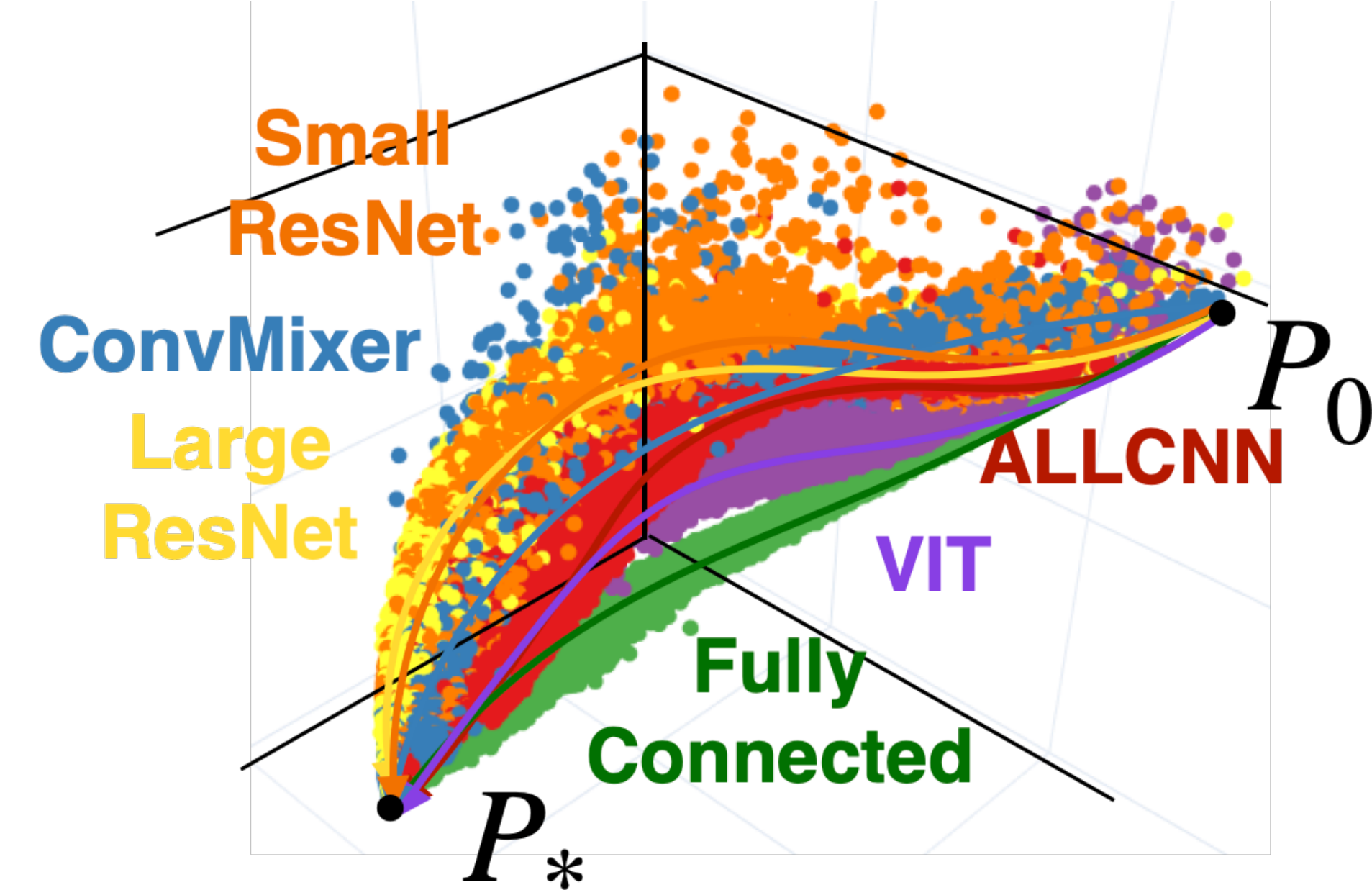}
\caption{Original InPCA}
\label{fig:all_models_no_outliers_inpca_3d}
\end{subfigure}
\hspace*{2ex}
\begin{subfigure}[b]{0.22\linewidth}
\includegraphics[width=\linewidth]{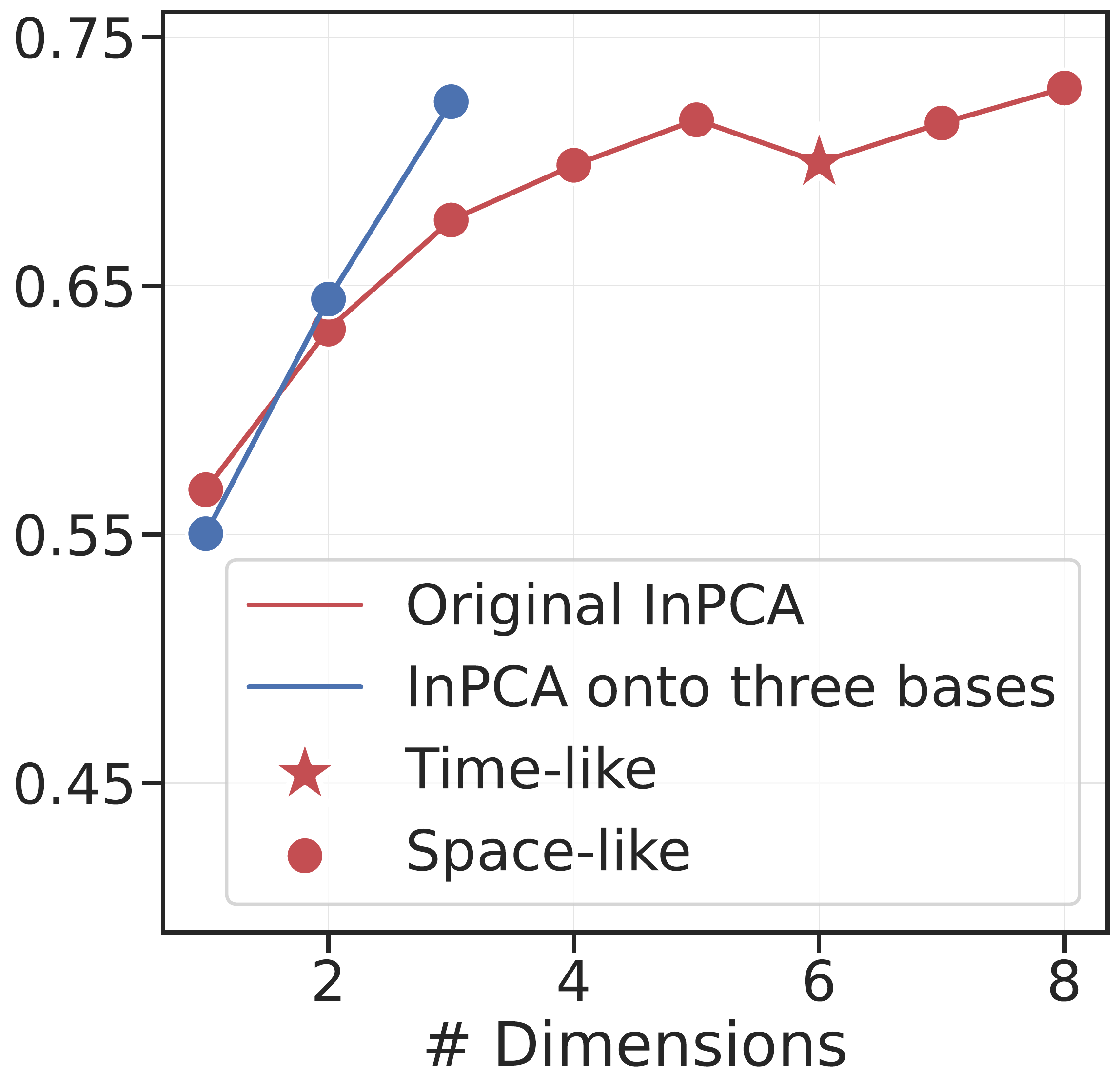}
\caption{Explained Pairwise Distances}
\label{fig:all_models_new_inpca_pairwise_distances}
\end{subfigure}
\caption{
All models in~\cref{fig:all_models_train_3d} with Bhattacharya distance $\dB(P, P_*) < 2$, which effectively removed the spread of points away from the train manifold (also see~\cref{fig:all_models_train_2d_ps}), were embedded using InPCA coordinates constructed using 4 points corresponding to three ``bases'' (straight line from the ignorance to truth, and tangents to the training trajectories at ignorance and truth) in \textbf{(a)} and using the original InPCA coordinates in~\cref{fig:all_models_train_3d} computed using all models in \textbf{(b)}. The top three coordinates in both \textbf{(a)} and \textbf{(b)} are space-like. The manifold in \textbf{(a)} is structurally similar to that of \textbf{(b)}, e.g., Small and Large ResNet models are close to those of ConvMixer models, and far from fully-connected models, some ResNets and ConvMixer models are away from the main manifold at intermediate training times. \textbf{(c)} shows that the explained pairwise Bhattacharyya distances between models in the new embedding is very high, and comparable to that of the first 8 dimensions in the original InPCA. We have drawn smooth curves denoting trajectories by hand to guide the reader.
}
\vspace*{-1em}
\label{fig:all_models_new_inpca}
\end{figure*}
We next performed the same analysis but with all models in~\cref{fig:all_models_train_3d} with $\dB(P, P_*) < 2$, which effectively removes models that lie away from the manifold. In~\cref{fig:all_models_new_inpca_3d}, we created an InPCA embedding using 4 points: ignorance $P_0$, truth $P_*$ and $P_{s_1}, P_{s_2}$ for $s_1 = 1-s_2 = 0.2$ by computing the average over all models $P'$ in~\cref{eq:P_s}, and projected the original probabilistic models into these new coordinates using the procedure in~\cref{eq:w_expanded} to visualize them. We rotated the top 3 non-trivial dimensions of this embedding to best align the embedding created using the original InPCA procedure that uses all models to compute the embedding. This alignment was done using the Kabsh-Umeyama algorithm~\cite{lawrence2019purely} which finds the optimal translation, rotation and sign-flips of the coordinates to align two sets of points; the root mean square deviation (RMSD) is 0.06. As~\cref{fig:all_models_no_outliers_inpca_3d} shows, there are structural similarities in the embedding computed using only the 4 points and the one computed using all models, e.g., Small and Large ResNet models are close to those of ConvMixer models, and far from fully-connected models, some ResNets and ConvMixer models are away from the main manifold at intermediate training times. \cref{fig:all_models_new_inpca_pairwise_distances} shows that the new embedding also preserves pairwise Bhattacharyya distances between the models to a similar degree.

This exercise gives us an interpretation for the low-dimensional embedding discovered by InPCA. It may point to a mechanistic explanation for our findings: the train and test manifolds are effectively low-dimensional because networks with different architectures, optimization algorithms, hyper-parameter settings and regularization mechanisms fit the same easy images in the dataset first and the same challenging images towards the end of training; this phenomenon has also been studied in~\cite{hacohen2020let}.

\paragraph{Why are the train and test manifolds effectively low-dimensional?}
It is remarkable that trajectories of networks with such different configurations lie on a manifold whose dimensionality is much smaller than the embedding dimension. To explore this further, we analyzed trajectories of networks trained on synthetic data: (a) sampled from a ``sloppy'' Gaussian, i.e., with eigenvalues of the covariance that are distributed uniformly on a logarithmic scale (this structure has been noticed in many typical problems~\cite{yang2021does,quinnInformationGeometryMultiparameter2021}), and (b) sampled from an isotropic Gaussian (non-sloppy data). We labeled these samples using a random two-layer fully-connected teacher network and trained student networks with different configurations to fit these labels. When students are initialized near ignorance $P_0$, train and test manifolds are effectively low-dimensional for both kinds of data (87\% explained stress in top ten dimensions). When students are initialized at different initial points $\{P_0^{(k)}\}_{k=1,\dots,10}$ similar to those in~\cref{fig:allcnn_from_corners}, train and test manifolds are still effectively low-dimensional for both kinds of data; top ten dimensions have 85\% explained stress. But the explained stress is higher in the top few dimensions if trajectories begin from near each other, e.g., from fewer initial points, or from ignorance. For sloppy input data, trajectories converge to the same manifold quickly even if they begin from very different initial points. \cref{s:app:details_synthetic} discusses this experiment further.

We therefore believe that the low-dimensionality of the manifold arises from (a) the structure of typical datasets~\cite{goldt2020modeling,d2021interplay,refinetti2021classifying}, e.g., spectral properties, and (b) the fact that typical training procedures initialize models near one specific point in the prediction space, the ignorance $P_0$. Along the first direction, recent work on understanding generalization~\cite{bartlett2020benign,yang2021does} has argued that deep networks, as also linear/kernel models, can interpolate without overfitting if input data have a sloppy spectrum. Work in neuroscience~\cite{simoncelliNaturalImageStatistics2001,fieldWhatGoalSensory1994} has also argued for visual data being effectively low-dimensional. Theories in machine learning~\cite{vapnik1998statistical,scholkopf2002learning} and information-theory~\cite{rissanen1978modeling,balasubramanianStatisticalInferenceOccam1997} for model selection are based on estimates of the number of models in a hypothesis class that are consistent with the data. In this context, our second suspect, namely initialization, suggests that even if the size of the hypothesis space might be very large for deep networks~\cite{dziugaiteComputingNonvacuousGeneralization2017,bartlett2017spectrally}, the subset of the hypothesis space explored by typical training algorithms might be much smaller.


%% file: ack.tex

\section*{Acknowledgments}
\label{s:ack}
JM, RR, RY and PC were supported by grants from the National Science Foundation (IIS-2145164, CCF-2212519), the Office of Naval Research (N00014-22-1-2255), and cloud computing credits from Amazon Web Services. IG was supported by the National Science Foundation (DMREF-89228, EFRI-1935252) and Eric and Wendy Schmidt AI in Science Postdoctoral Fellowship. HKT was supported by the National Institutes of Health (1R01NS116595-01). JPS was supported by the National Science Foundation (DMR-1719490), MKT was supported by the National Science Foundation (DMR-1753357). The authors would like to acknowledge Itai Cohen and Jay Spendlove for helpful comments on this material and manuscript.

%% file: appendix.tex

\onecolumn

\renewcommand\thesection{S.\arabic{section}}
\renewcommand\thefigure{S.\arabic{figure}}
\renewcommand\thetable{S.\arabic{table}}
\setcounter{figure}{0}
\setcounter{section}{0}
\setcounter{table}{0}
\renewcommand{\thesubsection}{\thesection.\arabic{subsection}}

\begin{appendix}

\section{Notation}
\label{s:app:notation}

\begin{center}
\resizebox{0.8\linewidth}{!}{
\renewcommand{\arraystretch}{1.25}
\begin{tabular}{c m{7cm}}
    \toprule
    \textbf{Symbol} & \textbf{Description}\\
    \midrule
    $N$                & Number of samples    \\
    $C$                & Number of classes    \\
    $x_n$              & Input sample with index $n \in \{1,\dots,N\}$ \\
    $y_n$              & Label assignment of sample with index $n \in \{1,\dots,N\}$ \\
    $y_n^*$              & Ground-truth label of sample with index $n \in \{1,\dots,N\}$ \\
    $w$                & Weights of the deep network \\
    $\vec y^*$           & Ground-truth labels for each of the $N$ samples, $\vec y^* = (y_1^*,\dots,y_N^*)$\\
    $\vec y$           & Label assignment for each of the $N$ samples, $\vec y \in \cbr{1, \dots, C}^N$ \\
    \midrule
    $p_w^n(y_n)$  & Probability that sample $x_n$ belongs to class $y_n \in \cbr{1,\dots,C}$, $p_w^n(y_n) \equiv p_w(y_n \mid x_n)$ \\
    $P_w(.)$           & Probabilistic model with weight $w$; assigns a probability to every sequence $\vec y$  \\
    $P_*$               & Truth ($P_* = \delta_{\vec y^*}(\vec y)$)\\
    $P_0$              & Ignorance, has $p_0^n(c) = 1/C$ for all classes $c$ and samples $n$\\
    $\dB$              & Bhattacharyya distance between two probability distributions  \\
    $\dG$              & Geodesic distance (great circle distance) between two probability distributions  \\
    \midrule
    $g(w)$             & Fisher Information Metric (FIM) at weight configuration $w$\\
    $(\sqrt{p_u^n(c)})_{c=1,\dots, C}$  &  Point on a $(C-1)$-dimensional sphere \\
    $P_{u,v}^{\a}$ & Geodesic between probability distributions $P_u$ and $P_v$ parameterized by $\alpha \in [0, 1]$ \\
    \midrule
    $T$                & Number of recorded checkpoints \\
    $(w(k))_{k=0, \cdots T}$  & A sequence of recorded checkpoints in the weight space\\
    $s_w$      & Progress of a probabilistic model $P_w$ with weights $w$\\
    $\a$ &      Interpolating parameter along a geodesic, $\a \in [0,1]$\\
    $\tt_w$ & A sequence of probabilistic models recorded during training, also denoted by $(P_{w(k)})_{k=0, \cdots T}$  \\
    $\t_w$ & A continuous curve in the space of probabilistic models, also denoted by $(P_{w(s)})_{s \in [0,1]}$\\
    $\dtraj(\t_u, \t_v)$  & Distance between trajectories $\t_u$ and $\t_v$  \\
    \midrule
    $D$                & Matrix ($\in \reals^{m \times m}$) of pairwise Bhattacharyya distances between $m$ probabilistic models, entries of this matrix are denoted by $D_{ij}, D_{uv}$ etc. depending
    upon the context\\
    $W$                & Matrix ($\in \reals^{m \times m}$) of centered pairwise Bhattacharyya distances, $W = -LDL/2$ where $L_{uv} = \delta_{uv} - 1/m$ performs the centering\\
    $X_w$                & Coordinates ($\in \reals^{p,m-p}$) of the InPCA embedding of a model with weights $w$\\
    $1 - \sqrt{\f{\sum_{ij} \rbr{W_{ij} - \sum_{k=1}^d \L_{k k} U_{i k} U_{k j}}^2}{\sum_{ij} W_{ij}^2}}$
    & Explained stress, used to estimate the fraction of the entries of the centered pairwise distance matrix $W$ that are preserved by an embedding; equivalent to explained variances in standard PCA (up to the square root)\\
    $1 - \f{\sum_{ij} \abr{D_{ij} - \norm{X_i -X_j}^2}}{\sum_{ij} D_{ij}}$
    & Explained pairwise distances, used to estimate the fraction of the entries of the pairwise Bhattacharyya distance matrix $D$ that are preserved by an embedding\\
    \bottomrule
\end{tabular}
}
\end{center}

\section{Derivation of the joint probability of predictions and the Bhattacharyya distance}
\label{s:app:derivation}

\change{
The quantity in~\cref{eq:def:Pw} is the joint likelihood of all the labels given the weights. Observe that
\[
    \aed{
    P_w(\vec y) &\equiv p(\cbr{(x_n,y_n)}_{n=1}^N; w)\\
    &= p(x_1,\ldots,x_N)\ p_w(y_1,\dots,y_N \mid x_1, \dots, x_N)\\
    &\overset{\text{(a)}}{=} p(x_1,\ldots,x_N) \prod_{n=1}^N p_w(y_n \mid x_1, \dots, x_N)\\
    &\overset{\text{(b)}}{=}  p(x_1,\ldots,x_N) \prod_{n=1}^N p_w(y_n \mid x_n)\\
    &\overset{\text{(c)}}{=} \rbr{\f{1}{N} \sum_{n=1}^{N} \delta_{x_n}(x_n)} \prod_{n=1}^N p_w(y_n \mid x_n)\\
    &= \prod_{n=1}^N p_w(y_n \mid x_n).
    }
\]
In this calculation, we have used the assumption that (a) predictions on two samples are independent of each other \emph{given the weights and the input samples} (if we marginalize on the weights, they are certainly dependent), (b) we are performing inductive inference, i.e., $p(y_n \mid x_1,\dots,x_n) = p(y_n \mid x_n)$, and (c) the samples are frozen to the ones in the training set for the analysis, i.e., the distribution $p(x_1,\dots,x_n) \equiv \rbr{\f{1}{N} \sum_{n=1}^{N} \delta_{x_n}(x_n)} = 1$. So we actually do not need to use the assumption that the training samples $x_1, \dots, x_N$ are independent of each other to write down the joint likelihood that factorizes over the samples in the training set. Certainly, if the training samples are independent, then this derivation also holds. Let us note that training samples being independent of each other is one of the most common assumptions in machine learning. This assumption is used to derive, for instance, the maximum likelihood estimator in~\cite[Equation 1.61]{bishop1995neural}.

The expression for the Bhattacharyya distance between two probability distributions $P_u$ and $P_v$ in~\cref{eq:dB} can be derived as follows. Note that $\yvec$ can take a total of $C^N$ distinct values, and each $y_n \in \{1,\dots, C\}$.
\[
    \aed{
        \dB(P_u, P_v)
        &\doteq -\f{1}{N} \log \sum_{\yvec} \sqrt{P_u(\yvec)} \sqrt{P_v(\yvec)}\\
        &= -\f{1}{N} \log \sum_{\yvec} \prod_{n=1}^N \sqrt{p^n_u(y_n)} \sqrt{p^n_v(y_n)}\\
        &= -\f{1}{N} \log \sum_{y_1} \dots \sum_{y_{N-1}} \prod_{n=1}^{N-1} \sqrt{p^n_u(y_n)} \sqrt{p^n_v(y_n)} \rbr{ \sum_{y_N} \sqrt{p^N_u(y_n)} \sqrt{p^N_v(y_n)}}\\
        &\vdots\\
        &= -\f{1}{N} \log \prod_{n=1}^N \sum_c \sqrt{p^n_u(c)} \sqrt{p^n_v(c)}\\
        &= -\f{1}{N} \sum_n \log \sum_c \sqrt{p^n_u(c)} \sqrt{p^n_v(c)}.
    }
\]
Calculations like the one above hold in general, the joint entropy of two independent random variables is the sum of their individual entropy. Just like the familiar cross-entropy loss used for training deep networks is an average over the samples, the Bhattacharyya distance is also an average over the training samples.
}

\section{Details of the experimental setup}
\label{s:app:details}

\paragraph{Datasets}
The experimental data in this paper was obtained by training deep networks on two datasets.
\begin{itemize}[nosep]
\item The CIFAR-10 dataset~\cite{krizhevsky2012imagenet} has $N=50,000$ RGB images in the training set of size 32$\times$ 32 from $C=10$ different categories (airplane, automobile, bird, cat, deer, dog, frog, horse, ship and truck). The test set has $N=10,000$ images. Both train and test sets have an equal number of images in each category.
\item The ImageNet dataset~\cite{dengImagenetLargescaleHierarchical2009} has $C=1000$ categories and a total of $N = 1.28 \times 10^6$ RGB images of size 224 $\times$ 224 in the training dataset. Different categories have slightly different numbers of images in the train set, but all categories have at least 1000 images. The test set consists of $N=50,000$ images, with 50 images from each category. 
\end{itemize}

\paragraph{Neural architectures}
For CIFAR-10, we used six neural architectures. These architectures were chosen and and configurations were chosen to ensure that these networks could fit all the images in the training dataset, i.e., achieve zero training error, for most training methods.
\begin{enumerate}[(i),nosep]
\item A multi-layer perceptron with rectified linear unit (ReLU) nonlinearities (fully-connected network) with 4 hidden layers, of size [1024, 512, 256, 128] respectively.
\item An ``all convolutional network'' (AllCNN~\cite{springenbergStrivingSimplicityAll2015}) with 5 convolutional layers followed by an average pooling layer; first three layers have 96 channels and the latter two have 144 channels.
\item Two different wide residual networks~\cite{zagoruyko2016wide}. The larger one has 16 layers and [64, 256, 1024, 4096] channels for the convolutional layers in the four blocks, and the smaller network has 10 layers with [8, 32, 128, 512] channels for the four blocks. Both networks have a ``widening factor'' of 4. We modified the implementation at \href{https://github.com/meliketoy/wide-resnet.pytorch}{https://github.com/meliketoy/wide-resnet.pytorch}.
\item The ConvMixer architecture~\cite{trockman2022patches} is a convolutional network but it uses very large receptive fields and maintains the same size for the activations across successive layers. We did not make any changes to the architecture from the original paper.
\item The ViT architecture~\cite{dosovitskiy2021an} is a self-attention based network that uses a set of disjoint patches of size 4$\times$4 from the input images. This network does not use convolutional operations and instead uses the so-called self-attention layer that is popularly in natural language processing. We use a linear layer size of 512, 8 self-attention heads and 6 transformer blocks (layers). We used the implementation from \href{https://github.com/lucidrains/vit-pytorch}{https://github.com/lucidrains/vit-pytorch}.
\end{enumerate}
We do not use Dropout~\cite{srivastava2014dropout} in any of the networks. All networks except ViT have a batch-normalization~\cite{ioffe2015batch} layer after each convolutional or fully-connected layer, except ViT which uses layer normalization~\cite{baLayerNormalization2016}.

For ImageNet, we used three architectures.
\begin{enumerate}[(i),nosep]
\item A smaller residual network~\cite{he2016deep} with 18 layers (ResNet-18). This residual network is different from the wide residual network used for CIFAR-10, primarily in that there are fewer channels in each block. A ResNet is architecturally similar to a wide residual network with a widening factor of 1. We replaced each strided convolution with a convolution followed by a BlurPool layer~\citep{zhang2019making}.
\item A larger residual network with 50 layers (ResNet-50). This is one of the most popular networks for training on ImageNet and widely used as a benchmark architecture in the field.
\item The ViT architecture which is similar to the one used for CIFAR-10 above except that the receptive field of the first layer is larger due to the larger images in ImageNet. We trained a smaller variant of ViT called ViT-S (with 22 million weights) which was introduced in~\citep{touvron2021training}. It operates on patches of size 16 $\times$ 16 and has 6 self-attention heads and 12 transformer blocks.
\end{enumerate}
Training multiple models on ImageNet is computationally expensive. To mitigate this, we used efficient data loaders, computed gradients in half-precision, and chose effective training hyper-parameters (FFCV~\citep{leclerc2022ffcv} for training ResNets and timm~\citep{rw2019timm} for training ViTs).

\paragraph{Training procedure}
For both datasets, we normalize images in the train and test sets by the channel-wise mean and variance of the images in the  training dataset. For CIFAR-10, we also augmented training images by randomly cropping a region of size 32$\times$ 32 after padding the original image by 4 pixels on each side, and performing horizontal flips with a probability of 0.5; our data contains models trained with and without such data augmentation.

All the networks are initialized using the default PyTorch weight initialization as follows. For fully-connected layers with an input dimension $d$, all weights and biases are sampled independently from a uniform distribution on $[-d^{-1/2}, d^{-1/2}]$. For convolutional layers with $c$ channels and a $k \times k$ convolutional kernel, all weights and biases are sampled independently from a uniform distribution on $[-(c k)^{-1/2}, (c k)^{-1/2}]$.

\begin{wrapfigure}{r}{0.4\linewidth}
    \centering
    \includegraphics[width=\linewidth]{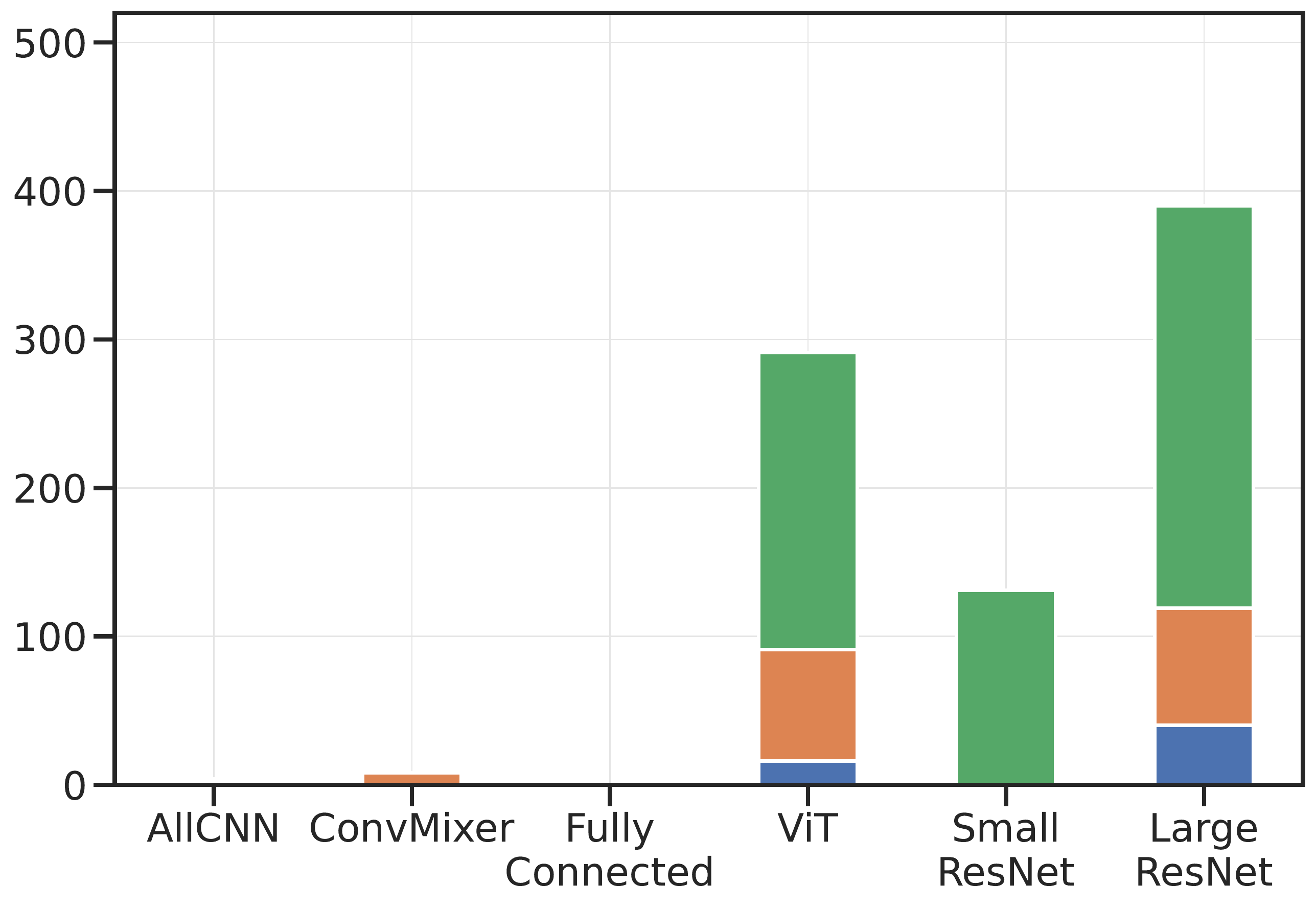}
    \caption{Number of networks that did not train beyond 90\% error for Adam (green), SGD (blue) and SGD with Nesterov's acceleration (orange). These models are not included in our analysis.}
    \label{fig:count_of_not_trained_models}
    \end{wrapfigure}
    We started with 3120 different configurations, 520 for each network architecture. Some networks did not finish training due to numerical errors during gradient updates, and we excluded them from our analysis. \cref{fig:count_of_not_trained_models} shows how many of the configurations did not finish training for each network architecture. Our data, with 2,296 different configurations, therefore contains fewer ViTs and Large ResNets than other architectures.

For CIFAR-10, we used three different optimization methods, stochastic gradient descent (SGD), SGD with Nesterov's momentum (with a coefficient of 0.9) and Adam~\cite{kingma2014adam}, three different batch sizes (200, 500 and 1000) and three different values of the weight decay coefficient ($\ell_2$ regularization) ($\{0, 10^{-3}\}$ when training with SGD and SGD with Nesterov's momentum, and $\{0, 10^{-5}\}$ when training with Adam). Fully-connected networks trained on augmented data are trained for 300 epochs to achieve zero training error, all other networks are trained for 200 epochs. One epoch corresponds to using each sample in the training dataset exactly once to compute a gradient update (i.e., mini-batches are sampled without replacement). As the batch-size in SGD is increased, the stochasticity of the weight updates decreases and this makes the iterations more susceptible to converging near local minima of the loss function, and thereby obtain poor test error. It has been noticed empirically that keeping the ratio of the learning rate to batch-size invariant helps mitigate this deterioration of test error for large batch sizes~\cite{goyal2017accurate}. This has also been argued theoretically via an analysis of the equilibrium distribution of SGD~\cite{chaudhari2017stochastic}. Therefore, for SGD and SGD with Nesterov's acceleration, we fixed this ratio to $5 \times 10^{-4}$, i.e., for a batch size 200, we use a learning rate of 0.1, and increase the learning rate proportionally for larger batch sizes. For Adam, this ratio was $5 \times 10^{-6}$, i.e., we used a learning rate of 0.001 for a batch-size of 200. For all experiments, we decreased the learning rate using a cosine annealing schedule over the course of training, i.e., for all networks the learning rate decays to zero at the end of training.

Residual networks on ImageNet were trained using SGD with Nesterov's acceleration for 40 epochs with a batch-size of 1024. The learning rate was decreased linearly from 0.5. We used a weight decay coefficient of $5 \times 10^{-5}$; no weight decay was applied to parameters associated with batch-normalization. To reduce the training time, we used mixed-precision training. We also used progressive resizing, i.e., we trained on images of size 196 $\times$ 196 for the first 34 epochs before using the full-sized images (224 $\times$ 224) for the remaining 6 epochs. We use random horizontal flips and random-resize-crops for data augmentations. For datasets with a large number of classes such as ImageNet, it helps to use label smoothing~\cite{he2019bag}, we used this with the smoothing parameter set of 0.9. This amounts to training towards a slightly different truth $P_*$ where the correct category has a probability of 0.9 and the remainder 0.1 is distributed uniformly across the other 999 categories (instead of them being zero).

ViT architectures are difficult to train well with SGD, especially on large datasets such as ImageNet. We therefore trained ViTs on ImageNet using AdamP~\citep{heo2021adamp} with a cosine-annealed learning rate schedule and an initial learning rate of 0.001. We trained for 200 epochs using a batch-size of 1024 and weight decay of 0.01 without any dropout. These networks also require a more extensive set of data augmentations, we used horizontal flips with probability 0.5, cropping the image to get a patch of the desired size at a random location (images in ImageNet are not of the same size), and mixup regularization~\citep{zhang2017mixup} which uses mini-batches that consist of convex combinations (with a random parameter) of images and ground-truth probability distributions.

\begin{wrapfigure}{r}{0.4\linewidth}
\centering
\includegraphics[width=\linewidth]{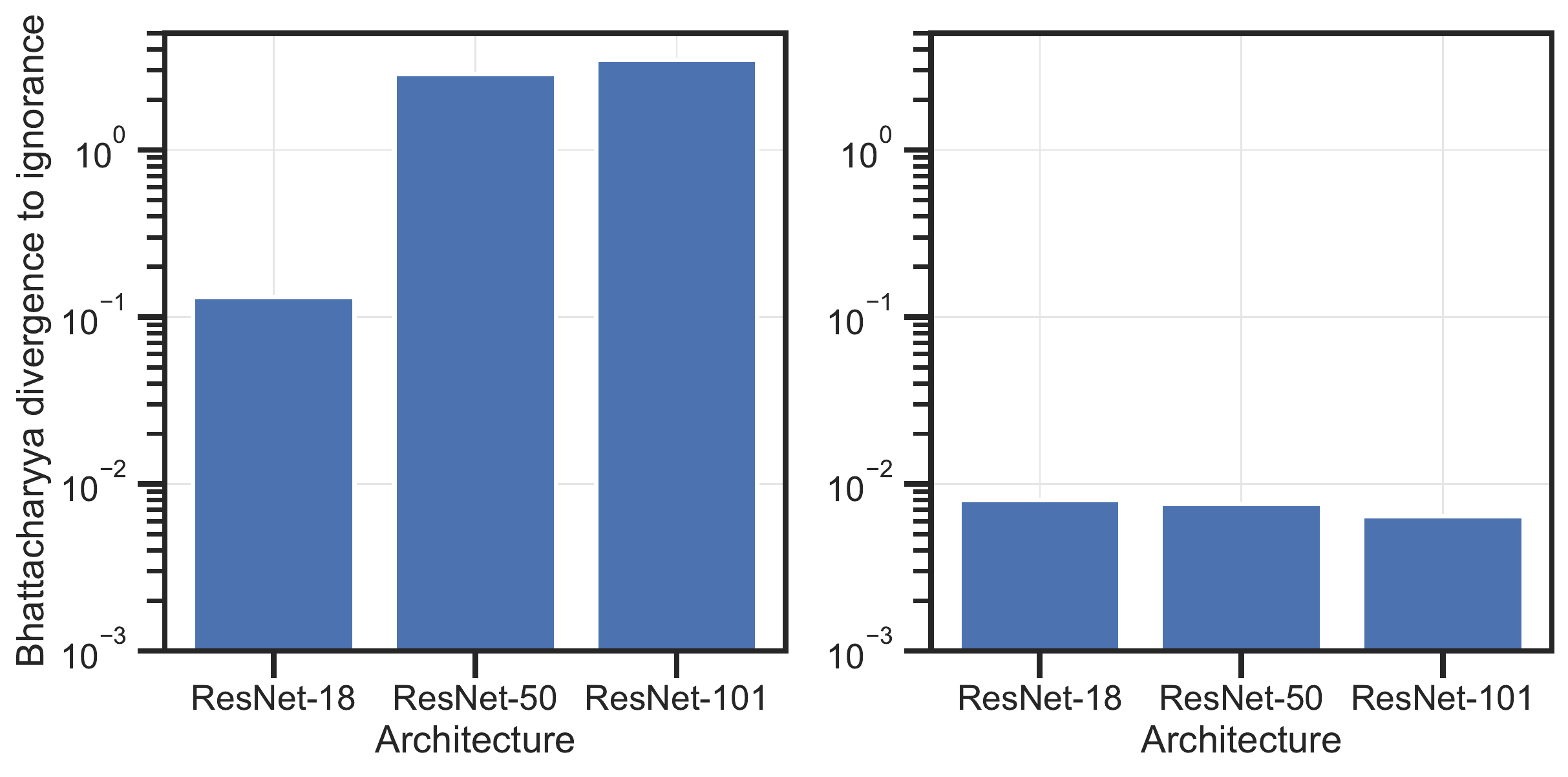}
\caption{Bhattacharyya distance from ignorance $P_0$ for networks at the beginning of training for standard off-the-shelf implementations of ResNet (left). If we initialize the estimates of the mean and standard deviation of the batch-normalization layers by doing a forward pass on a few mini-batches, then networks are close to ignorance at the beginning of training (right).}
\label{fig:imagenet_resnet_ignorance_dist}
\end{wrapfigure}

\paragraph{Some ImageNet models are not initialized near ignorance $P_0$}
We noticed that some randomly initialized models have a large Bhattacharyya distance from ignorance $P_0$. For example, the distance between a randomly initialized ResNet-50 model and ignorance is as much as 0.91 times the Bhattacharyya distance between ignorance and truth $\dB(P_0, P_*)$. We found that this is due to the batch-normalization layer~\citep{ioffe2015batch} being incorrectly initialized at the beginning of training. Batch-normalization subtracts the channel-wise mean of the activations (computed from samples in a mini-batch) and divides the result by an estimate of the channel-wise standard deviation of the activations (computed using the samples in the mini-batch). During training, typical deep learning libraries such as PyTorch maintain an exponentially moving average of the mean and standard deviation of activations of mini-batches. And it is these averaged estimates that are used to compute the output probabilities for test data. In PyTorch, the mean is initialized to zero and the standard deviation is initialized to 1. This causes the magnitude of the activations to be very large in the final few layers at initialization and that is why the probabilistic model is very far from ignorance at initialization, as shown in~\cref{fig:imagenet_resnet_ignorance_dist}.

This phenomenon is seen in most popular off-the-shelf implementations of a ResNet, and could also be present in other architectures. When training in a supervised learning setting, this finding of ours is only marginally relevant because the estimates of the mean and standard deviation stabilize to reasonable values within 5--10 mini-batch updates. But there are many sub-fields of machine learning, few-shot learning~\cite{dhillon2019a}, meta-learning~\cite{thrun2012learning} to name some, where the number of mini-batch updates of a trained model is a key parameter and where our finding has practical relevance. To fix this issue, we can initialize the batch-normalization mean and variance estimates---easily---by doing a forward pass on a few mini-batches from the training data before beginning the training. This ensures that the model starts training from near ignorance. When we collected data from our training trajectories on ImageNet, we did not have this fix. We therefore did not plot the first checkpoint for the ImageNet experiments in~\cref{fig:imagenet_all_models_train_3d,fig:imagenet_all_models_test_3d}.


\section{Addendum to Methods}
\label{s:app:methods}

\subsection{InPCA creates an isometric embedding}
\label{s:app:isometry}
InPCA, like standard PCA, relies on an embedding directed by the centered pairwise distances~\cref{eq:w}. Observe that the centering in~\cref{eq:w} is the same as the centering performed in standard PCA, indeed it ensures that rows and columns of the pairwise distance matrix $W$ sum to zero. Since InPCA involves pairwise Bhattacharyya distances, not pairwise Euclidean distances, such a centering is not trivially equivalent to a translation of points in a vector space. We show next that the embedding created using InPCA is isometric, i.e., it satisfies~\cref{eq:isometric}. The argument developed below also holds for other embedding techniques, e.g., the IsKL method discussed in~\cref{eq:isKL} that uses the symmetrized Kullback-Leibler divergence as the distance between probability distributions.

Given a real symmetric matrix $D \in \reals^{m \times m}$, we can write $D_{ij} = \sum_k U_{ik} \L_{kk} U_{jk}$ where the eigenvalues $\L_{kk} \in \reals$ and columns of $U$ are the eigenvectors. We can define an ``eigen-embedding'' of such a matrix:
\[
    \reals \ni X_{ik} \equiv \sqrt{\lvert\L_{kk}\rvert} U_{ik};\quad i,k \leq m
\]
and a quasi inner-product $\inner{a}{b}_D \doteq \sum_k \sign(\Lambda_{kk}) a_k b_k$ for $a, b \in \reals^{p,m-p}$, with metric signature $(p,m-p)$ derived from the $p$ positive eigenvalues of $D$. The quasi inner-product of the points in an eigen-embedding of a real symmetric matrix $D$ allows us to reconstruct the entries of $D$:
\beq{
\label{eq:innerreconstruction}
    D_{ij} = \inner{X_i}{X_j}_D.
}
Now consider a finite symmetric premetric space $\MM = (M, D)$ with $\abs{M} = m$ points\footnote{A premetric space satisfies two properties: that the distance between two points is non-negative, and the distance of a point from itself is zero.}. If $D$ is a matrix of pairwise distances between these points, then it has a vanishing diagonal. The embedding of $-D/2$ denoted by $\cbr{X_i \in \reals^{p,m-p}}_{i=1}^m$ satisfies $\inner{X_i}{X_i}_{-D/2} = -D_{ii}/2 = 0$ for any $i \leq m$. Now observe that the distance between any $X_i$ and $X_j$ is the squared Minkowski interval between them, i.e.,
\beq{
    \sum_k \norm{X_{ik}-X_{jk}}^2_{-D/2} = \inner{X_i-X_j}{X_i-X_j}_{-D/2} = -(D_{ii} + D_{jj} - 2 D_{ij})/2 = D_{ij}.
    \label{eq:dx_minkowski}
}
In other words, the $m$ points in $\MM$ can be isometrically embedded in a Minkowski space as the eigen-embedding of $-D/2$. The centering operation using a matrix $L_{ij} = \delta_{ij}-1/m$ which we use to compute $W = -L D L/2$ ensures that
\[
    W_{ij} = \inner{X_i - \overline X}{X_j - \overline X}_{-D/2}
\]
where $\reals^{p,m-p} \ni \overline X = m^{-1} \sum_i X_i$ is the mean of the eigen-embedding of $-D/2$; in other words, the centered pairwise distance matrix is equal to the cross-covariance matrix in a Minkowski space.
\begin{theorem}
Given a finite symmetric premetric space $\MM = (M, D)$ with $\abs{M} = m$ points, if $D \in \reals^{m \times m}$ is the matrix of pairwise distances between these points, then the eigen-embedding of $W = -LDL/2$ where $L_{ij} = \delta_{ij} - 1/m$ is the centering matrix, is isometric to $\MM$.
\end{theorem}
\begin{proof}
Let the eigen-embeddings of $-D/2$ and $W$ be $\cbr{X_i}_{i=1}^m$ and $\cbr{Y_i}_{i=1}^m$ respectively.  We know that the eigen-embedding of $-D/2$ is isometric to $\MM$. From~\cref{eq:innerreconstruction}, we have that $\inner{Y_i}{Y_j} = W_{ij}$ and so $\inner{Y_i-Y_j}{Y_i-Y_j}_W = W_{ii}+W_{jj}-2W_{ij}$. Since the centered pairwise distance matrix is equal to the cross-covariance matrix, we also have $W_{ij} = \inner{X_i - \overline X}{X_j - \overline X}_{-D/2}$ and therefore
\[
    \aed{
    \inner{Y_i - Y_j}{Y_i - Y_j}_W
    &= \inner{X_i - \overline X}{X_i - \overline X}_{-D/2}+\inner{X_i - \overline X}{X_j - \overline X}_{-D/2}-2 \inner{X_i - \overline X}{X_j - \overline X}_{-D/2}\\
    &= \inner{X_i - X_j}{X_i - X_j}_{-D/2}\\
    &= D_{ij}.
    }
\]
\end{proof}

\subsection{Relationship between progress and error}
\label{s:app:progress_vs_error}

Progress is related to the error but they are not the same. Suppose we have a model $P$ that predicts very confidently, i.e., $p^n(c) \in \cbr{0,1}$ for all $c \in \cbr{1,\dots,C}$ and all samples $n$. The progress of this model is given by
\[
    \aed{
    \a^* &= \argmin_{\a \in [0,1]} \dG(P, P_{0,*}^\a)\\
    &= (1-\e) \cos^{-1}\rbr{\f{\sin((1-\a)\dG^n)}{\sin (\dG^n)} \cos (\dG^n) + \f{\sin (\a\dG^n)}{\sin (\dG^n)}}
    + \e  \cos^{-1}\rbr{\f{\sin ((1-\a) \dG^n)}{\sin (\dG^n)} \cos (\dG^n)}
    }
\]
where $\e = N^{-1}\sum_n \ind{\argmax_c p^n_w(c) \neq y_n^*}$ is the fraction of errors made by the model on the $N$ samples and $\dG^n = \cos^{-1}(1/\sqrt{C})$ if there are $C$ classes. We can show that if $\e < 1-1/\sqrt{C}$, then the progress $\a^* = 1$.
%
This suggests that progress and error are not directly analogous to each other: models with high progress do not necessarily have small errors. In practice, if the number of samples $N$ is small and the number of classes is large, then we will find instances of models with high progress and high error. This is not often the case in our experiments for the training data, but we do see very high progress for some models on the test data (see~\cref{fig:progress_vs_error}).

\subsection{Emphasizing different models using a weighted embedding}
\label{s:app:weighted_embedding}
To study the details of the model manifold, we have found it useful to emphasize certain models in the visualization. There are many works~\cite{vo2008weighted,gabriel1979lower,greenacre2005weighted,delchambre2015weighted} that do similar things, e.g., those that modify the underlying objective of MDS to optimize a weighted Euclidean distance (but this does not do a good job of preserving pairwise distances between points), or those that learn a set of orthogonal transformations to highlight points of interest. We can also repeat models while computing InPCA: this shifts the center of mass and,at the same time transforms the visualization (via rotations and Lorentz boosts). It emphasizes the repeated models in the visualization. However, such a naive approach is computationally expensive because the size of the distance matrix $D$ increases due to these repetitions.

We present a different approach called weighted-InPCA next. Let $D \in \reals^{m \times m}$ be the matrix of pairwise Bhattacharyya distances $D_{uv} = \dB(P_u, P_v)$ and let $\mu_u \in \naturals$ be multiplicity of the model with weights $u$, i.e., the relative importance that we would like for it in the visualization. The normalized multiplicities are $\hat \mu_u = \mu_u/\sum_{v'} \mu_{v'}$. Weighted-InPCA is a modification of InPCA. It (a) uses a different centering matrix $L_{uv} = \delta_{uv} - \hat \mu_u$, (b) performs an eigen-decomposition of $W \text{diag}(\hat \mu_u)$, i.e., each column of $W$ is multiplied by $\hat \mu_u$, and (c) then scales back each of the eigenvectors $U_i$ using the expression $U_i/\sqrt{U^\top \diag(\hat \mu) U}$. This procedure gives the same embedding as the one obtained by repeating points before calculating standard InPCA and is also equivalent to the procedure in~\cref{eq:w_expanded} when the weights $\mu_u$ of the new points are zero.



\begin{figure}
\centering
\begin{subfigure}{0.3\linewidth}
\centering
\includegraphics[width=\linewidth]{explained_pairwise_distances_subsample_train}
\caption{Train manifold}
\label{fig:explained_pairwise_distances_subsample_train_app}
\end{subfigure}
\hspace*{3ex}
\begin{subfigure}{0.3\linewidth}
\centering
\includegraphics[width=\linewidth]{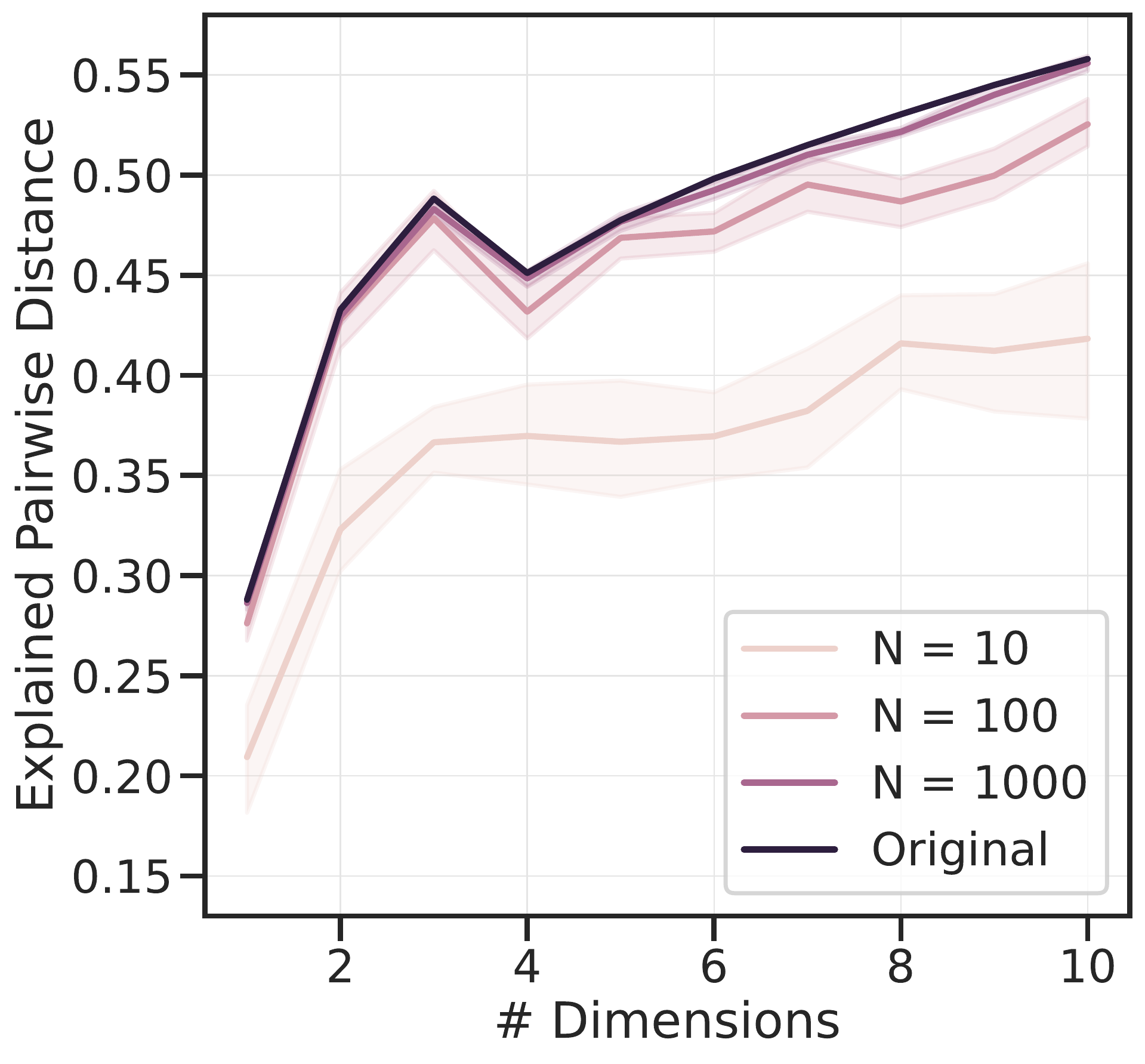}
\caption{Test manifold}
\label{fig:explained_pairwise_distances_subsample_test}
\end{subfigure}
\caption{The explained pairwise Bhattacharyya distances (computed using~\cref{eq:explained_pairwise_distances}) of the embedding when projected onto the principal components computed using a subset of the samples in the train and test data. Even for very small values of $N$, the explained pairwise distance is close to the explained distance of the original embedding computed from all the samples.}
\label{fig:explained_pairwise_distances_subsample}
\end{figure}

\begin{figure}
\centering
\begin{subfigure}[b]{0.325\linewidth}
\centering
\includegraphics[width=\linewidth]{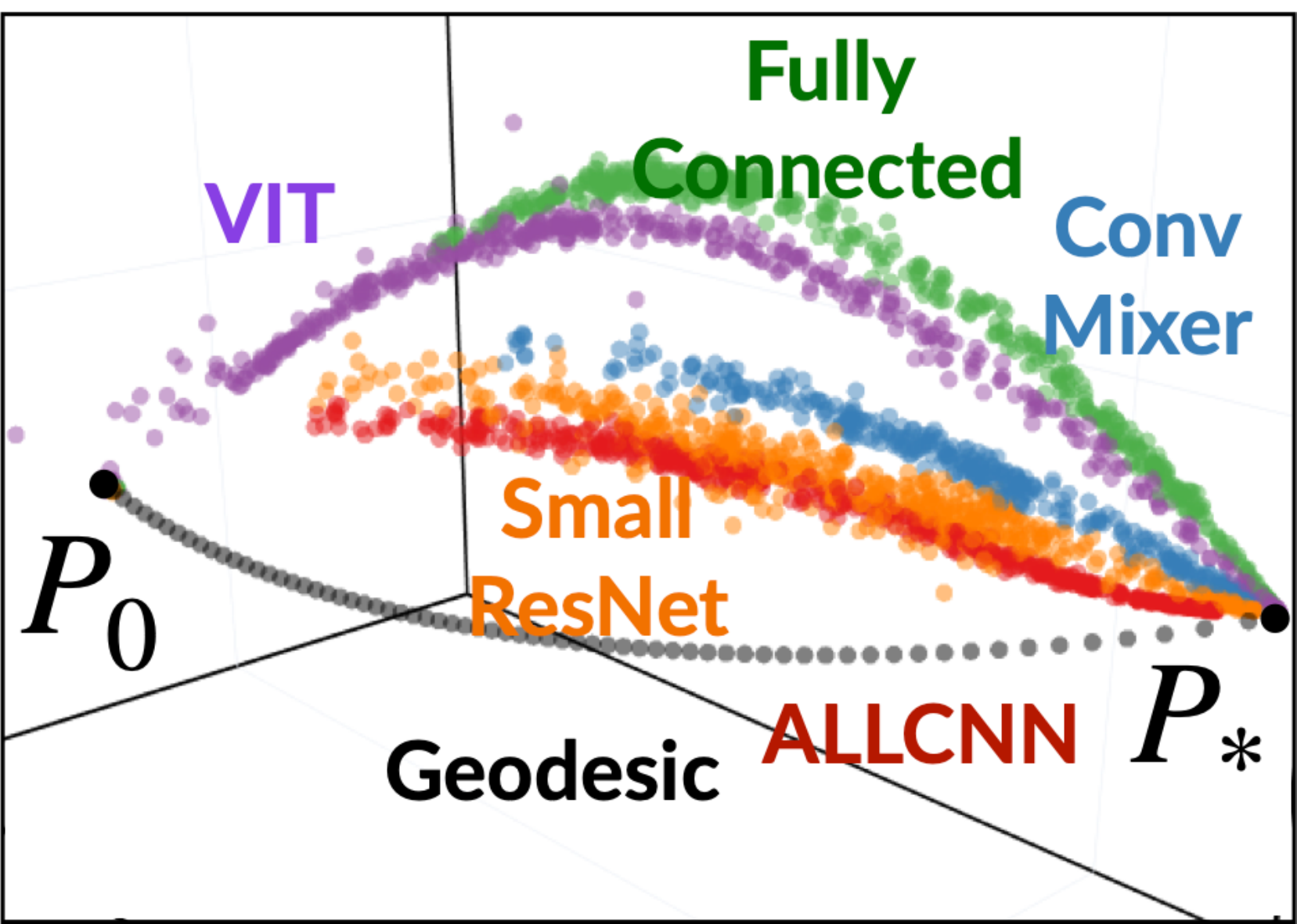}
\caption{Train $N=5000$}
\label{fig:train_subsample5000}
\end{subfigure}%
\hfill
\begin{subfigure}[b]{0.28\linewidth}
\centering
\includegraphics[width=\linewidth]{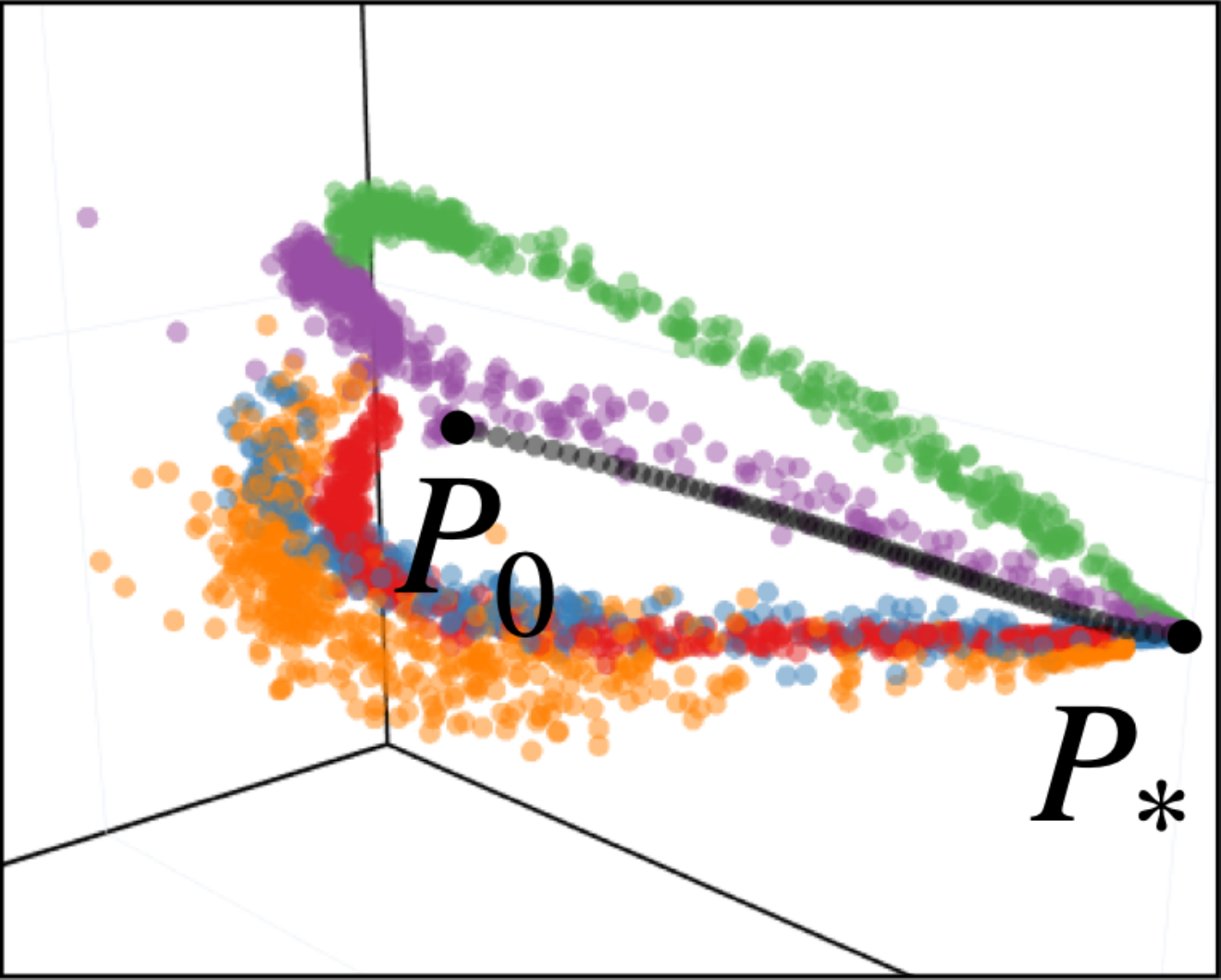}
\caption{Train $N=500$}
\label{fig:train_subsample500}
\end{subfigure}
\hfill
\begin{subfigure}[b]{0.33\linewidth}
\centering
\includegraphics[width=\linewidth]{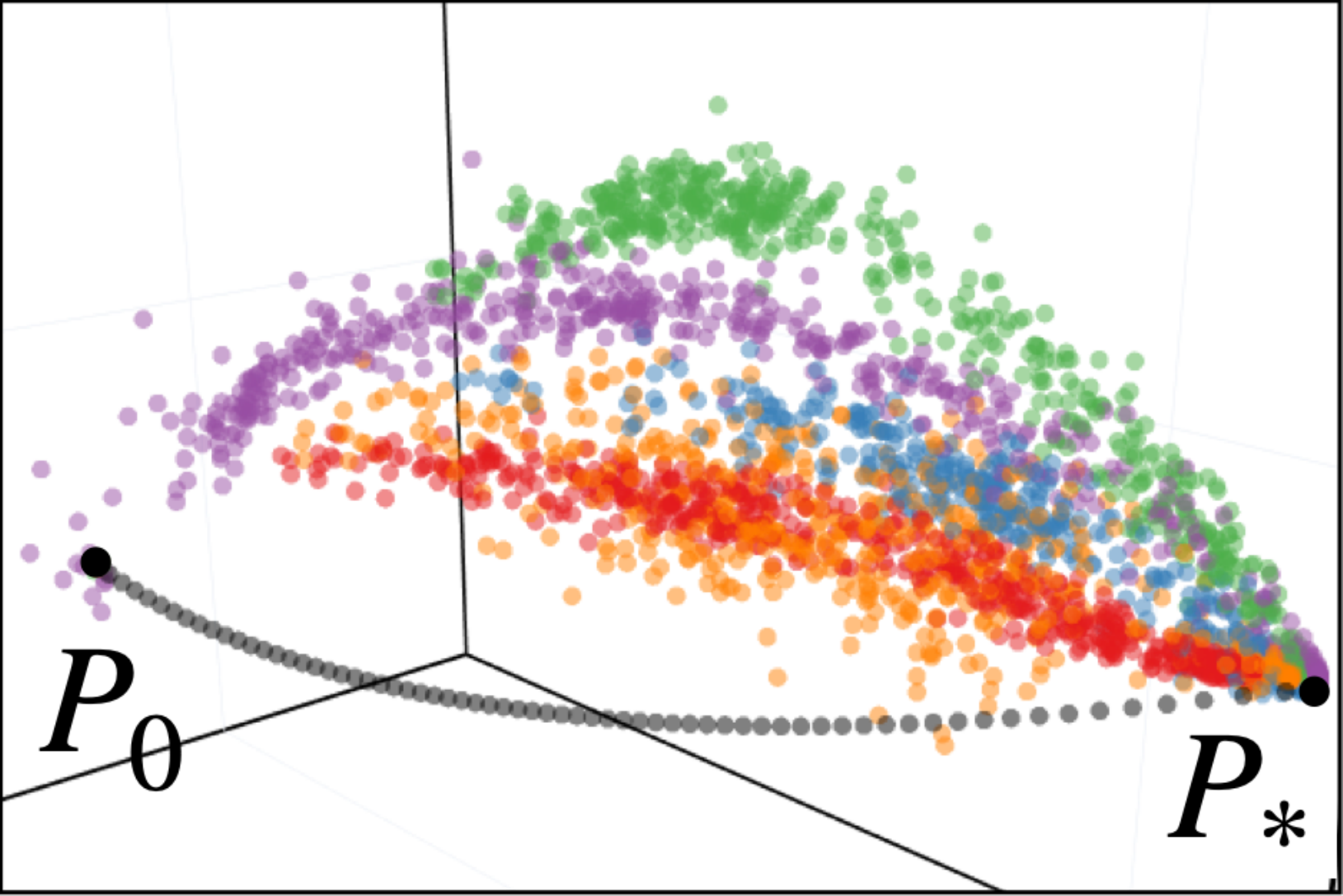}
\caption{Train $N=50$}
\label{fig:train_subsample50}
\end{subfigure}
\begin{subfigure}[b]{0.325\linewidth}
\centering
\includegraphics[width=\linewidth]{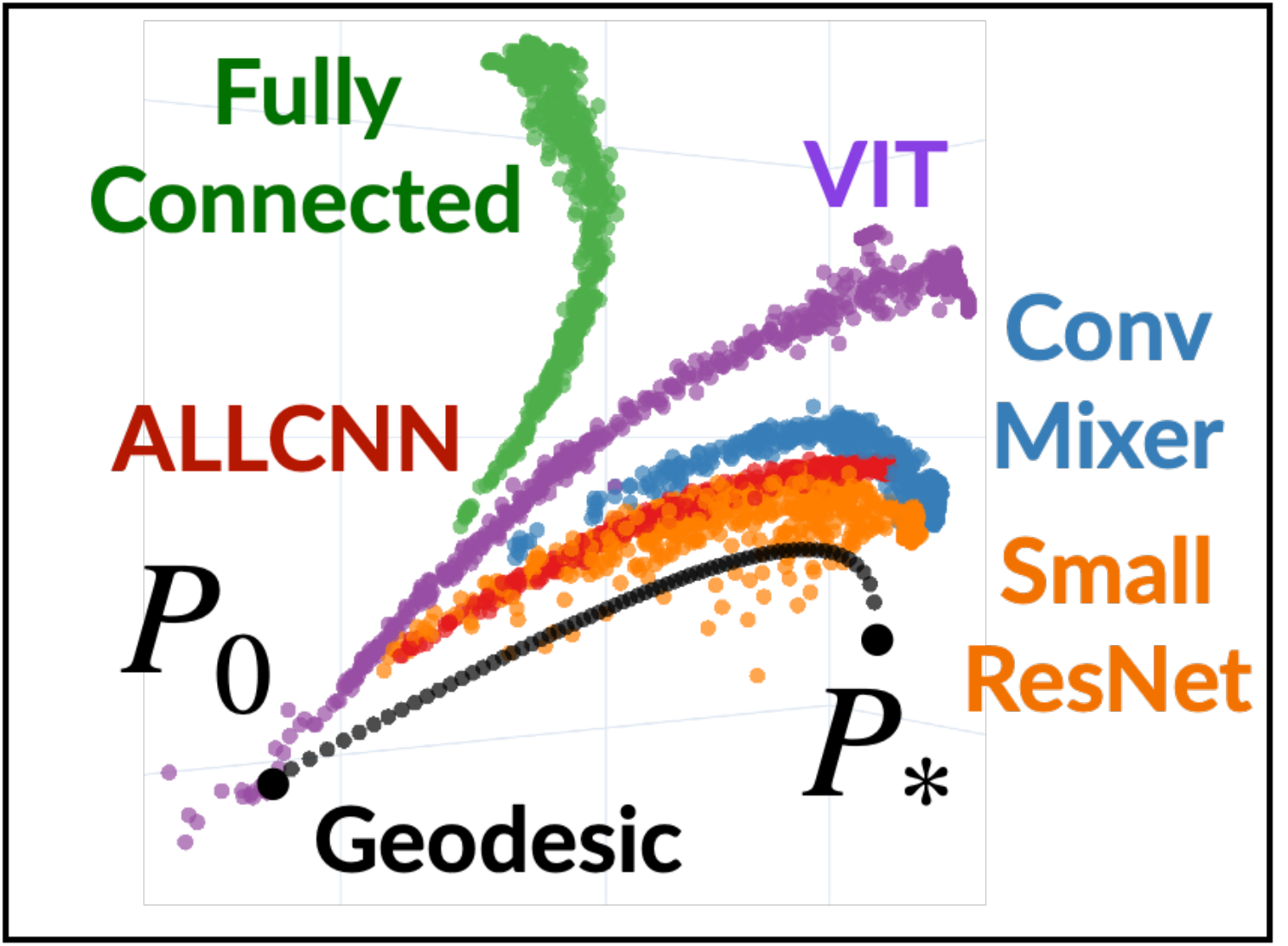}
\caption{Test $N=1000$}
\label{fig:test_subsample_1000}
\end{subfigure}
\hfill
\begin{subfigure}[b]{0.29\linewidth}
\centering
\includegraphics[width=\linewidth]{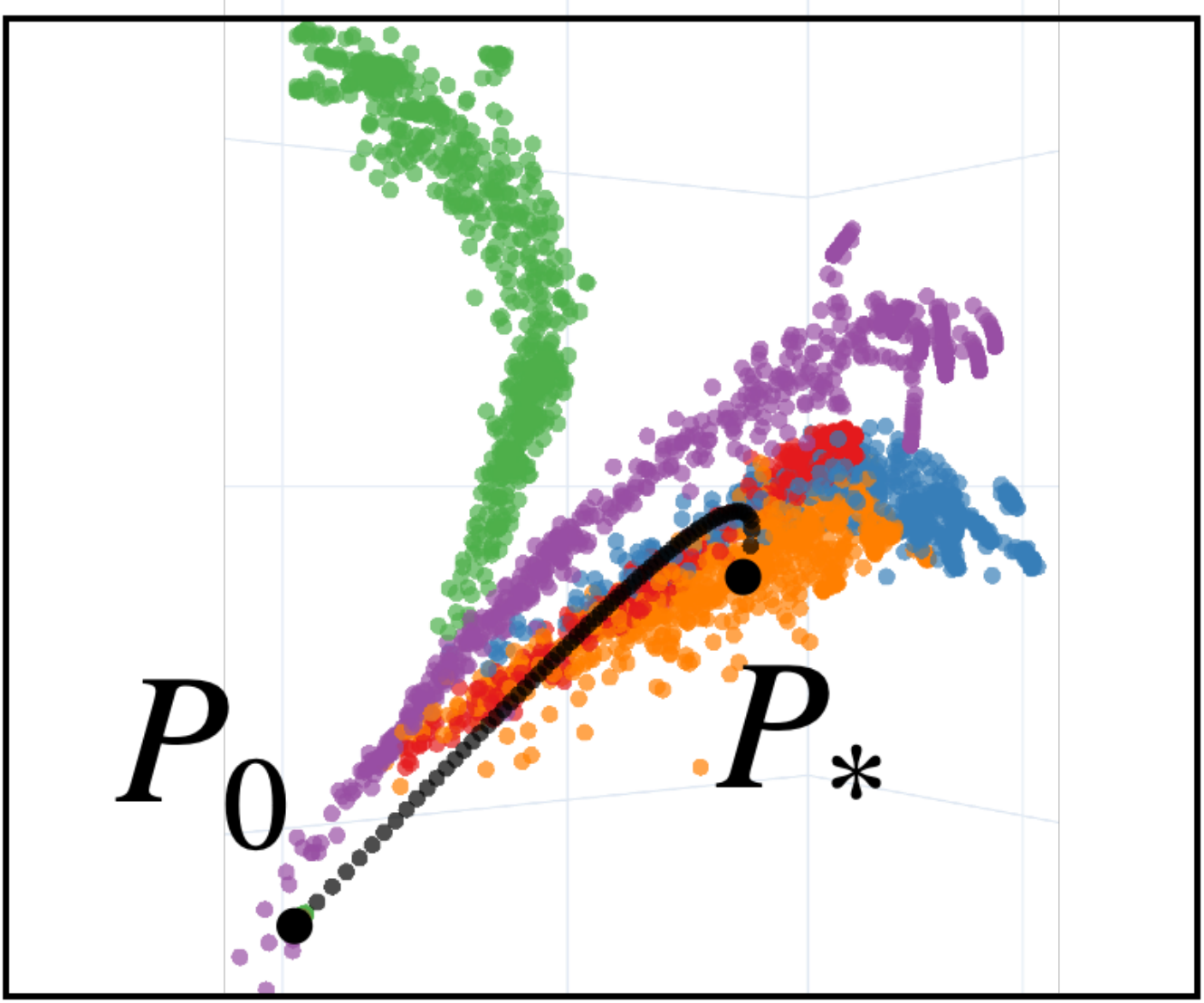}
\caption{Test $N=100$}
\label{fig:test_subsample100}
\end{subfigure}
\hfill
\begin{subfigure}[b]{0.3\linewidth}
\centering
\includegraphics[width=\linewidth]{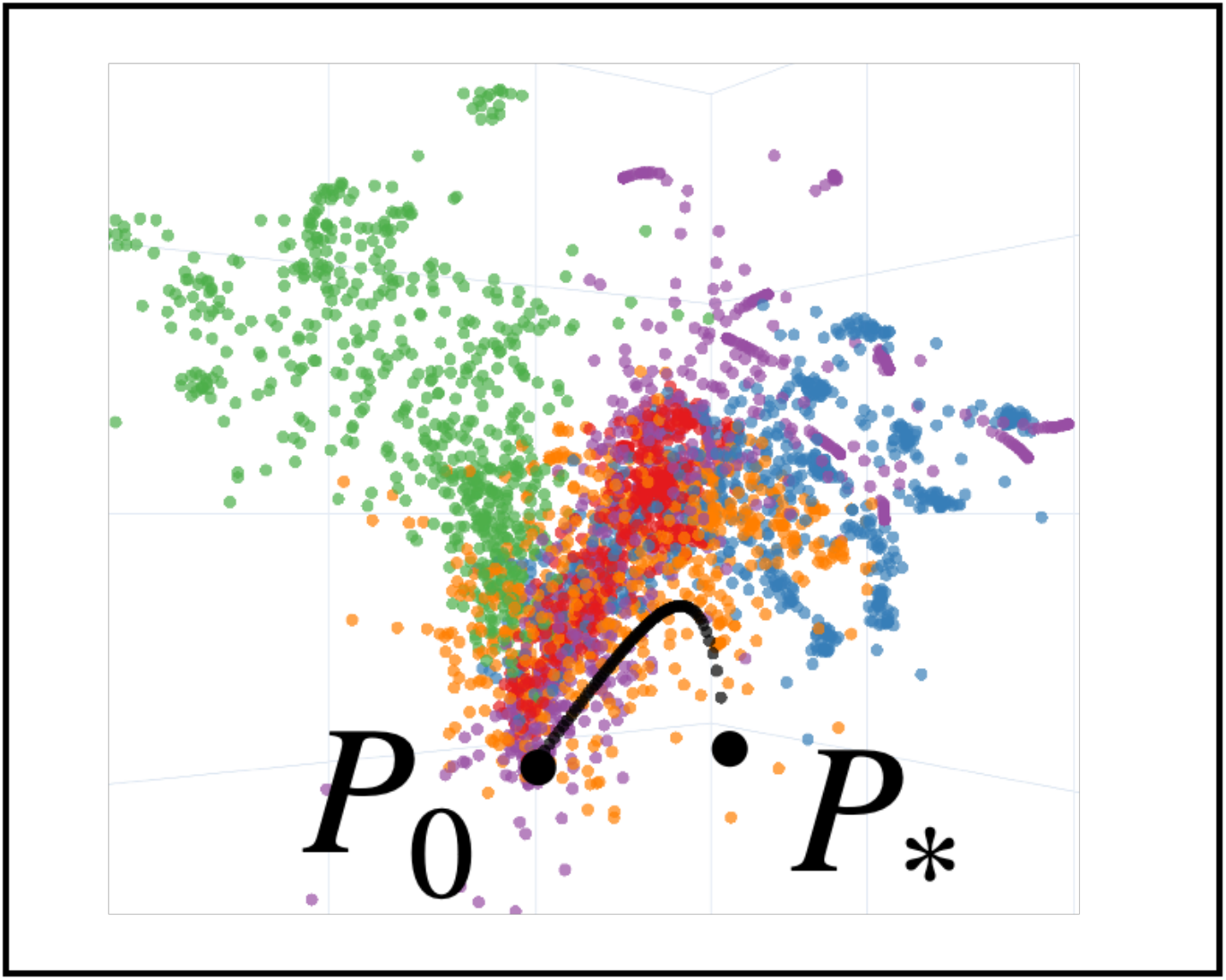}
\caption{Test $N=10$}
\label{fig:test_subsample10}
\end{subfigure}
\caption{
Projecting the original probabilistic models and pairwise Bhattacharyya distances computed on all samples into InPCA coordinates created using a distance matrix on a subset of samples (\textbf{(a-c)} for $N=5000, 500, 50$ respectively for the train data and \textbf{(d-f)} for $N=1000,100, 10$ respectively for test data). On the train data, even with as few as 1\% of the samples, these embeddings are qualitatively similar to the original embeddings (\cref{fig:all_models_train_3d,fig:all_models_test_3d}). For the test data, explained pairwise distances is low in~\cref{fig:explained_pairwise_distances_subsample_test} and manifolds are more diffuse.
}
\label{fig:subsampling}
\end{figure}

\subsection{Computing pairwise distances in InPCA using only a subset of the samples gives a faithful representation of the train and test manifolds}
\label{s:app:subset_inpca}

We computed the InPCA coordinates using a subset of the samples in the train and test sets to calculate the pairwise Bhattacharyya distance matrix. Using the procedure in~\cref{eq:w_expanded}, we then embedded the models in the original pairwise distance matrix computed using all samples into these InPCA coordinates. \cref{fig:explained_pairwise_distances_subsample,fig:subsampling} show that the explained pairwise distances by the top three dimensions of these new InPCA embeddings is quite high. This suggests that our visualization methods could be used effectively, even for large datasets with a large number of samples $N$, by sub-sampling the data before computing InPCA.

\section{Addendum to Results}

\subsection{Further analysis of the train trajectories}
\label{s:app:analysis_train_trajectories}

\begin{wrapfigure}{r}{0.35\linewidth}
\vspace*{-2em}
\centering
\includegraphics[width=\linewidth]{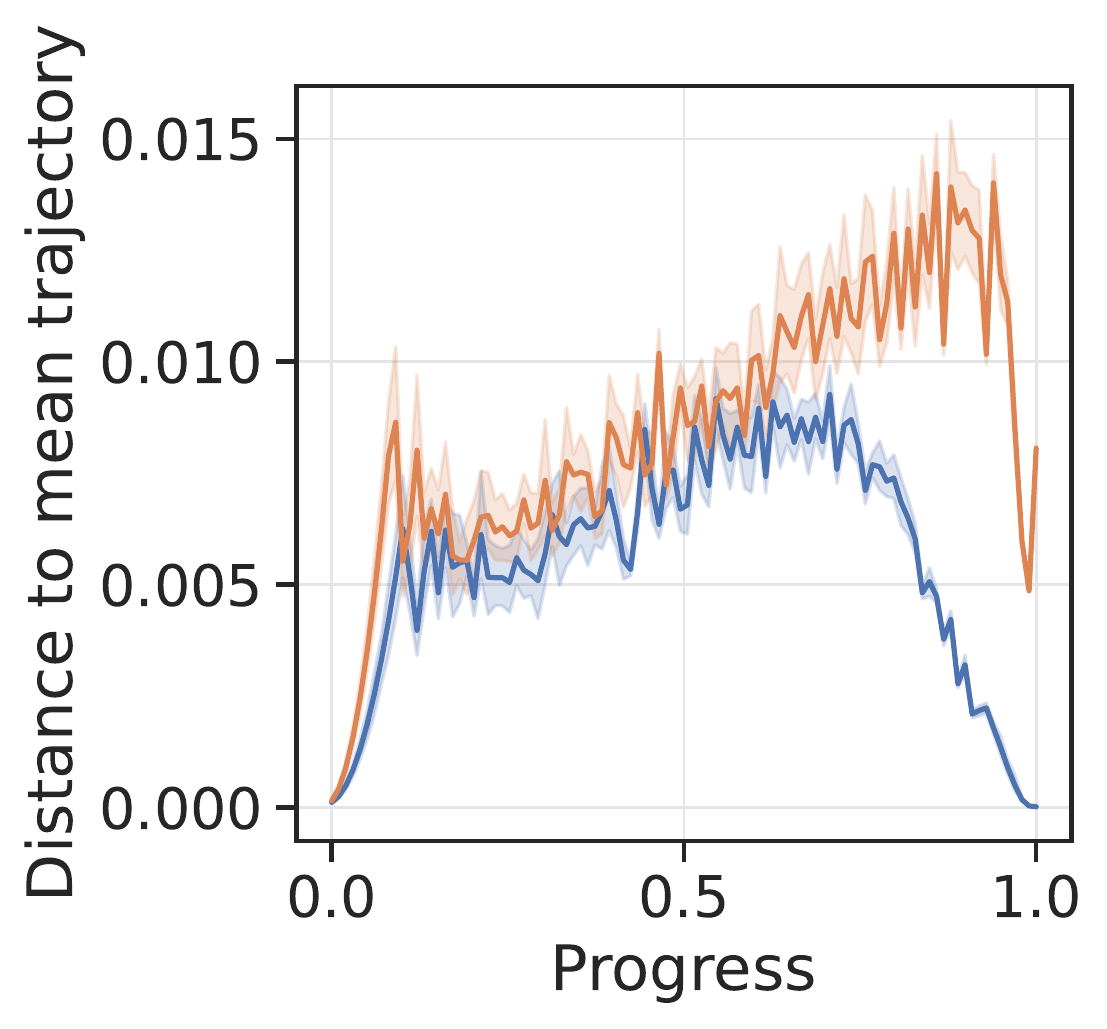}
\caption{Towards the end of training at large values of progress, models trained with augmentation (orange) have larger tube widths than models trained without augmentation (blue), on the train manifold. The corresponding figure for the test manifold looks similar.}
\vspace*{-2em}
\label{fig:allcnn_train_tube_width_aug}
\end{wrapfigure}

\begin{figure}
\centering
\begin{subfigure}[b]{0.35\linewidth}
\centering
\includegraphics[width=\linewidth]{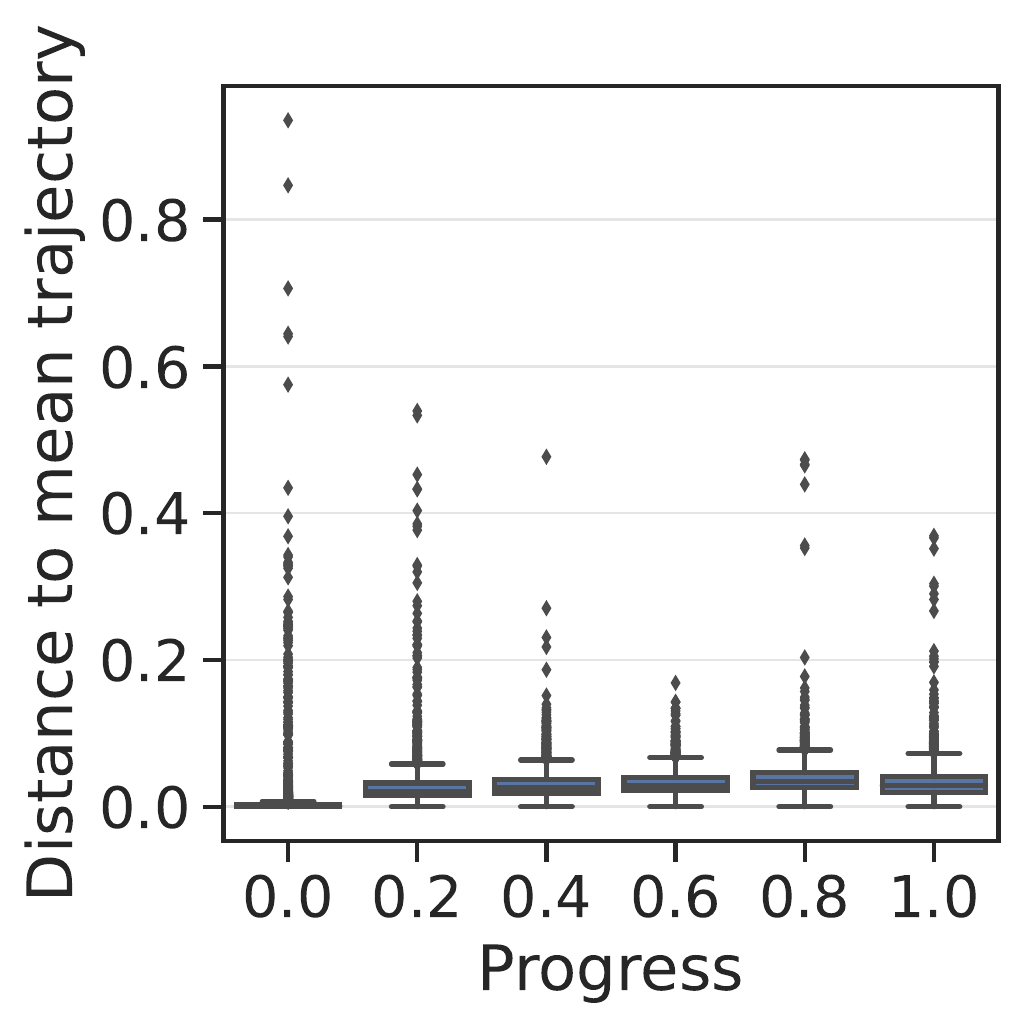}
\caption{}
\label{fig:all_models_train_tube_width}
\end{subfigure}%
\hspace*{3ex}
\begin{subfigure}[b]{0.35\linewidth}
\centering
\includegraphics[width=\linewidth]{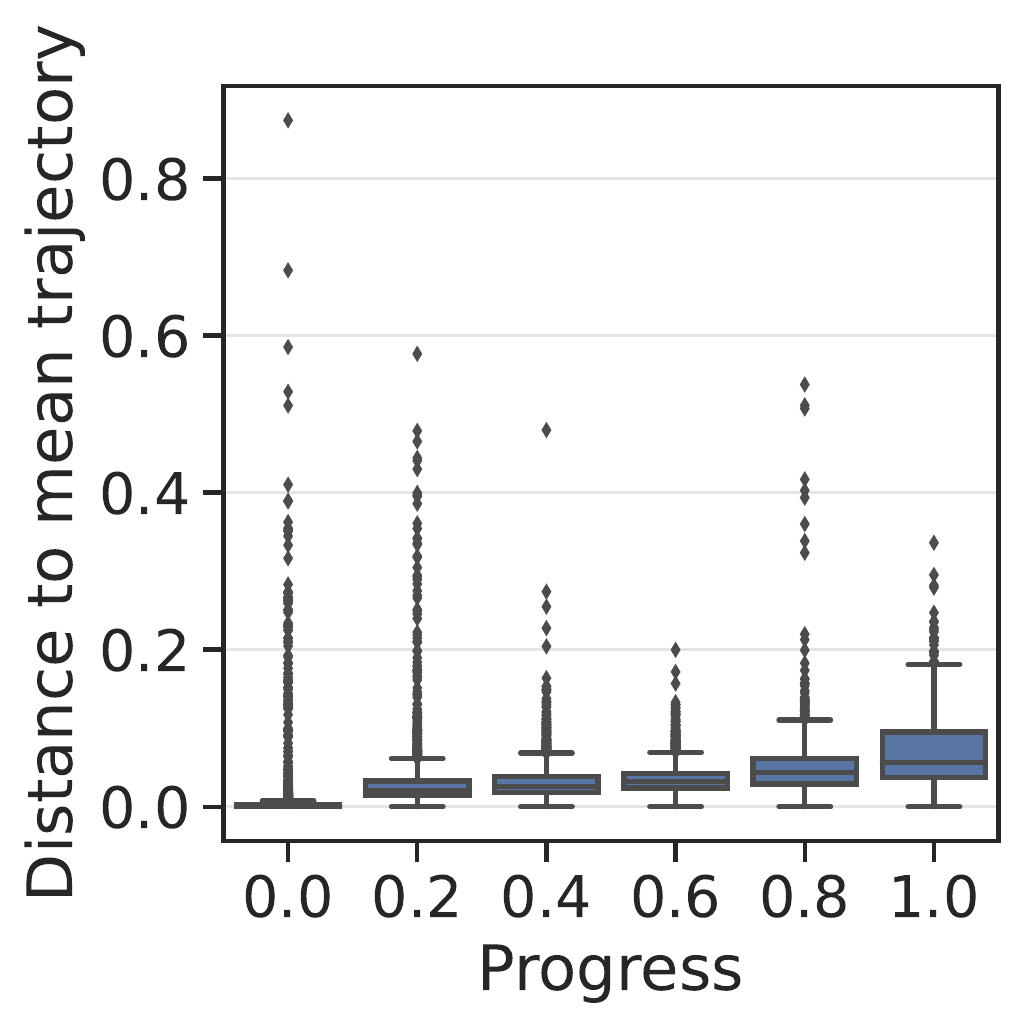}
\caption{}
\label{fig:all_models_test_tube_width}
\end{subfigure}
\caption{A boxplot (horizontal line denotes median, boxes denote 25 percentile, whiskers denote 1.5$\times$ the inter-quantile (25--75 percentile) range) of the Bhattacharyya distance between a model and the Euclidean mean of probabilities of models with the same configuration but obtained from different weight initializations for train \textbf{(a)} and test \textbf{(b)} trajectories. There are minor differences in tube widths of different configurations and therefore we have not distinguished them here. All tube-widths are quite small, which indicates that training trajectories whose configurations only differ in weight initializations are tightly clustered together in the prediction space.
}
\label{fig:all_models_tube_width}
\end{figure}

\paragraph{Understanding the differences between the trajectories of different configurations}
Using the interpolated trajectories, for each configuration, we calculated the Euclidean mean of the probabilities of the models corresponding to different weight initializations at the same progress. The distance of the model to such a configuration-specific mean model gives us an understanding of the ``tube width'', i.e., how different in prediction space models with the same progress but corresponding to different weight initializations are. \cref{fig:all_models_train_tube_width} shows that---for all configurations, for all values of progress---models are very close to their respective mean model. The median tube width is about 0.05 in terms of Bhattacharyya distance throughout training; this should be compared to the abscissae of~\cref{fig:dendrogram_train_end} where a cut at a distance of 0.05 separates all configurations (except some AllCNNs, and very few fully-connected and ConvMixer architectures). The dendrogram in~\cref{fig:dendrogram_train_end} averages models for the same progress;~\cref{fig:all_models_train_tube_width,fig:all_models_test_tube_width} indicate that such averaging is a reasonable thing to do. The test manifold in~\cref{fig:all_models_test_tube_width} is similar, except that tube widths increase slightly with progress. This suggests that networks with different weight initializations train along very similar trajectories in prediction space.

One can dig deeper into the differences in models caused by weight initialization. Tube widths of different architectures at the same progress are similar on the train manifold, but there are more pronounced differences on the test manifold. We have found that variations coming from optimization methods and regularization do not result in large tube widths. In general, towards the end of training, at large values of progress, models trained with augmentation have larger tube widths than models trained without augmentation, on both train and test manifolds (\cref{fig:allcnn_train_tube_width_aug}). Training a deep network is a non-convex optimization problem, and as such the solution depends upon the initialization of weights in a non-trivial way. Each point in the prediction space corresponds to a large set of weight configurations that lead to this same prediction. Our results therefore suggest that, even if different weight initializations could lead to different eventual weights for these non-convex optimization problems, the probabilistic models obtained at the end of training are very similar (they are more similar on the training data than the test data).

\begin{figure}
\centering
\begin{subfigure}[b]{0.45\linewidth}
\centering
\includegraphics[width=\linewidth]{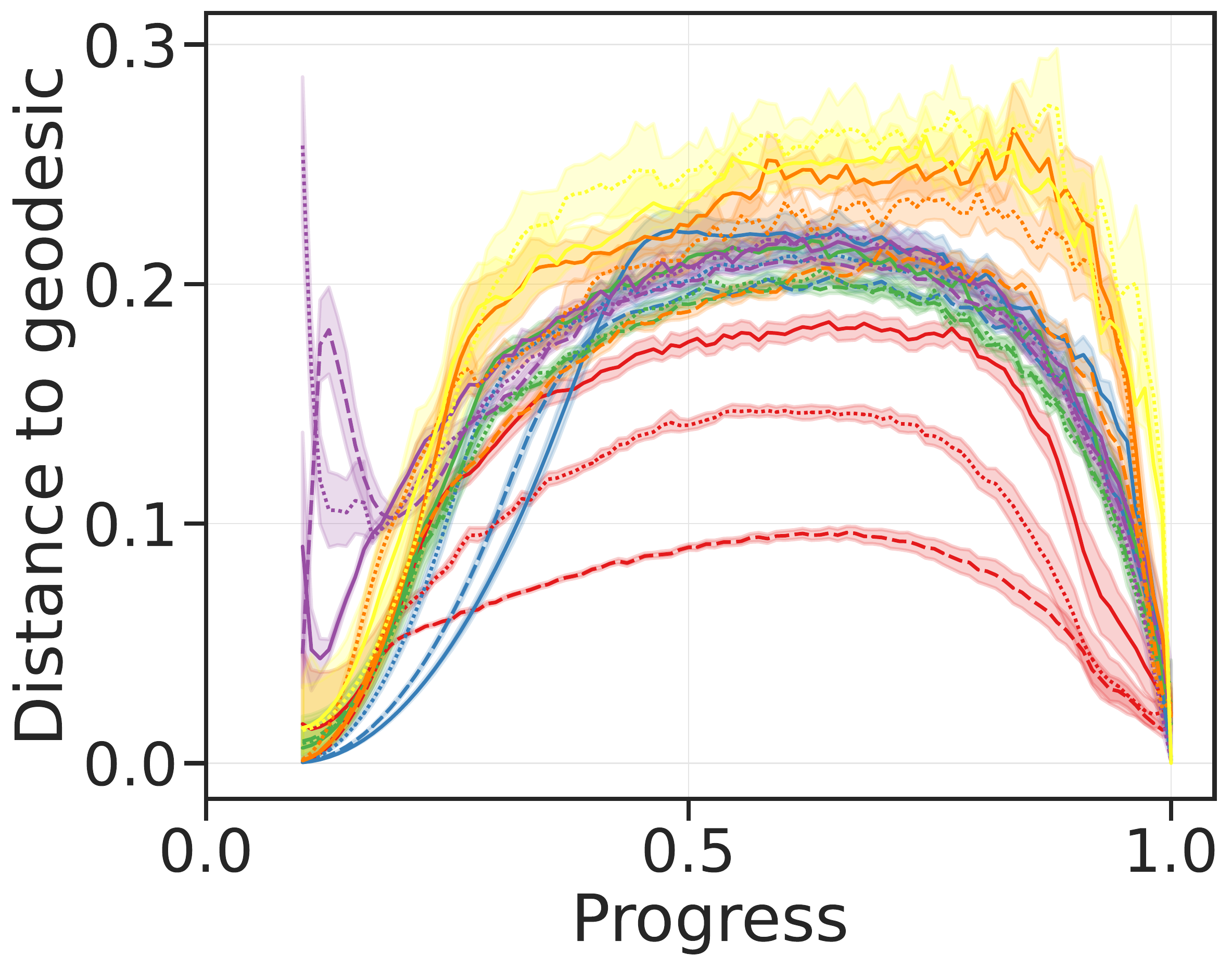}
\caption{}
\label{fig:all_models_train_d2geod}
\end{subfigure}
\hspace*{3ex}
\begin{subfigure}[b]{0.45\linewidth}
\centering
\includegraphics[width=\linewidth]{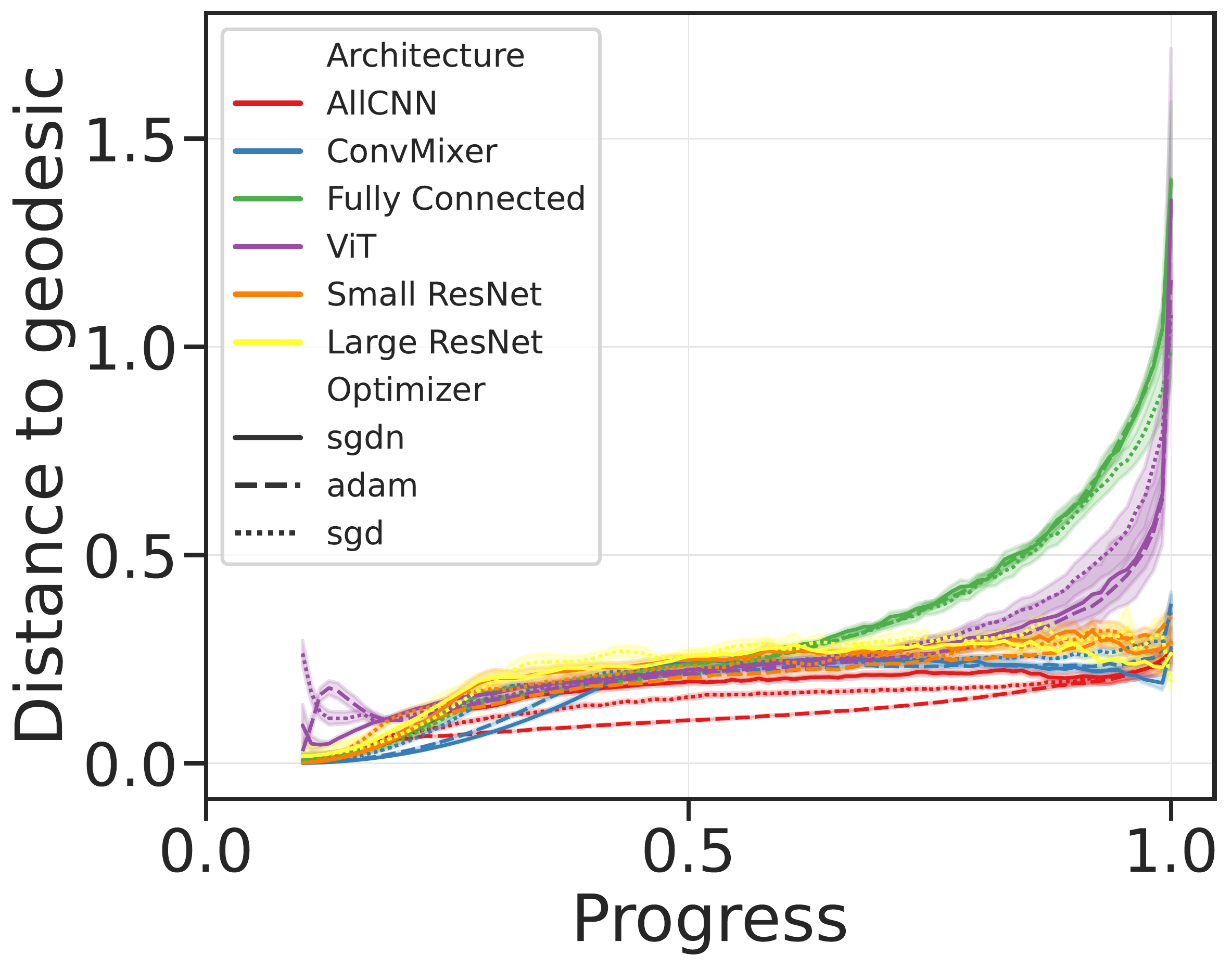}
\caption{}
\label{fig:all_models_test_d2geod}
\end{subfigure}
\caption{Bhattacharyya distance of models with different configurations to the geodesic at different progress for train \textbf{(a)}  and test \textbf{(b)}  trajectories.
}
\label{fig:all_models_d2geod}
\end{figure}

\begin{figure}
\centering
\begin{subfigure}{0.4\linewidth}
\centering
\includegraphics[width=\linewidth]{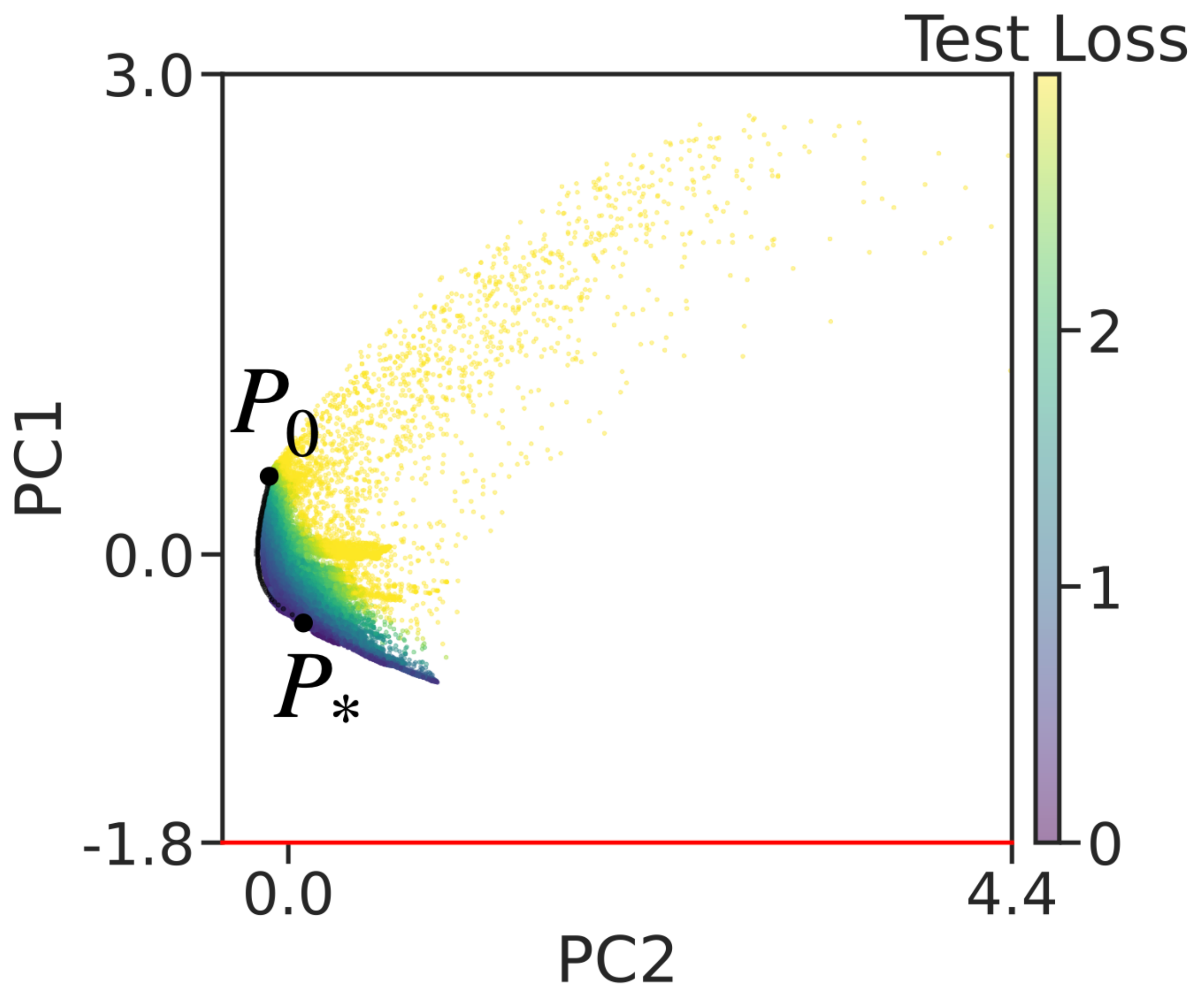}
\caption{}
\label{fig:all_models_test_2d_ps}
\end{subfigure}
\hspace*{3ex}
\begin{subfigure}{0.4\linewidth}
\centering
\includegraphics[width=\linewidth]{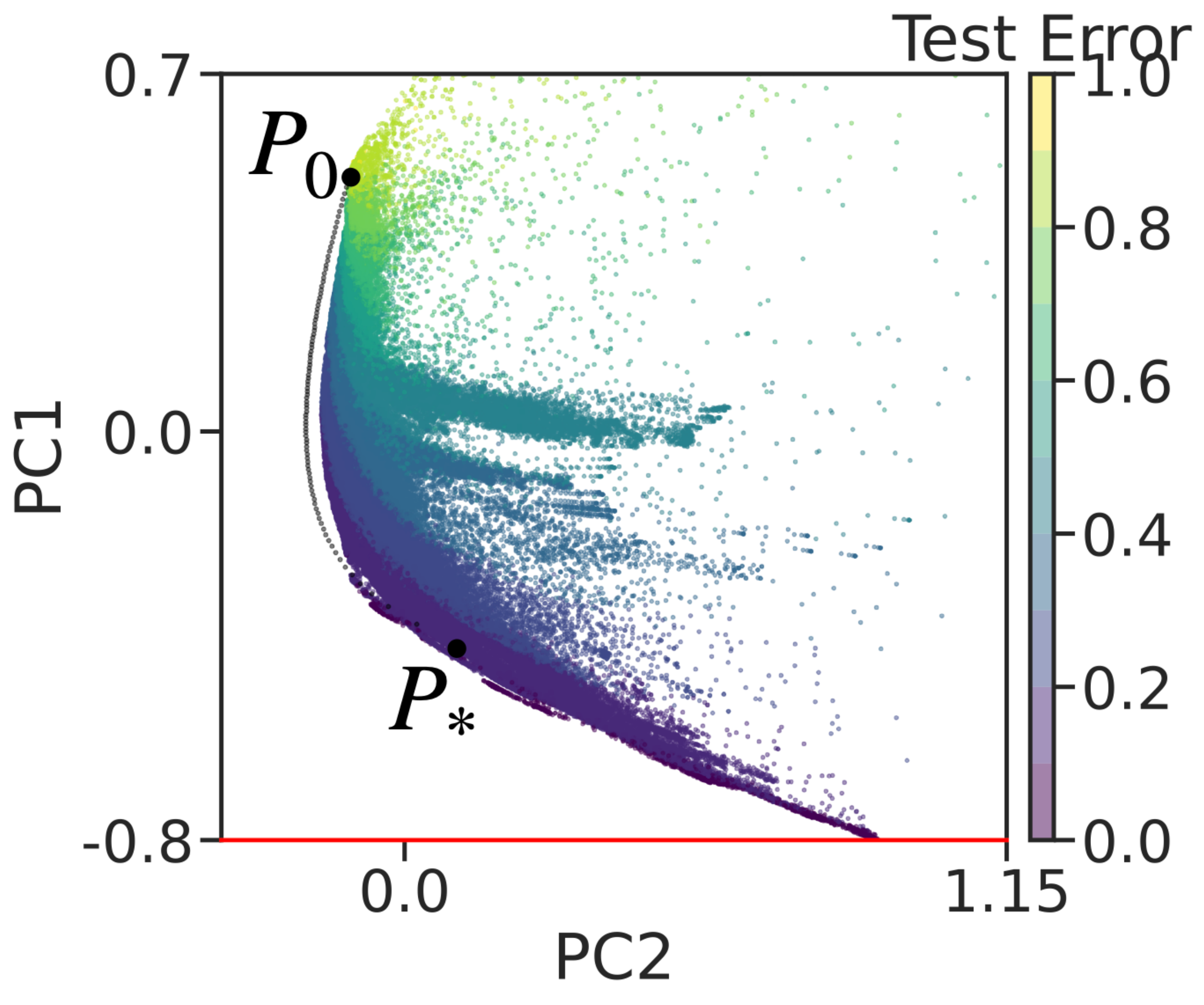}
\caption{}
\label{fig:all_models_test_2d_error}
\end{subfigure}
\begin{subfigure}{0.4\linewidth}
\centering
\includegraphics[width=\linewidth]{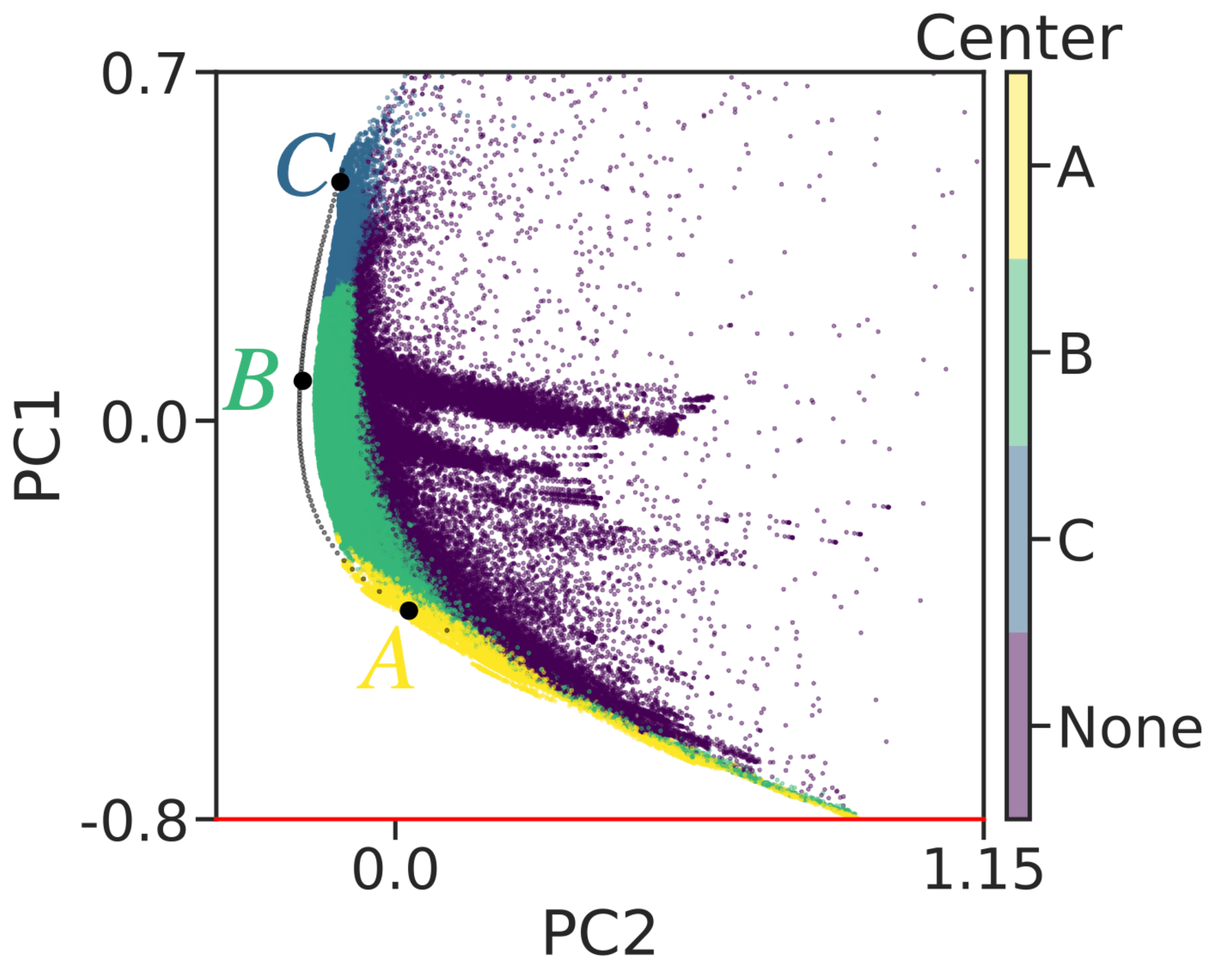}
\caption{}
\label{fig:all_models_test_2d_spread}
\end{subfigure}
\hspace*{3ex}
\begin{subfigure}{0.4\linewidth}
\centering
\includegraphics[width=\linewidth]{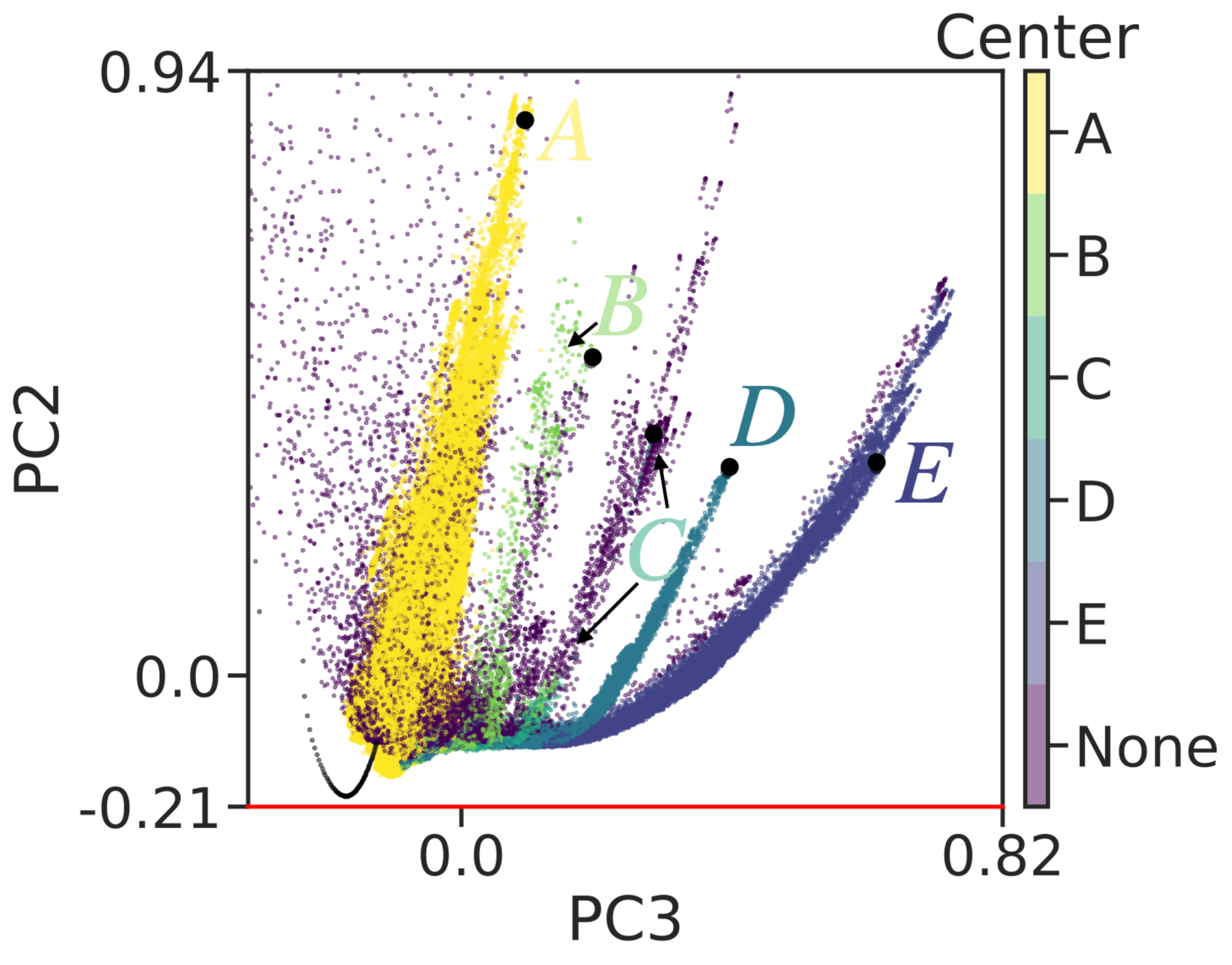}
\caption{}
\label{fig:all_models_test_2d_branches}
\end{subfigure}
\caption{Comparison of two principal components of an InPCA embedding using test data of all models on CIFAR-10 colored by test loss \textbf{(a)}, by test error \textbf{(b)}, by whether they are within a Bhattacharyya distance \textless\ 0.3 from models marked A, B, and C on the geodesic in \textbf{(c)}, and by whether they are within a distance 0.45 from the models marked A--E in \textbf{(d)}. These figures should be studied together with~\cref{fig:all_models_test_2d}.}
\label{fig:all_models_test_2d_details}
\end{figure}

\begin{figure}
\centering
\begin{subfigure}[c]{0.65\linewidth}
\centering
\includegraphics[width=\linewidth]{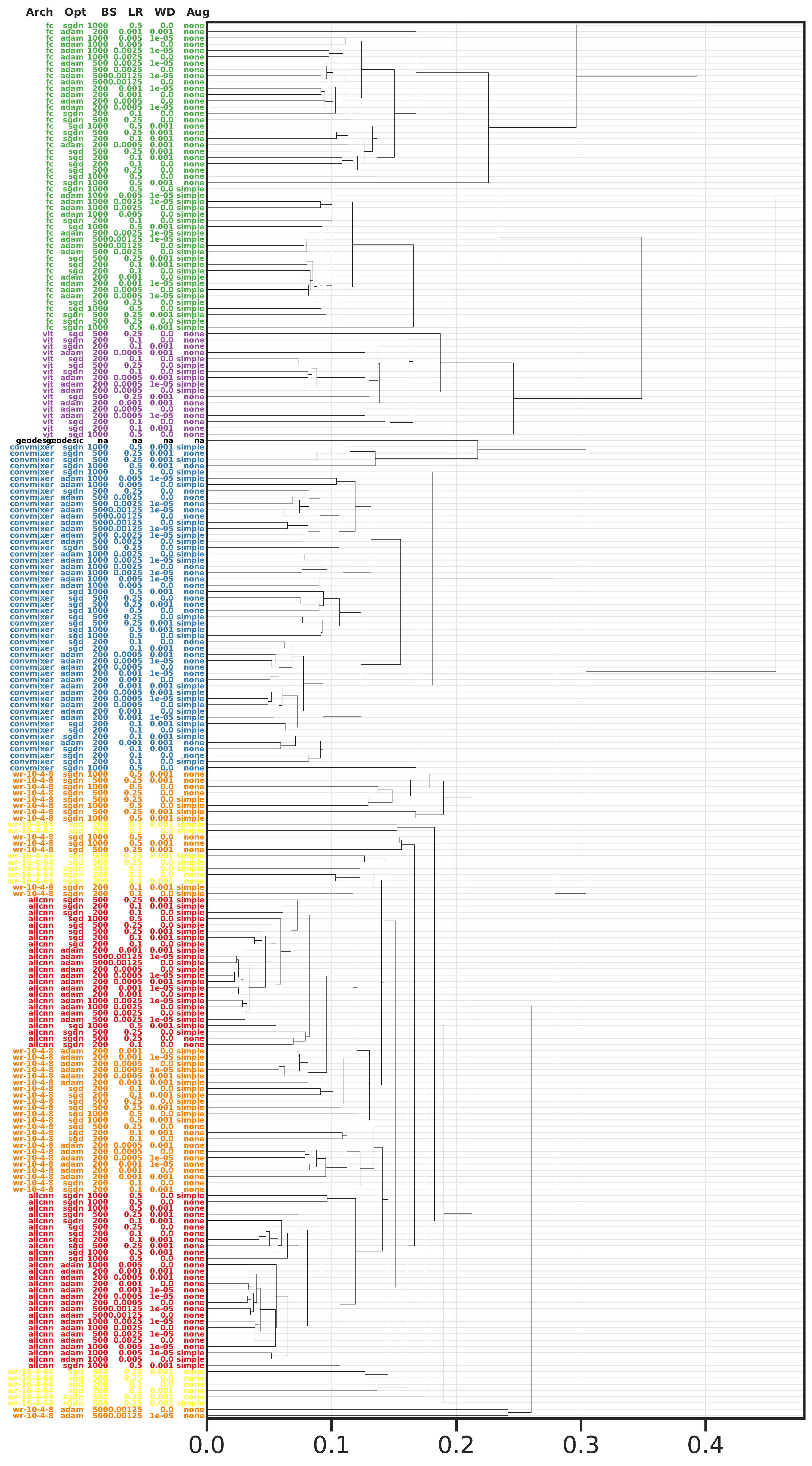}
\caption{}
\label{fig:dendrogram_test_end}
\end{subfigure}
\begin{subfigure}[c]{0.34\linewidth}
\begin{subfigure}[b]{\linewidth}
\centering
\includegraphics[width=\linewidth]{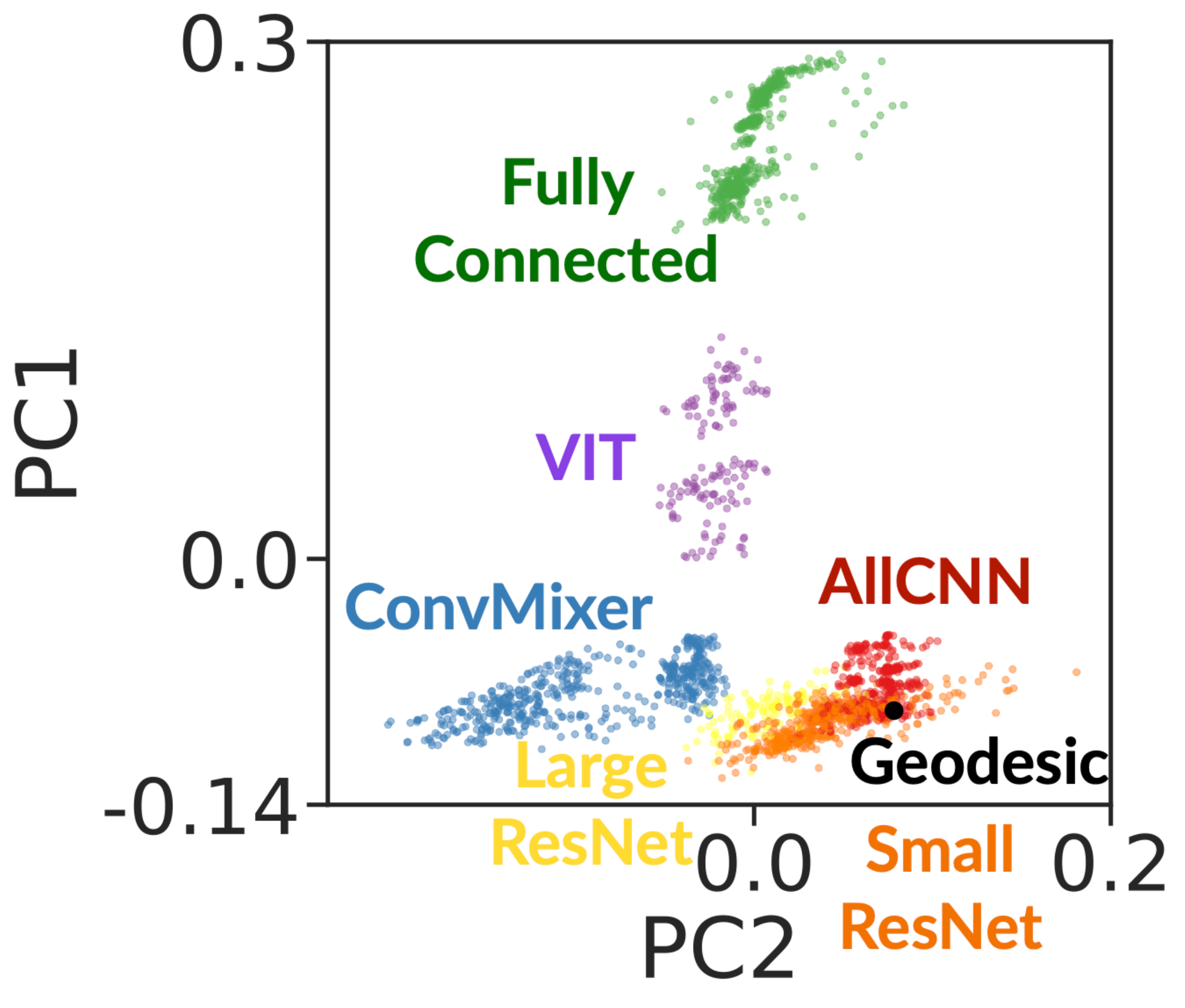}
\caption{}
\label{fig:test_trajectories_inpca}
\end{subfigure}
\begin{subfigure}[b]{\linewidth}
\centering
\includegraphics[width=\linewidth]{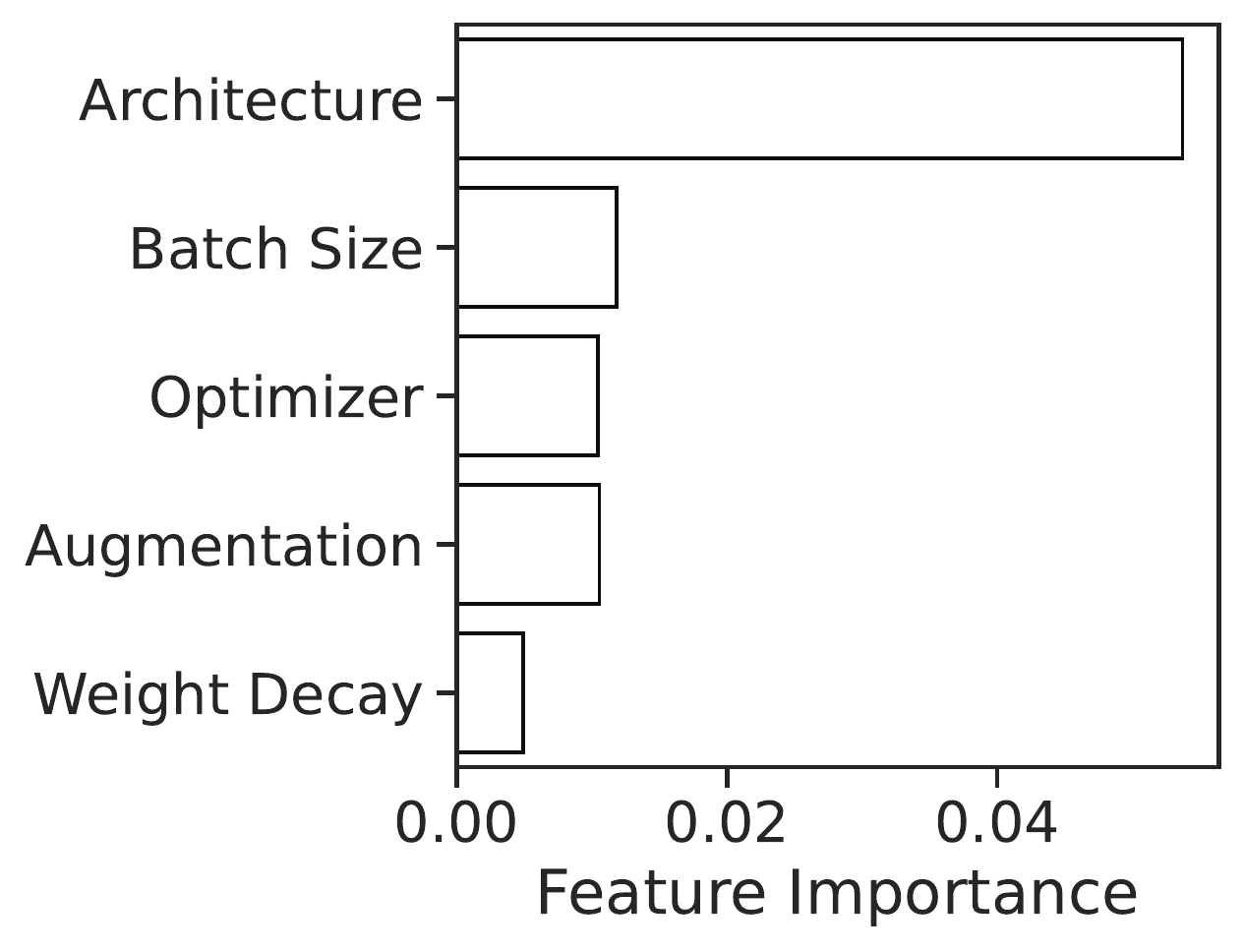}
\caption{}
\label{fig:test_trajectories_feature_importance}
\end{subfigure}
\end{subfigure}
\caption{
\textbf{(a)}: dendrogram obtained from hierarchical clustering of pairwise distances (averaged over weight initializations) between trajectories using distances calculated on testing samples. X-labels correspond to architecture, optimization algorithm, batch-size, learning rate, weight-decay coefficient and augmentation strategy. Compared to the equivalent figure on training data~\cref{fig:dendrogram_train_end}, trajectories still form clear clusters according to architecture, the distances between different trajectories are in general larger on test data, and the clusters of large and small wide ResNets are less distinguishable. \textbf{(b)} the first two components of an InPCA embedding (without averaging over weight initializations) of these trajectories, each point is one trajectory; explained stress of top two dimensions is 73.7\%. \textbf{(c)} variable importance from a permutation test ($p < 10^{-6}$) using a random forest to predict pairwise distances. These three plots suggest that for test data, architecture is still the primary distinguishing factor of trajectories in the prediction space, and the picture of different trajectories is very similar to those evaluated on training data, even though they appear to have a larger difference in the InPCA embedding.}
\label{fig:result2_test}
\end{figure}

We next study the distances of models along the interpolated trajectories to the geodesic. On the train manifold (\cref{fig:all_models_train_d2geod}), all models are very close to the geodesic at the beginning (small progress) and at the end of training (large progress). At intermediate progress, all trajectories have large distances to the geodesic; as we discussed above this deviation away from the geodesic could be an indicator of the range of difficulties of learning different samples. Trajectories corresponding to different architectures and optimization methods are at different distances from the geodesic at intermediate progress. Train trajectories of AllCNN are closest to the geodesic; there are marked differences between the three optimization algorithms in this case. But this is not so for other architectures. For test trajectories (\cref{fig:all_models_test_d2geod}), the distance to the geodesic is roughly the same, and larger that that of the train manifold, for all architectures and all values of progress. At large progress, test trajectories of fully-connected and ViT networks are very far from the geodesic; this is also visible in~\cref{fig:all_models_test}.

\subsection*{Models initialized at very different parts of the prediction space converge to the truth along a similar manifold}
\change{
The manifold in our analysis is the set of probabilistic models explored during the training process; this is a subset of the space of all probabilistic models (which is the simplex in $[0,1]^{NC}$ and not low-dimensional). Our manifold is a subset of the manifold of all probabilistic models that can be expressed by the network $\cbr{P_w(\vec y): \forall w}$ (which is also not expected to be low-dimensional) because the training process does not explore all parts of the weight space. To understand why our trajectories seem to lie on effectively low-dimensional manifolds, using CIFAR-10, we created three different tasks by randomly assigning labels to the images, e.g., each image of a dog is labeled independently as any of the 10 possible classes. This gives us three random initial models denoted by $P_0^{(k)}$ for $k \in \cbr{1,2,3}$, and we can now train networks to fit these random labels. Both train and test manifolds of training to such random tasks are effectively low-dimensional. This suggests that the low-dimensionality is not necessarily due to there being learnable patterns in the labels.
}
\begin{figure}
\centering
\begin{subfigure}[c]{0.49\linewidth}
\centering
\includegraphics[width=\linewidth]{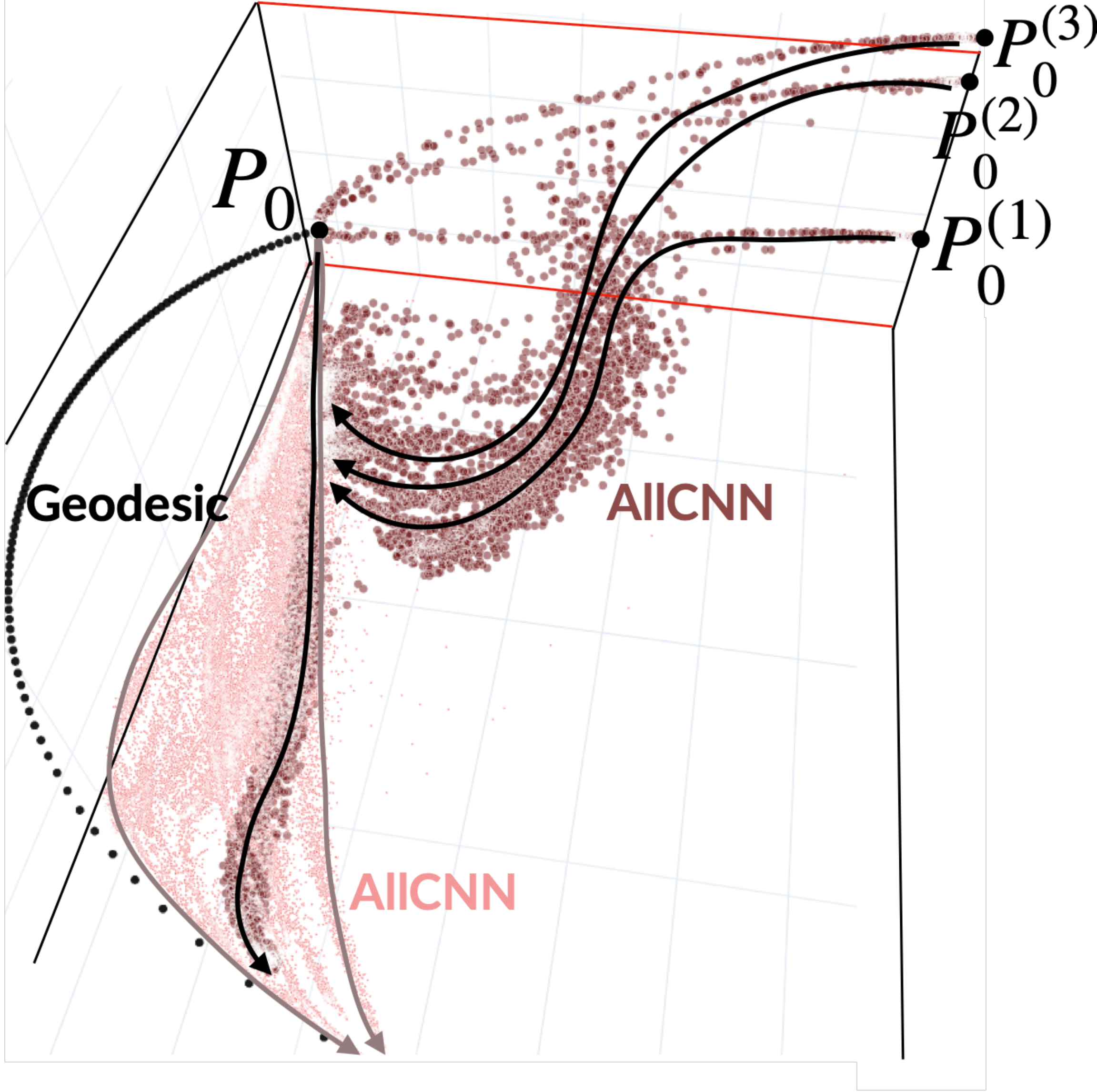}
\caption{}
\label{fig:allcnn_train_from_corners}
\end{subfigure}%
\begin{subfigure}[c]{0.42\linewidth}
\centering
\includegraphics[width=\linewidth]{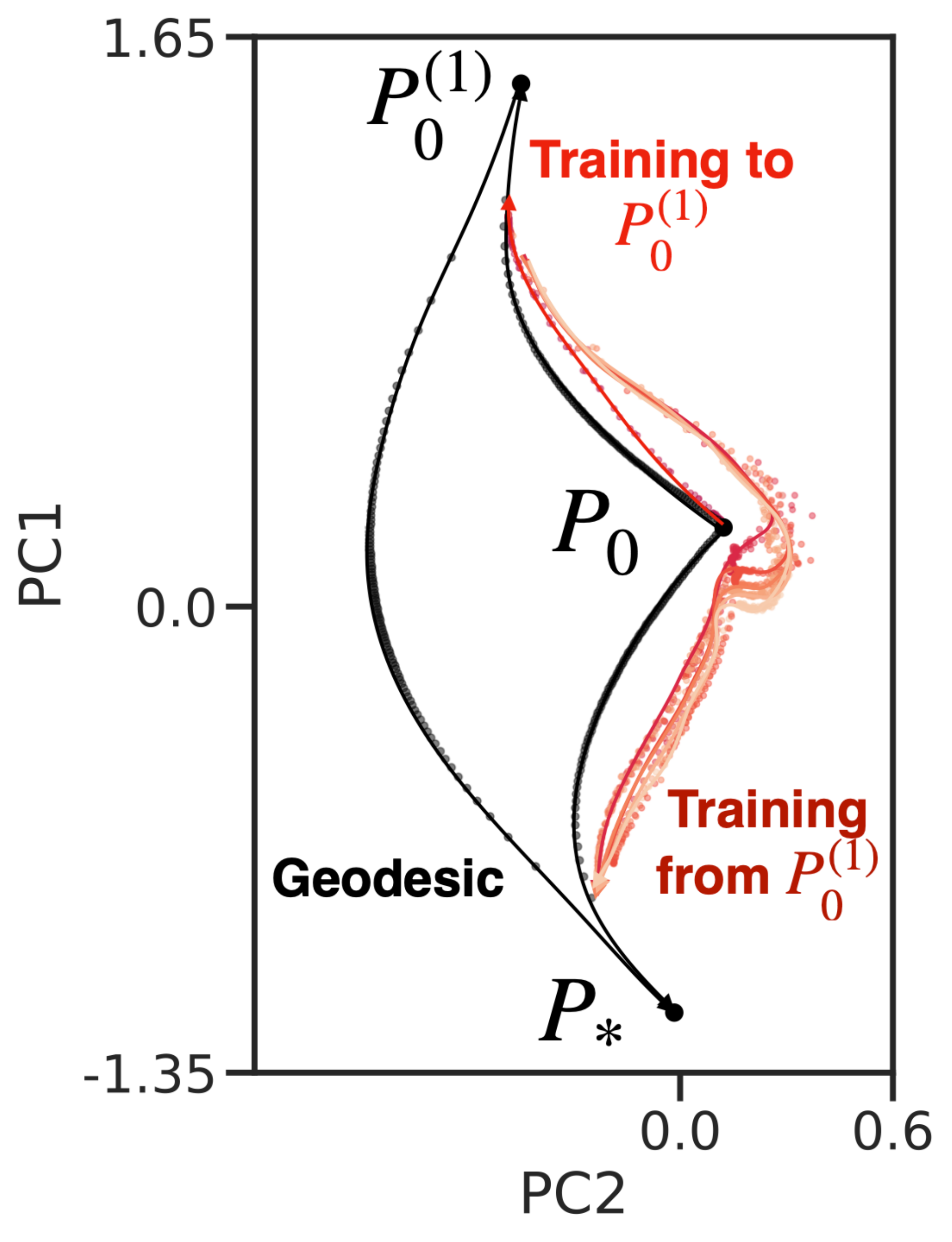}
\caption{}
\label{fig:allcnn_train_from_one_corner}
\end{subfigure}
\caption{\textbf{(a)} shows the top three dimensions of an InPCA embedding of some configurations with AllCNN architectures when networks are initialized near ignorance and trained to truth $P_*$ (light brown), and when they are first trained to tasks $P_0^{(k)}$ for $k=1,2, 3$ with random labels (stream of brown points heading towards these corners) and then further trained to the truth $P_*$. Trajectories from random tasks join the original train manifold before heading to truth (black curves in \textbf{(a)} for trajectories that begin at different random tasks and red in \textbf{(b)} for trajectories corresponding to different weight initializations from the same random task). These trajectories are very different from geodesics. We have drawn smooth curves denoting trajectories by hand to guide the reader.
Note that the trajectories that begin at corners with random labels rejoin the trajectories that begin from near ignorance quite close to ignorance but along paths
}
\label{fig:allcnn_from_corners}
\end{figure}

We next performed a second stage of training where networks were initialized to the endpoints of the trajectories to $P_0^{(k)}$ for $k\in \cbr{1,2,3}$ (models do not reach these points exactly during training), and trained on the actual CIFAR-10 task, i.e., to the actual truth $P_*$.  In this case, we only trained one particular configuration (AllCNN architecture, SGD without Nesterov's acceleration, no augmentation or weight-decay) from 10 different weight initializations chosen to be near $P_0^{(k)}$. This two-stage training procedure also results in effectively low-dimensional train and test manifolds (\cref{fig:allcnn_train_from_corners,fig:allcnn_test_from_corners}); the top three dimensions explain more than 87\% of the stress. It is interesting to note that the networks don't just forget the wrong labels before learning the correct ones, trajectories rejoin the original training trajectory at a variety of points before following it to the truth.

In \cref{fig:allcnn_train_from_one_corner} we show the training trajectories to (light red) and from (red) $P_0^{(1)}$, together with the geodesics connecting $P_0$, $P_*$ and $P_0^{(1)}$. The geodesic from $P_0^{(1)}$ to the truth does not pass near ignorance $P_0$. In fact, a random task $P_0^{(k)}$ agrees with the truth on approximately $1/C$ of the samples, and the Bhattacharyya distance of the geodesic from $P_0^{(k)}$ to the truth is at least a distance $\log(C)/(2C)+ ((C-1) \log(C/2))/(2C)$ ($\approx 0.83$ for $C=10$) from ignorance. As a reference, the distance between training trajectories of two different configurations is about 0.15 in~\cref{fig:dendrogram_train_end}. Unlike the geodesic from $P_0^{(k)}$, trajectories from $P_0^{(1)}$ come much closer to ignorance; the smallest distance from $P_0$ ranges from 0.1--0.5 for different weight initializations. There is a large spread in the models near ignorance and trajectories with different weight initializations join along separate paths (\cref{fig:allcnn_train_from_one_corner}). After progress of 0.27 ± 0.15 (which is typically achieved within 3 epochs), most models have a distance of less than 0.15 from models that began training from ignorance $P_0$. This suggests a remarkable picture for the train manifold: not only do trajectories that begin near ignorance $P_0$ lie on it, but even if trajectories begin at very different parts of the prediction space, they still join this manifold before heading to the truth. Conclusions on test data in~\cref{s:app:analysis_test_trajectories} are similar.

\subsection{Further analysis of trajectories on the test data}
\label{s:app:analysis_test_trajectories}

\paragraph{Dendrogram and InPCA embedding of test trajectories}
\cref{fig:result2_test} shows a dendrogram, similar to the one in~\cref{fig:dendrogram_train_end}, obtained from hierarchical clustering of pairwise distances (averaged over weight initializations) between trajectories using distances calculated on the test samples. \cref{fig:test_trajectories_inpca} shows an InPCA embedding of the test trajectories and~\cref{fig:test_trajectories_feature_importance} shows a variable importance plot using a random-forest to predict the pairwise distances between test trajectories. The conclusions drawn from these plots on the test data are very similar to those on the train data in~\cref{fig:result2_train} discussed in the main paper.

\paragraph{Characterizing the details of the test manifold}
We will first study the spread of points away from the test manifold. Consider~\cref{fig:all_models_test_2d_ps}, which shows points in the first two components colored by their distance to truth $P_*$. Points colored purple have the smallest distance and the best test loss. This is corroborated by~\cref{fig:all_models_test_2d_spread} where we took three points on the geodesic and colored models in terms of whether they are within a Bhattacharyya distance of 0.3 from these centers. Points that are away from the test manifold at early training times are colored yellow in~\cref{fig:all_models_test_2d_ps}; they consequently have high errors (90\% in many cases, colored yellow in~\cref{fig:all_models_test_2d_error}). We checked that these are the same models that are far from the train manifold near ignorance $P_0$ (yellow in~\cref{fig:all_models_train_2d_ps}). Some (about half) of these models did not reach zero training error, and correspondingly they also have poor test error.

In~\cref{fig:all_models_test_2d_ps}, we see that there is a large number of models that form a sliver of the manifold near truth $P_*$; these are primarily ConvMixer and Large ResNet architectures. Their test errors are \textless\ 10\% (see~\cref{fig:all_models_test_2d_error}), and their Bhattacharyya distance to the truth is \textless\ 1. In the train manifold, the spread in the visualization was coming due to InPCA amplifying small differences in the models, all with zero error, towards the end of training. In the test manifold, these models also have similar predictions (as seen in~\cref{fig:all_models_test_2d_spread}) but they do not have zero error. InPCA is again identifying differences in the underlying probabilistic models.

\begin{wrapfigure}{r}{0.35\linewidth}
\centering
\includegraphics[width=\linewidth]{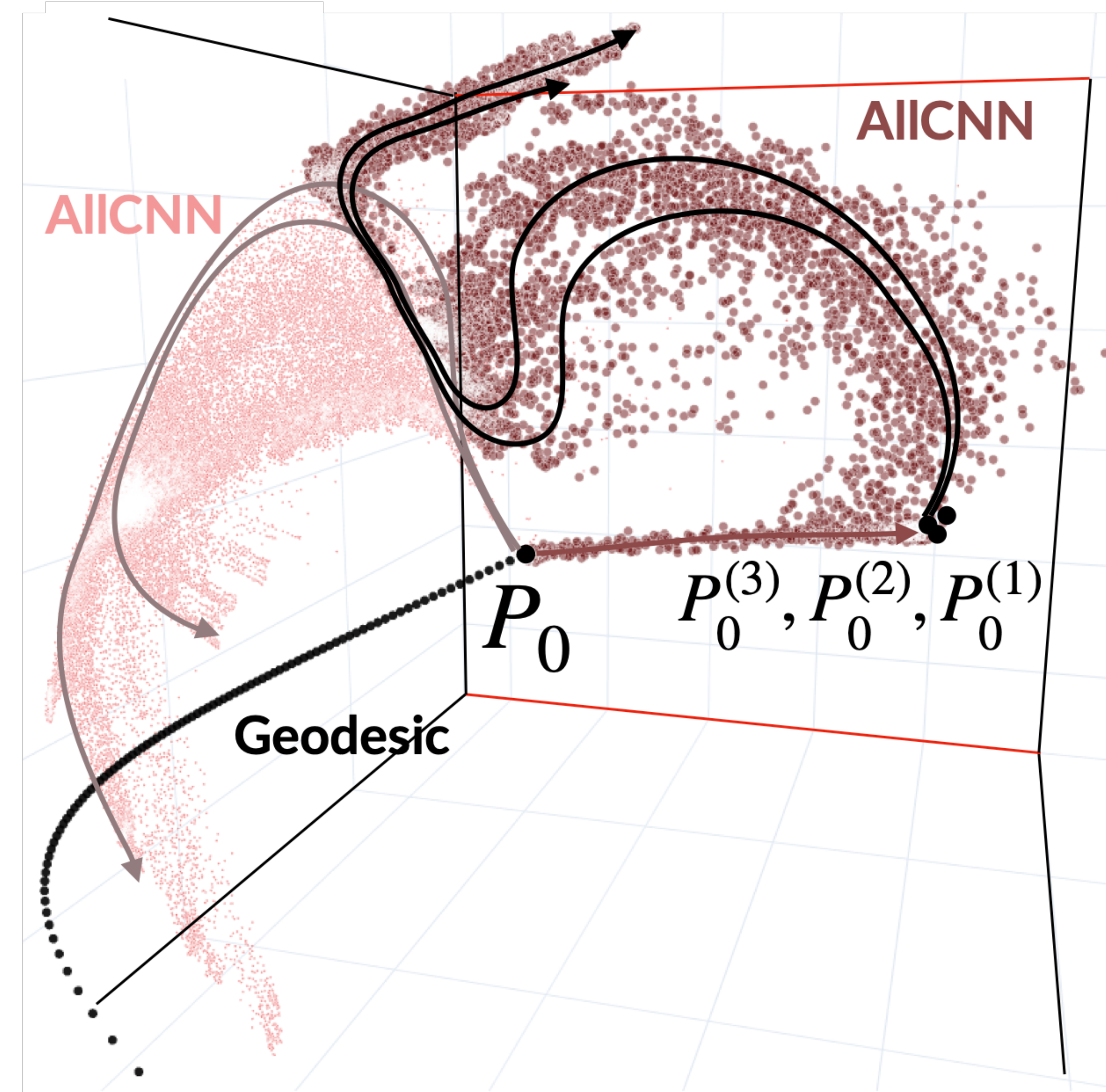}
\caption{Predictions on test set of a subset of AllCNN models (the same set as in~\cref{fig:allcnn_from_corners}) trained from ignorance (light brown) and from three different corners (dark brown). Networks trained from corners still seem to come close to the normally trained models mid-training, but they divert from the main manifold and end at a higher testing error in the later part of training.}
\label{fig:allcnn_test_from_corners}
\end{wrapfigure}

For the same error, models on the test manifold show a large spread (see~\cref{fig:all_models_test_2d_error}) as compared to those on the train manifold in~\cref{fig:all_models_train_2d_error}. In particular, different ConvMixer networks which eventually reach low test errors predict similarly at intermediate levels of train/test error, not only on the training data but also on the test data (blue/purple in~\cref{fig:all_models_test_2d_spread}). But fully-connected networks predict very differently from each other at intermediate errors (error of, say 0.3--0.4 in~\cref{fig:all_models_test_2d_error}), i.e., their spread is more pronounced on the test manifold. This could indicate that architectures with strong inductive biases (e.g., convolutions) explore a smaller part of the prediction space, even on the test data. It has implications for theoretical analyses of generalization in deep learning using ideas such as algorithmic stability.

Using PC2 and PC3, in~\cref{fig:all_models_test_2d_branches}, we chose five specific endpoints, corresponding to fully-connected and ViT networks trained with and without augmentation (B--E), and for comparison, one more endpoint from the trajectory of ConvMixer trained with augmentation (A). We colored models in terms of whether they lie within a Bhattacharyya distance \textless\ 0.45 from their closest center. Models colored purple are far from all centers. For fully-connected and ViT networks, models having the same test error can lie in very different parts of the test manifold. For example, for test error within 0.3--0.4 (see~\cref{fig:all_models_test_2d_error}) some models lie on the manifold (e.g., green in~\cref{fig:all_models_test_2d_spread}), some on one branch (e.g., one of the purple branches or the smaller green branch in~\cref{fig:all_models_test_2d}), while some others can lie on other branches (e.g., other purple branches in~\cref{fig:all_models_test_2d}).

\paragraph{Models initialized at very different parts of the prediction space converge to the truth along a similar manifold}

For the test data, there is a larger spread in how models initialized near $P_0^{(k)}$ join the main manifold, and also how their endpoints are different from endpoints of trajectories that begin near ignorance $P_0$ (see~\cref{fig:allcnn_test_from_corners}).

\subsection{Observations remain consistent with other intensive distances}
\label{s:app:iskl}

We can also use other distances in place of the Bhattacharyya distance. For example, the IsKL method~\cite{teohVisualizingProbabilisticModels2020} uses the symmetrized Kullback-Leibler (KL) divergence to compute the distances between pairs of points $D$ in~\cref{eq:w}
\beq{
    \dsKL(P_u, P_v) = \f{1}{N} \sum_{n=1}^N \sum_{c=1}^C \rbr{p^n_u(c) - p^n_v(c)} \log \rbr{\f{p^n_u(c)}{p^n_v(c)}}.
    \label{eq:isKL}
}
For exponential families, we can obtain an analytical formula for the IsKL embedding and in this case, the embedding has at most twice the number of dimensions as the dimensionality of the sufficient statistic (for CIFAR-10, this has $9 \times 10^5$ dimensions). Our models $P_u$ and $P_v$ are vectors that lie on a sphere of radius $N$ (probabilities of each image sum up to 1). We could also use the geodesic distance on this sphere
\beqs{
    \sqrt{N} \cos^{-1} \prod_{n=1}^N \sum_{c=1}^C \sqrt{p_u(y_n)} \sqrt{p_v(y_n)};
}
but this has poor behavior in high dimensions because points along the trajectory jump abruptly from ignorance to truth. This is similar to the saturation of the Hellinger distance in high dimensions that is discussed in the main text. Since our models live on a product space of hyper-spheres (samples in the dataset are independent of each other) we can use the geodesic distance on the product of spheres instead
\beq{
    \dG(P_u, P_v) = \f{1}{N} \sum_n \cos^{-1} \sum_c \sqrt{p_u^n(c)} \sqrt{p_v^n(c)}.
    \label{eq:dG}
}

\begin{wrapfigure}{r}{0.5\linewidth}
\centering
\includegraphics[width=\linewidth]{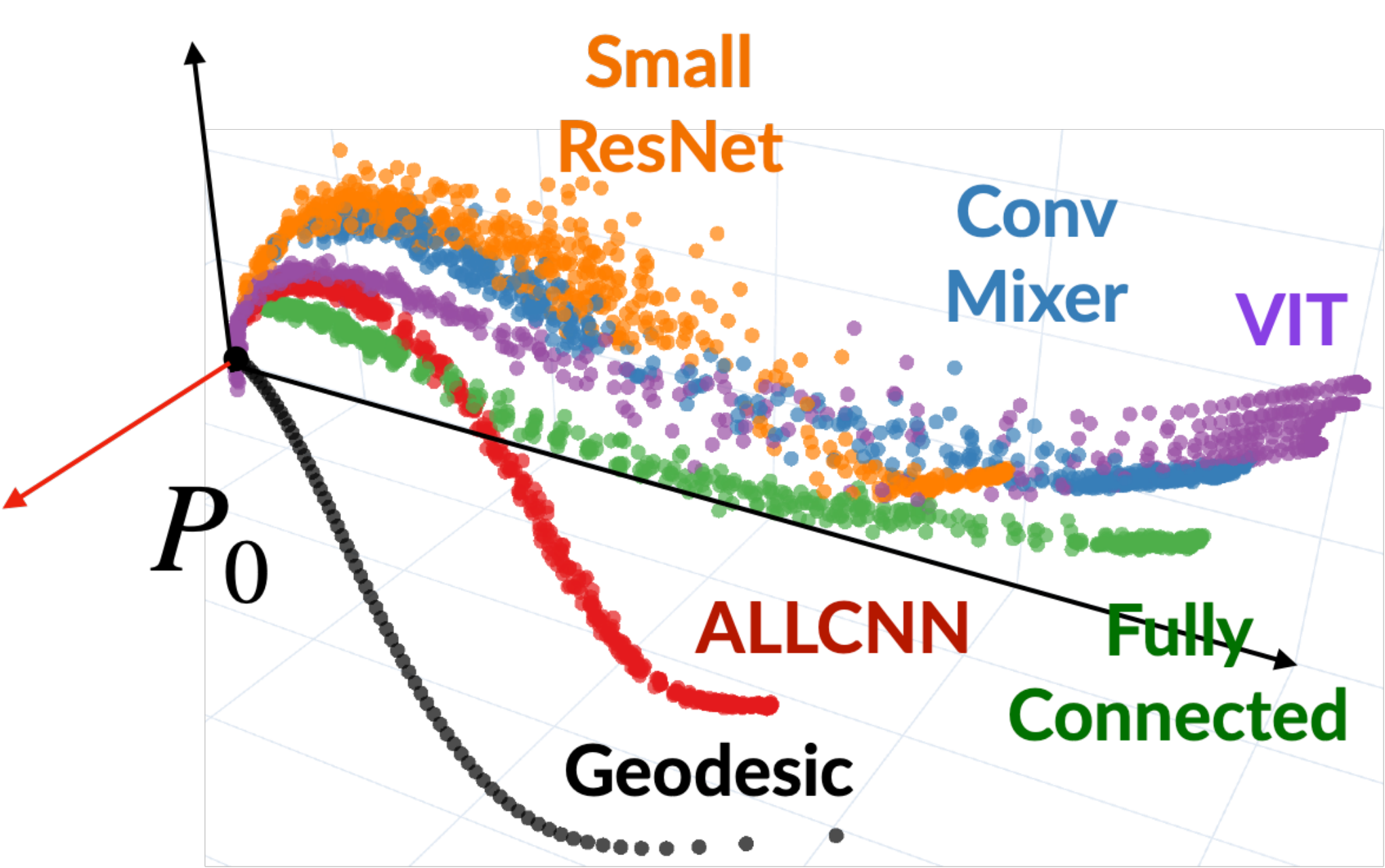}
\caption{The top three dimensions of the IsKL embedding using the train data for a subset of the models trained on CIFAR-10 (this is the same subset as in~\cref{fig:train_test_inpca_3d}). The IsKL embedding carries a different kind of information than the InPCA embedding in~\cref{fig:all_models_train_3d,fig:all_models_test_3d}. Trajectories exhibit a larger spread towards the end of training and truth $P_*$ (not seen here) is at infinity. The IsKL embedding emphasizes the differences among the trajectories towards the end of training.}
\label{fig:iskl_3d}
\end{wrapfigure}

All the above distances respect the natural Fisher Information Metric in probability space. The IsKL, InPCA and Geodesic embeddings carry different pieces of information on the structure of the space of probability distributions. For example, IsKL places truth $P_*$ infinitely far away, and it therefore stretches the last part of the training trajectories in our experiments. This allows us to investigate the behavior of trajectories towards the end of training in more detail (although we do not do so in this paper). We have noticed in smaller-scale experiments that IsKL captures a slightly higher explained stress in the top three dimensions that InPCA. The geodesic embedding maps geodesics to straight lines which may be useful to construct a simpler, more interpretable, set of coordinates.

\paragraph{Embeddings using standard principal component analysis (PCA)}

\begin{figure}
\centering
\begin{subfigure}[b]{0.4\linewidth}
\centering
\includegraphics[width=\linewidth]{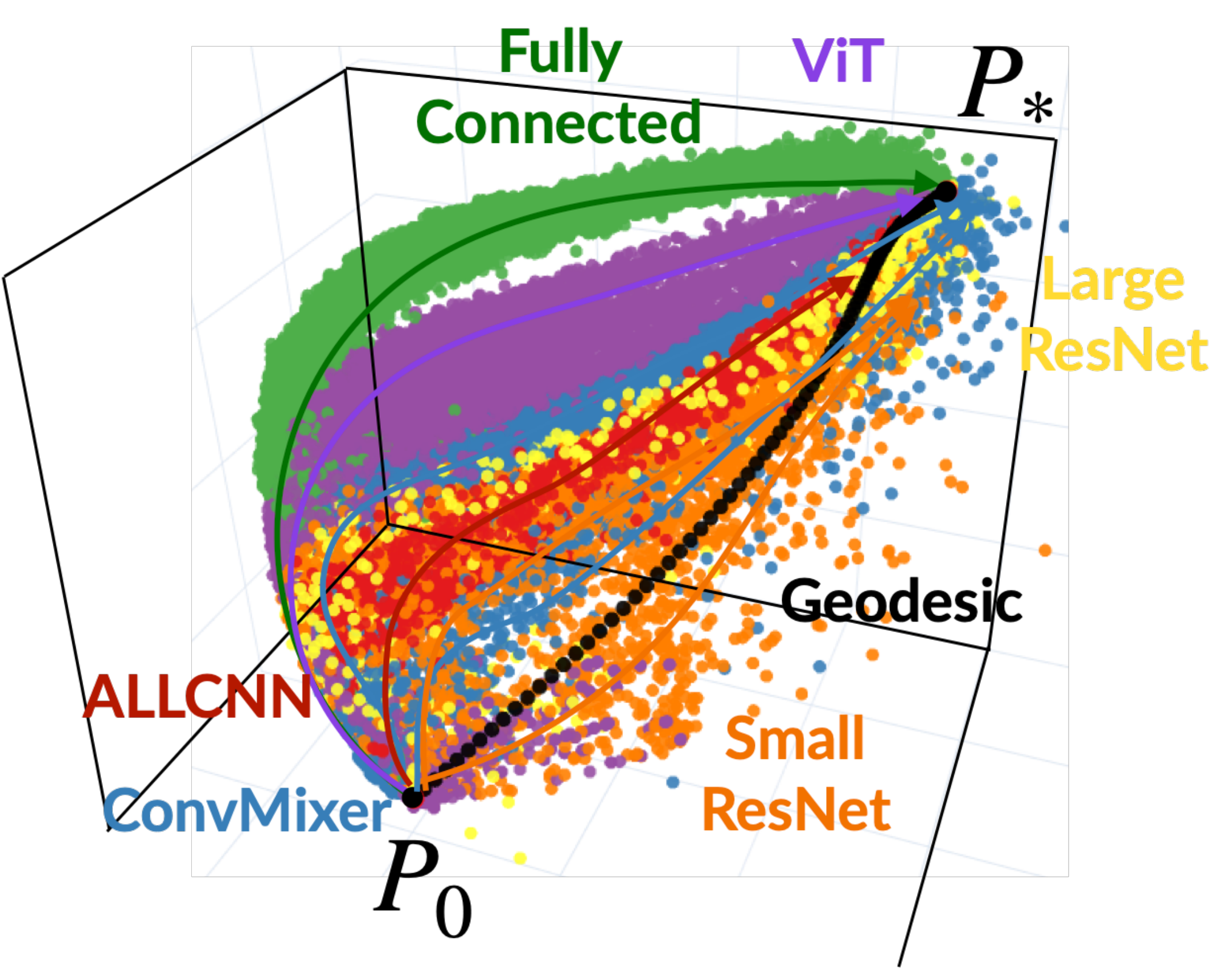}
\caption{}
\label{fig:all_models_train_3d_pca}
\end{subfigure}%
\begin{subfigure}[b]{0.14\linewidth}
\centering
\includegraphics[width=\linewidth]{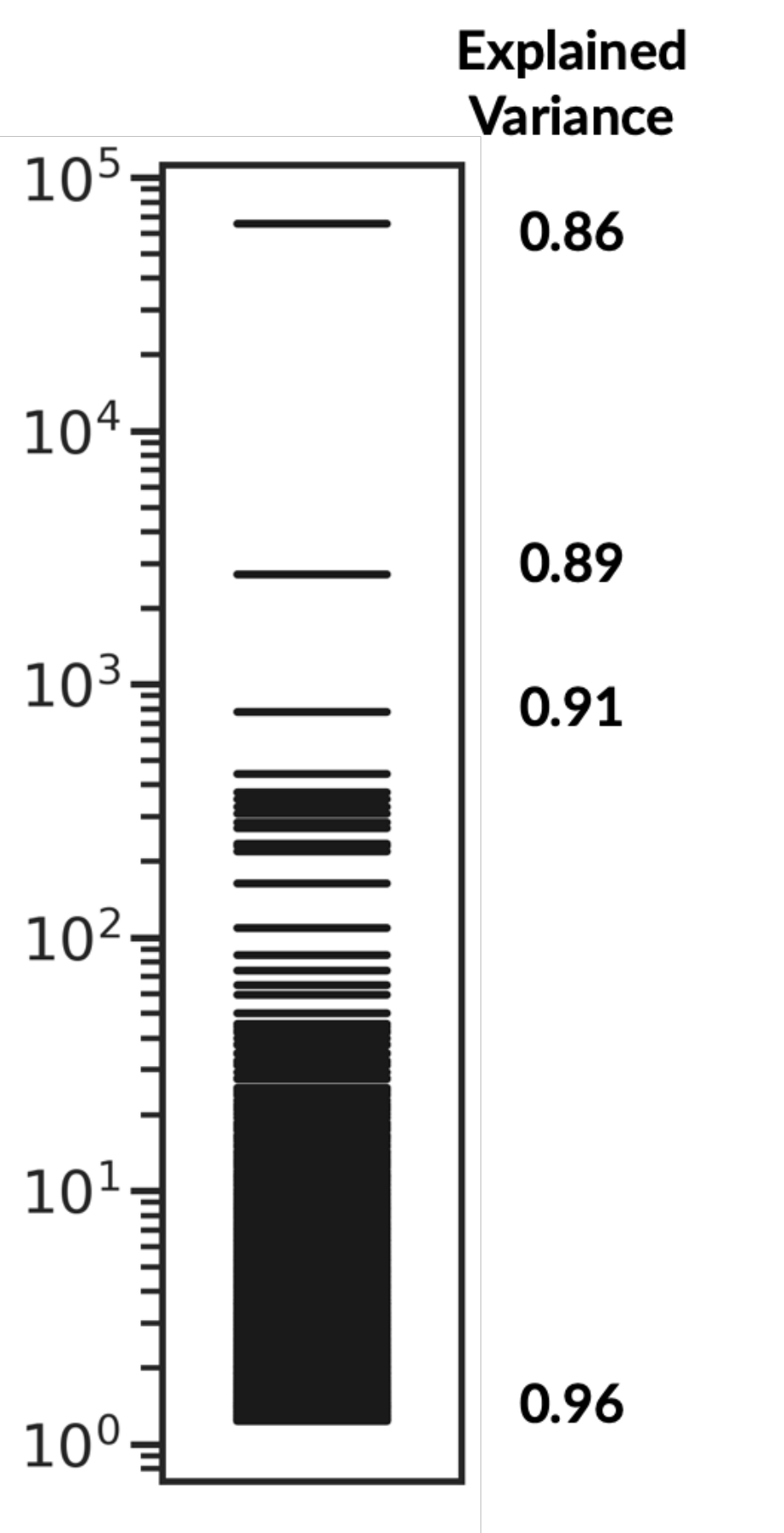}
\caption{}
\label{fig:all_models_train_3d_pca_eig}
\end{subfigure}
\begin{subfigure}[b]{0.3\linewidth}
\centering
\includegraphics[width=\linewidth]{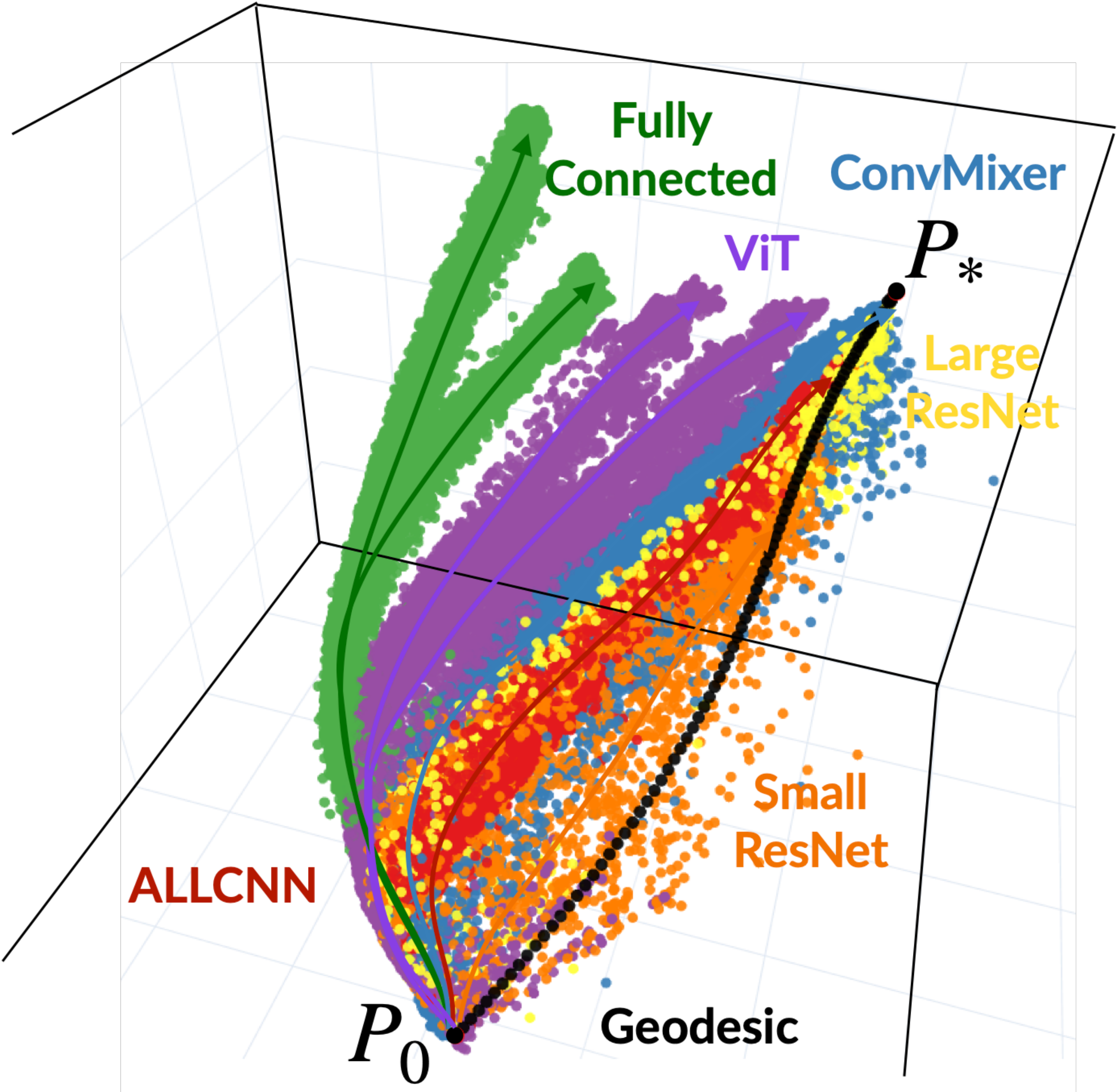}
\caption{}
\label{fig:all_models_test_3d_pca}
\end{subfigure}
\begin{subfigure}[b]{0.14\linewidth}
\centering
\includegraphics[width=\linewidth]{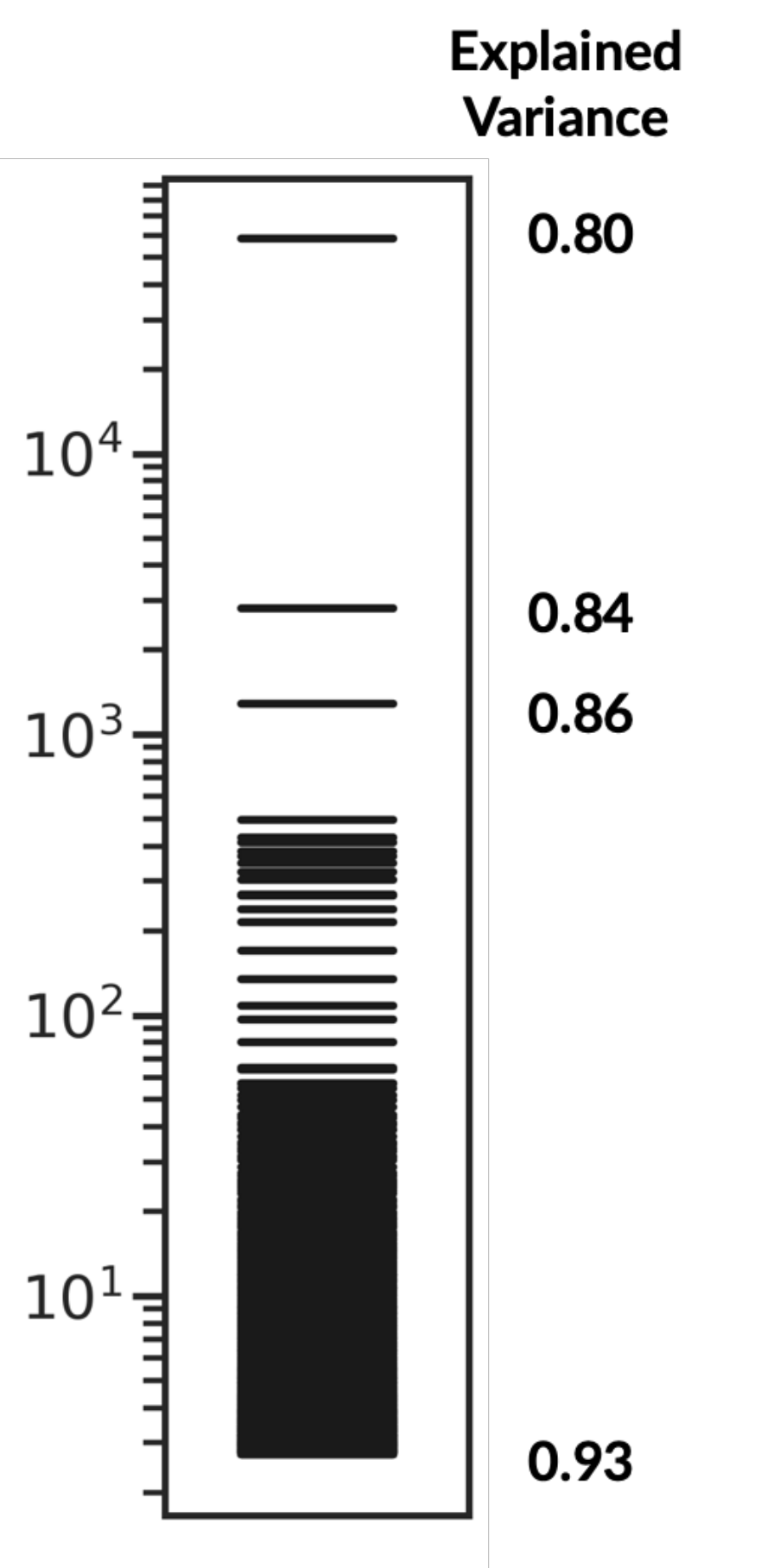}
\caption{}
\label{fig:all_models_test_3d_pca_eig}
\end{subfigure}
\caption{The top three dimensions of an embedding obtained using standard PCA for all the networks on CIFAR-10 using train data \textbf{(a)} and the test data \textbf{(c)}. The explained variance in \textbf{(b,d)} for train and test data respectively is very high but the structure of the low-dimensional manifold identified by PCA is very different from that obtained by InPCA in~\cref{fig:all_models_train_3d,fig:all_models_test_3d}. In particular, although this embedding is low-dimensional it does not respect the natural metric in probability space because the second derivative of the divergence is not the same Fisher Information Matrix as that of, say, the Bhattacharya distance.}
\label{fig:all_models_3d_pca}
\end{figure}

Our data consists of probability distributions and therefore a meaningful embedding of such data should seek to preserve distances between probability distributions. But it is reasonable to ask how well standard dimensionality reduction and embedding techniques, e.g., standard principal component analysis (PCA), can reveal the inherent low-dimensional structure in the data. For this calculation, we created a matrix of pairwise distances
\[
    D_{uv} = \f{1}{N} \sum_n \sum_c \rbr{p^n_u(c) - p^n_v(c)}^2
\]
and computed the eigen-decomposition of this matrix (after centering) to get the coordinates. One should note two important choices here: (a) the Euclidean distance between the probability distributions $p^n_u(\cdot)$ and $p^n_v(\cdot)$ treats them as standard vectors in $\reals^C$, and (b) the averaging over the samples using $N^{-1}$ ensures that $D_{uv}$ remains non-trivial even for a large number of samples.

We show an embedding calculated using PCA for the train and test manifolds in~\cref{fig:all_models_train_3d_pca} and~\cref{fig:all_models_test_3d_pca} respectively. In both cases, an embedding using PCA suggests that the data lies on an effectively low-dimensional manifold, the explained variance is quite large (91\% and 86\% in the first three dimensions for train and test manifolds respectively). This is consistent with the results we have discussed using InPCA in the main text. But because it uses an unusual distance between probability distributions, PCA distorts the structure of the manifold as compared to InPCA. The salient differences are as follows: (a) trajectories corresponding to different architectures are very close to each other in~\cref{fig:all_models_train_3d} and~\cref{fig:dendrogram_train_end} but there are marked differences in these trajectories in~\cref{fig:all_models_train_3d_pca}; (b) the geodesic is far from all trajectories in the original data but this is not so in the PCA embedding; (c) the cloud of points that lie away from the main manifold, which we have analyzed in~\cref{fig:all_models_train_2d_details}, is not visible in the PCA embedding. For the test manifold, we see some similarities between~\cref{fig:all_models_test_3d,fig:all_models_test_3d_pca}: (a) there are  multiple branches for fully-connected and ViT networks; and (b) networks that obtain good test error (ConvMixer and Large ResNet) are closer to the truth. There are also some differences: (a) the geodesic is far from all trajectories in the InPCA embedding while it is close to the trajectories of the Small ResNet in the PCA embedding; (b) InPCA reveals the fact that trajectories of AllCNN are closest to the geodesic in terms of the Bhattacharyya distance for both train and test manifolds (\cref{fig:all_models_d2geod}) but PCA does not show this.

Altogether, while we can corroborate the claim that the trajectories explore an effectively low-dimensional manifold of predictions on both the train and test data using both methods, PCA distorts the structure of the manifold and conclusions that one may derive from the embedding are not consistent with those derived from analysis of the trajectories in the original high-dimensional space. Also, observe that InPCA distinguishes the small differences between the probability distributions towards the end of training while PCA does not.

\begin{table}
\centering
\resizebox{0.6\linewidth}{!}{
\renewcommand{\arraystretch}{1.5}
\begin{tabular}{c c r}
\toprule
Divergence $d(p,q)$ & \multicolumn{2}{c}{Centroid $(p^{(1)}, p^{(2)}, \dots)$ }\\
\midrule
$\sum_n (p_n-q_n)^{2}$                             & $\propto \sum_k p^{(k)}_n$ & Arithmetic mean (AM)\\
$\sum_n (\sqrt{p_n} - \sqrt{q_n})^2$      & $\propto \sum_k \sqrt{p^{(k)}_n}$ & Sqrt. Arithmetic mean\\
$\sum_n (\log p_n - \log q_n)$                  & $\propto (\prod_k p^{(k)}_n)^{1/N}$ & Geometric mean (GM)\\
$\sum_n n (p_n^{-1} - q_n^{-1})$                   & $\propto (\sum_k 1/p^{(k)}_n)^{-1}$ & Harmonic mean (HM)\\
$ -\log \rbr{\sum_n \sqrt{p_n} \sqrt{q_n}}$         & & Bhattacharyya centroid~\cite{nielsen2011burbea}\\
$ \sum_n (p_n - q_n)  \log (p_n/q_n)$  & $\text{AM}/W \rbr{e\text{AM}/\text{GM}}$ & Jeffrey's centroid~\cite{nielsen2013jeffreys}\\
\bottomrule
\end{tabular}
}
\caption{\textbf{Different divergences and their corresponding centroids.} \textnormal{We have showed the formulae for two $N$-dimensional probability distributions $(p_n)_{n=1,\dots,N}$ and $(q_n)_{n=1,\dots,N}$ and the centroid of a set of distributions $\{p^{(1)}, p^{(2)}, \dots \}$. The Lambert omega function is denoted by $W(\cdot)$ and $e$ is Euler's number.}}
\label{tab:mean_calculation}
\end{table}

\begin{figure}
\centering
\begin{subfigure}[b]{0.4\linewidth}
\centering
\includegraphics[width=\linewidth]{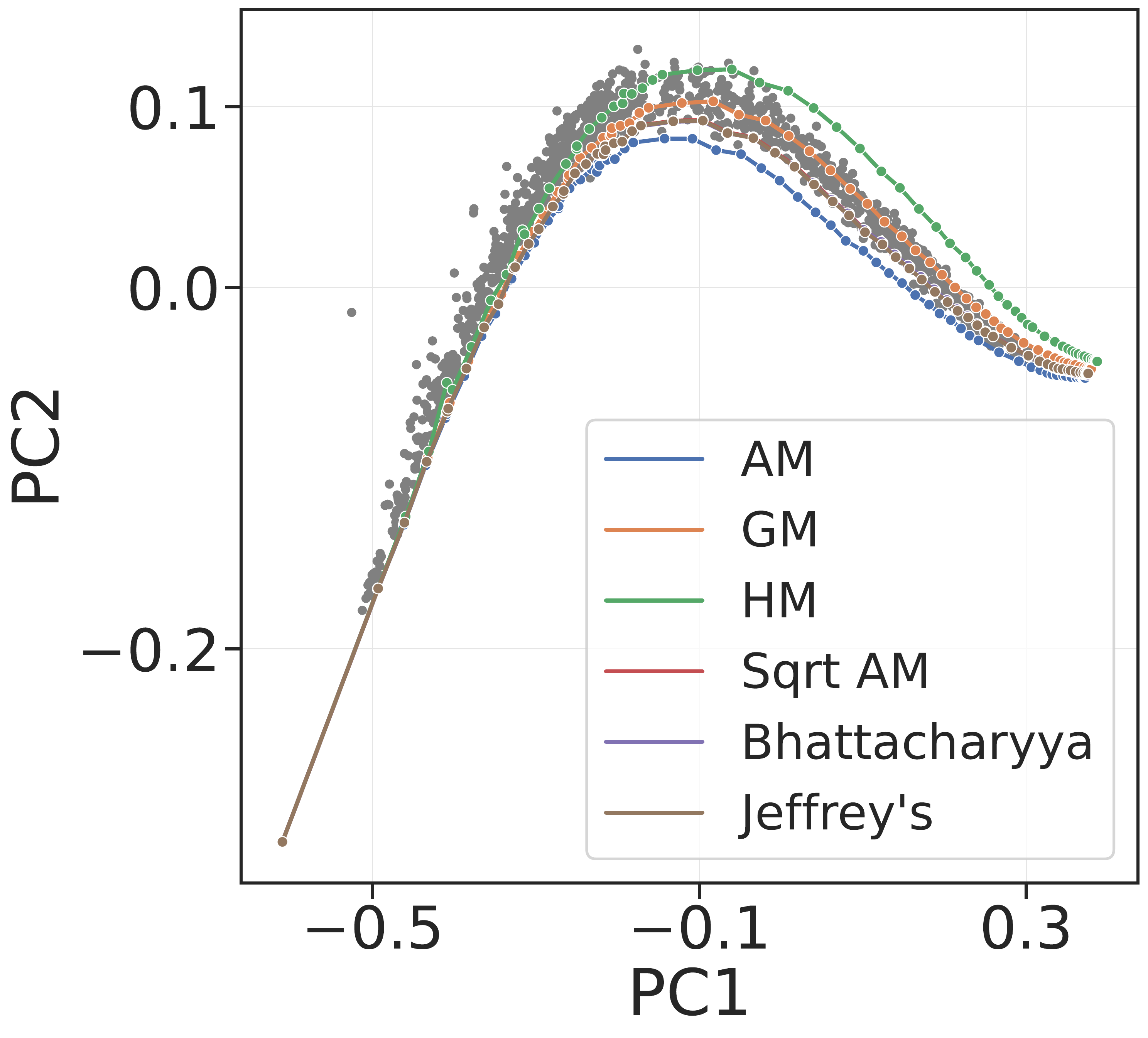}
\caption{}
\label{fig:allcnn_train_mean}
\end{subfigure}
\hspace*{5ex}
\begin{subfigure}[b]{0.4\linewidth}
\centering
\includegraphics[width=\linewidth]{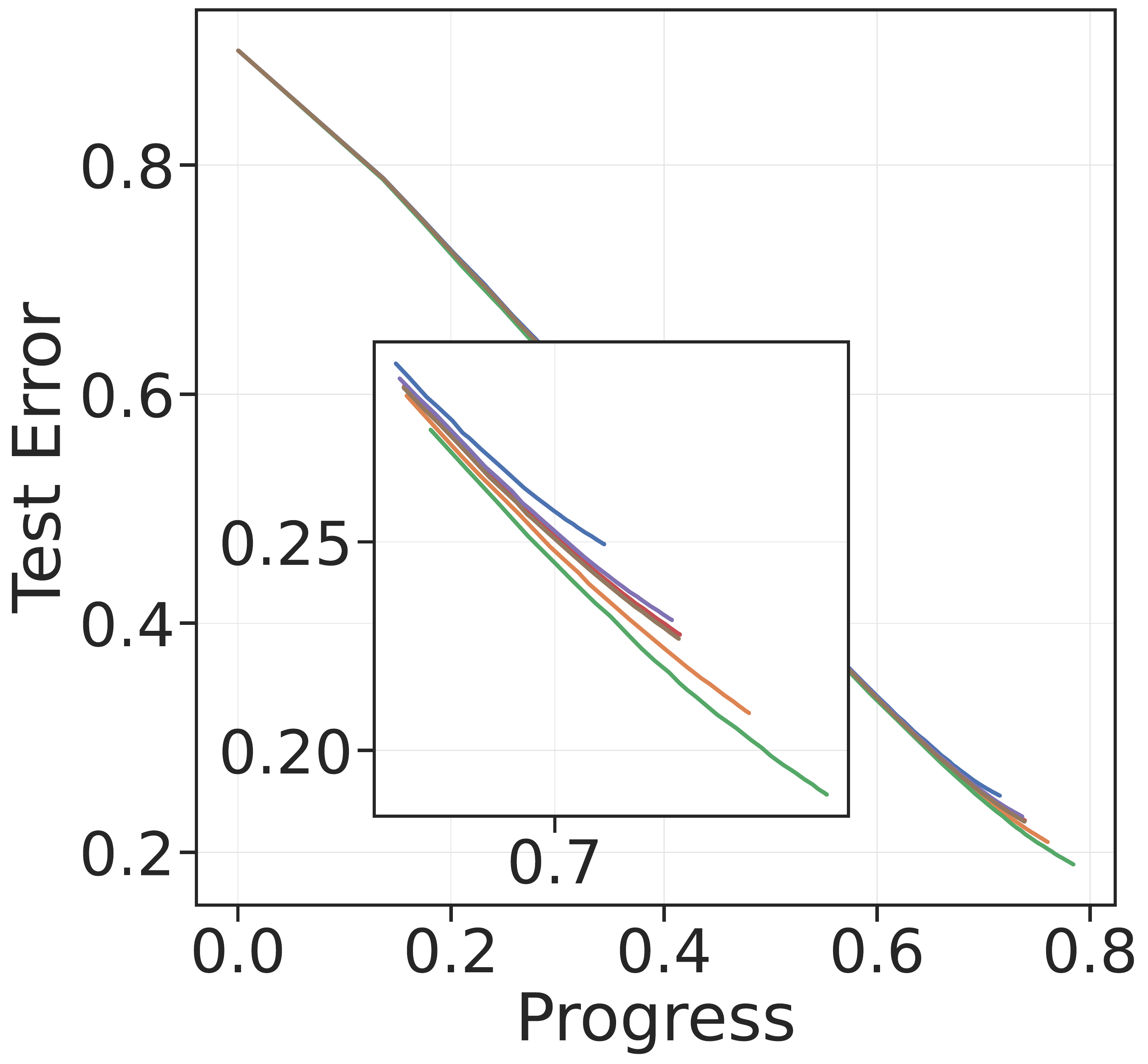}
\caption{}
\label{fig:allcnn_test_mean_progress}
\end{subfigure}
\caption{
\textbf{(a)}: the top two principal components obtained from InPCA for the train data for one particular configuration on CIFAR-10 (AllCNN architecture, trained with SGD without augmentation or weight-decay). We computed the arithmetic mean (AM), geometric mean (GM), harmonic mean (HM), the arithmetic mean of the square roots of probabilities appropriately normalized (Sqrt AM), the Bhattacharya centroid and Jeffrey's centroid for models with the same progress. It is noticeable that different means do not always lie on the manifold. In particular, the arithmetic mean and the harmonic mean are the farthest away visually. \textbf{(b)}: the test error as a function of progress for the different ways of computing the mean. The test errors are AM (25.0\%), GM (20.9\%), HM (18.9\%), Sqrt AM (22.8\%), Bhattacharyya centroid (23.1\%), Jeffrey's centroid (22.7\%): therefore computing the harmonic mean of the probabilities of the models in the ensemble leads to a slightly better test error than computing the arithmetic mean of their probabilities which is typically done in machine learning.}
\label{fig:allcnn_mean}
\end{figure}

\subsection{Harmonic mean of an ensemble of deep networks has a better test error}
\label{s:app:harmonic_mean}

We saw previously that a small network with higher eventual test error trains along the same manifold as that of a large network with lower eventual test error, more slowly. There is a classical technique that also achieves better test errors, namely ensembling. We therefore investigated whether an ensemble also exhibits higher progress towards the truth than that of the individual models that constitute the ensemble.

The standard way of building an ensemble in machine learning is to calculate the arithmetic mean of the class probabilities; this corresponds to the $\ell_2$ distance in the space of probability distributions. As~\cref{tab:mean_calculation} shows, different distances correspond to different ways of computing the centroid. We choose five other candidates: (i) the arithmetic mean of the square roots of the probabilities, which corresponds to the centroid of the Hellinger distance, (ii) the geometric mean, (iii) the harmonic mean, (iv) the centroid of the Bhattacharyya distance, which can be calculated using an iterative procedure given in~\cite{nielsen2011burbea}, and (v) Jeffrey's centroid which corresponds to the symmetric KL-divergence which is known in closed-form~\cite{nielsen2013jeffreys}. In~\cref{fig:allcnn_mean}, for 30 different weight initializations, for both train and test trajectories pertaining to one particular configuration (AllCNN architecture, trained with SGD without augmentation or weight-decay), we show these different centroids, after the same number of mini-batch updates for each model.

\begin{wrapfigure}{r}{0.4\linewidth}
\centering
\includegraphics[width=\linewidth]{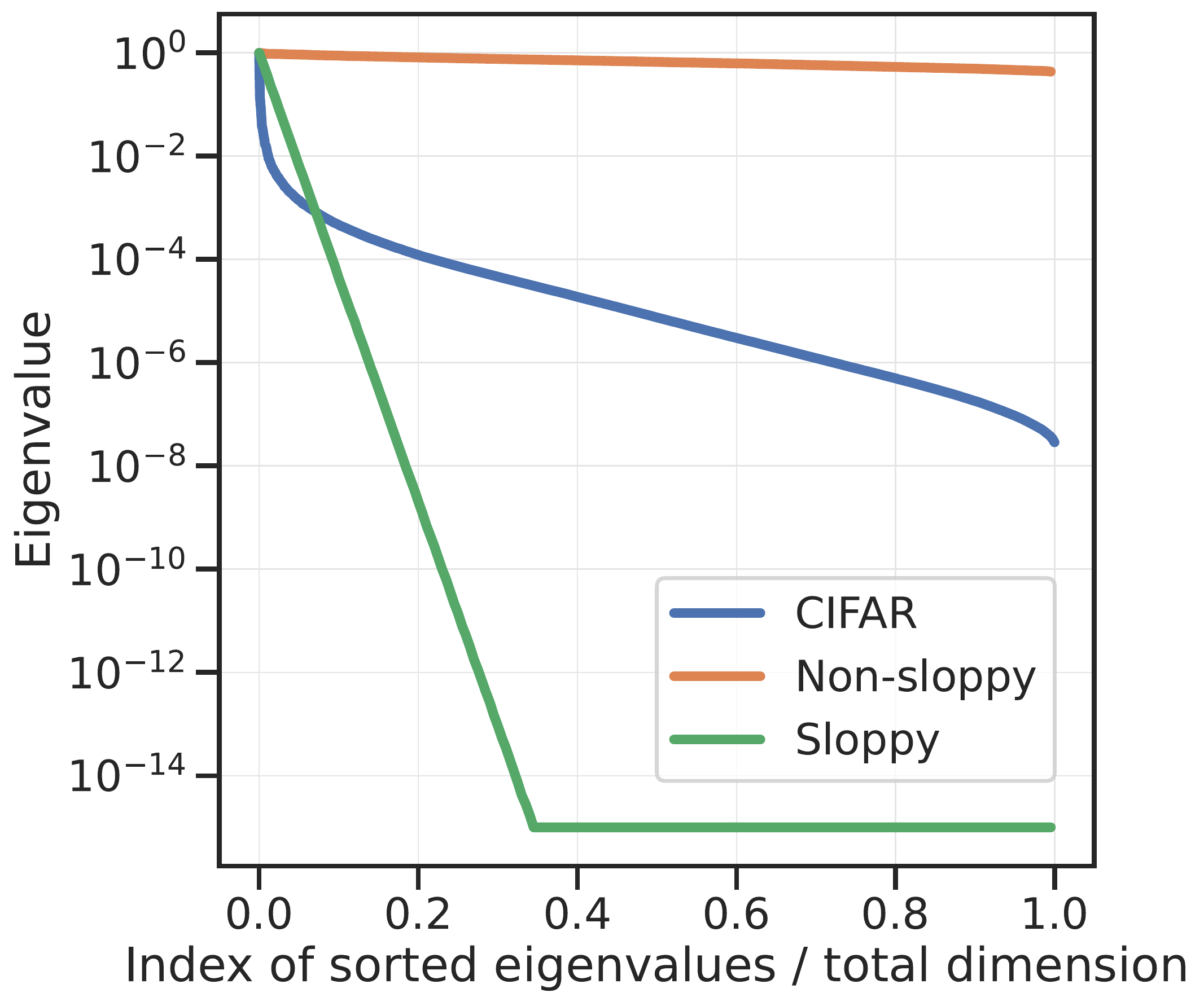}
\caption{Eigenvalues of the the input correlation matrix $\E[x x^\top]$ for 32$\times$32 RGB images $x$ in CIFAR-10 (blue), i.e., $x \in \reals^{3072}$, non-sloppy synthetic inputs ($x \in \reals^{200}$) sampled from an isotropic zero-mean Gaussian (orange) and sloppy synthetic inputs ($x \in \reals^{200}$) sampled from a Gaussian distribution with zero mean and covariance matrix whose eigenvalues decay as $\l_i = 50 c \exp(-c i)$ for $c=0.5$ (green).}
\label{fig:input_evals_all}
\end{wrapfigure}

The arithmetic mean lies noticeably outside the manifold in the visualization for both train and test manifolds. Different centroids have different trajectories in the embedding. But the harmonic mean (green) makes the highest progress towards the truth on the test manifold and also has the lowest test error at the end~\cref{fig:allcnn_test_mean_progress}. This suggests that ensembles that use the harmonic mean of the probabilities to compute the final model could lead to a slightly better test error.

\section{Experiments using synthetic data}
\label{s:app:details_synthetic}

\paragraph{Datasets}
We sampled $N=5000$ samples for the training set and $N = 1000$ samples for the test set from a $d=200$ dimensional Gaussian with mean zero and a diagonal covariance $\L = \diag(\l_1, \ldots, \l_d)$. We experimented with two types of data: those sampled from an isotropic Gaussian ($\L = I_{d \times d}$) and those sampled from a Gaussian distribution with a covariance matrix that has eigenvalues that decay linearly on a logarithmic scale, i.e., $\l_i = 50 c e^{-ci}$. The latter setup is the so-called sloppy dataset~\cite{yang2021does,transtrumGeometryNonlinearLeast2011}. We can control the sloppiness of the dataset by choosing different values of $c$, i.e., larger the value of $c$, sharper the decay. We created a 5-class classification problem using labels from a teacher  (a fully-connected network with one hidden layer of width 50). The largest logit among the 5 logits of the teacher is taken to be the ground-truth label. We train student networks of different architectures using these teacher-generated labels using the cross-entropy loss.  All networks were trained with batch-normalization and without dropout.

\paragraph{Neural architectures and training procedure}
We studied the difference in training trajectories when networks are trained on data with different sloppiness. We used two values: $c = 0.001$ (which is effectively non-sloppy data) and $c = 0.5$ (which is sloppy data). We trained 160 different configurations: (1) fully connected networks of one and two hidden layers (both with a width of 512), (2) training with SGD and SGD with Nesterov's momentum of coefficient 0.9, (3) two values of batch-size 200 and 500, (4) two values of the weight decay coefficient $\{0, 10^{-5}\}$, and (5) 10 different weight initializations.

\begin{figure}
\centering
\begin{subfigure}[b]{0.35\linewidth}
\centering
\includegraphics[width=\linewidth]{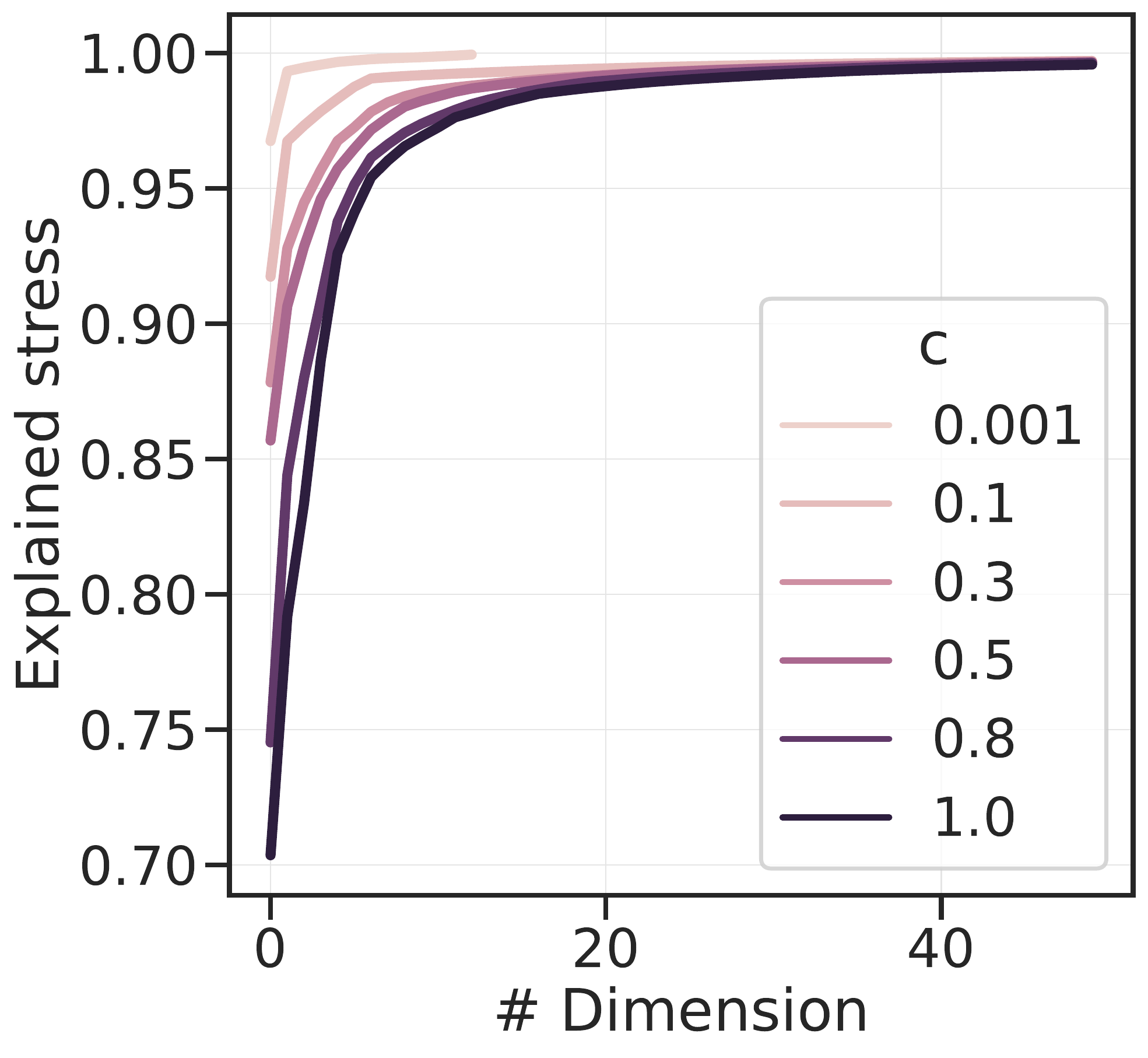}
\caption{}
\label{fig:synthetic_explained_stress_c}
\end{subfigure}
\hspace*{3ex}
\begin{subfigure}[b]{0.55\linewidth}
\centering
\includegraphics[width=\linewidth]{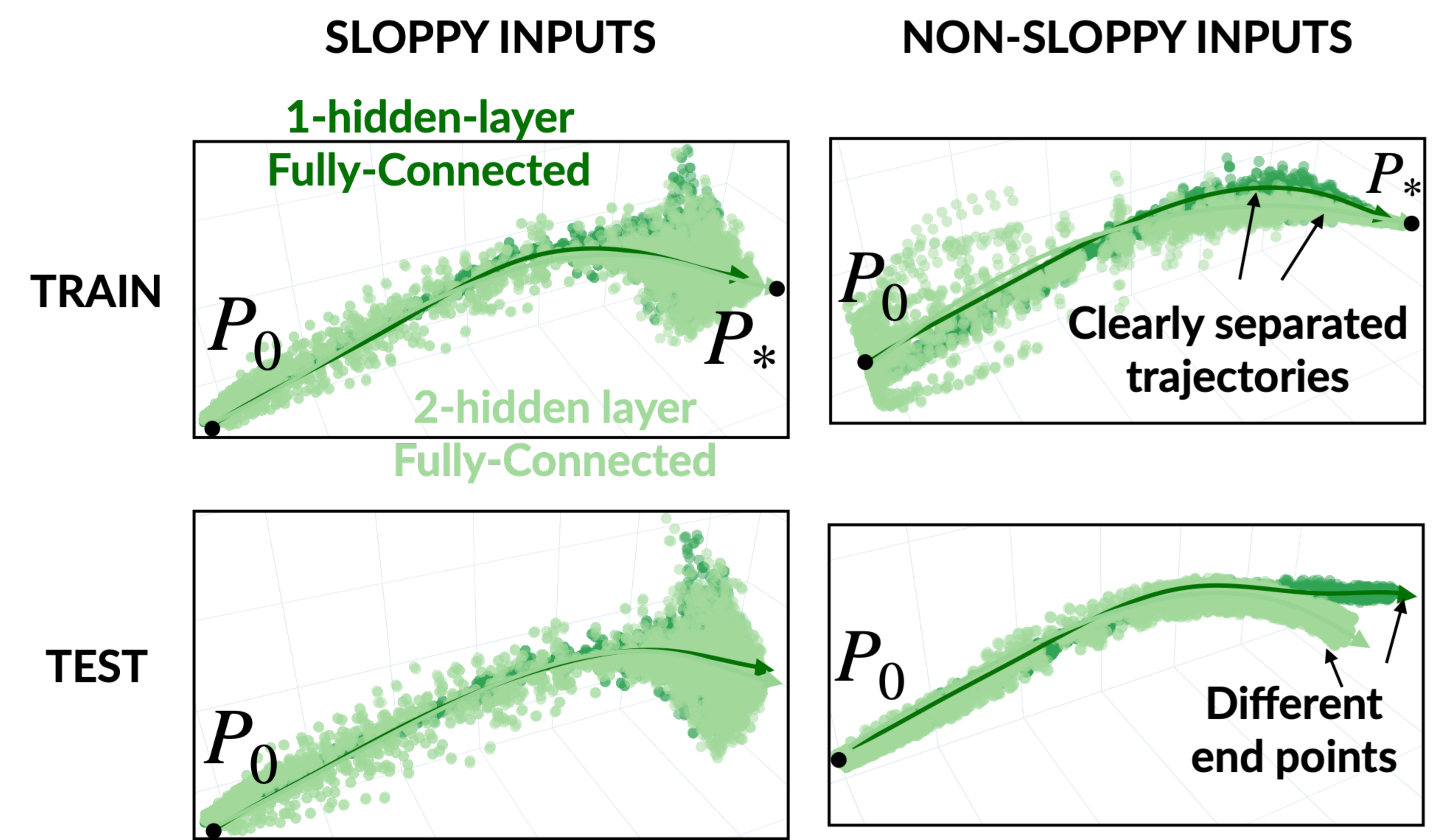}
\caption{}
\label{fig:synthetic_from_ignorance_panel}
\end{subfigure}
\caption{
\textbf{(a)}: the explained stress in the top few dimensions (X-axis) of an InPCA embedding of models along training trajectories when input data are sampled from a Gaussian distributions with zero mean and covariance matrix whose eigenvalues decay as $\l_i = 50 c \exp(-c i)$ for different values of $c$. For all values of $c$ (small values indicate that inputs were sampled from a near-isotropic Gaussian and large values indicate that input data were sampled from a Gaussian with a sloppy covariance matrix), the explained stress is high.
\textbf{(b)}: the top three dimensions of an InPCA embedding of models along train and test trajectories for synthetic sloppy and non-sloppy input data for two different architectures (1-hidden-layer fully-connected networks in dark green and 2-hidden-layer fully-connected networks in light green) and multiple training configurations for each architecture.
}
\label{fig:synthetic_from_ignorance}
\end{figure}

\paragraph{Analysis}
Train and test manifolds are effectively low-dimensional for both sloppy and non-sloppy data. \cref{fig:synthetic_explained_stress_c} shows how the explained stress increases in the top few dimensions of the InPCA embedding; it reaches 99\% in the first 10 dimensions of an InPCA embedding. In general, when inputs are sloppy (larger value of $c$ is more sloppy inputs), the explained stress is slightly lower. We speculate that this is due to the increased difficulty of the underlying optimization problem which makes the details of the optimization procedure, e.g., the learning rate, important---and thereby leads to a larger spread in the models of different configurations. The explained stress on test data is essentially the same. As the embeddings in~\cref{fig:synthetic_from_ignorance} show qualitatively, when input data is not sloppy, training trajectories show a more clear separation between different training configurations. It is therefore important to choose the architecture (in this case) when we fit models on non-sloppy data. On the other hand, if input data is sloppy, choosing the architecture or the parameters of the optimization algorithm carefully is less important. We noticed that the larger spread of the points in the InPCA embedding towards the end of training near $P_*$ in~\cref{fig:synthetic_from_ignorance_panel} is coming from models trained with SGD with Nesterov's acceleration. A heuristic explanation of this phenomenon, using a linear regression objective for sloppy vs.\@ non-sloppy data,  is that overshoots in the weight space caused by momentum terms in Nesterov's acceleration lead to more diverse trajectories if the underlying objective is not isotropic.

\begin{figure}[b]
\centering
\begin{subfigure}[t]{0.245\linewidth}
\centering
\includegraphics[width=\linewidth]{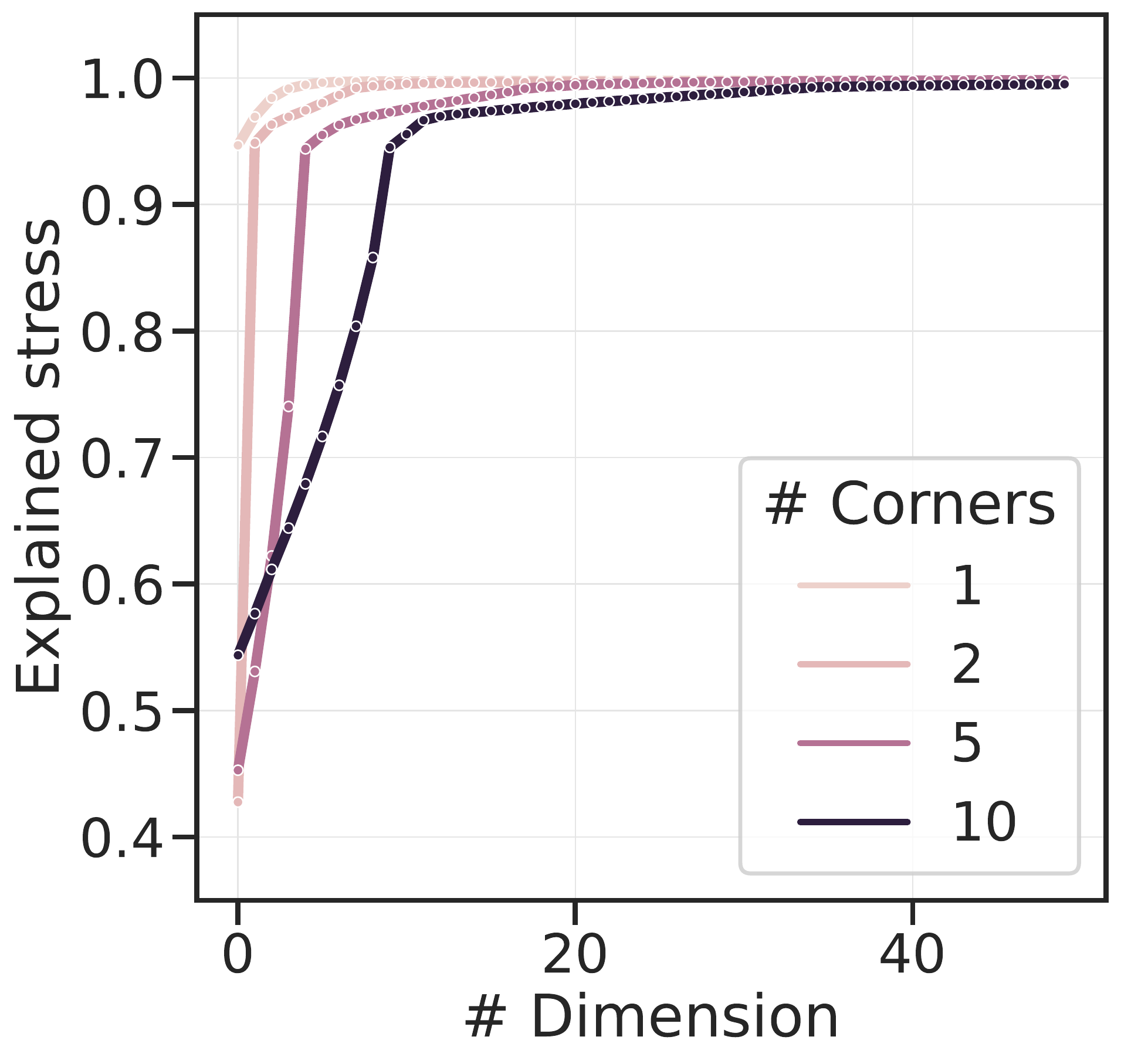}
\caption{Non-sloppy input data}
\label{fig:synthetic_corners_explained_stress_nonsloppy}
\end{subfigure}
\begin{subfigure}[t]{0.245\linewidth}
\centering
\includegraphics[width=\linewidth]{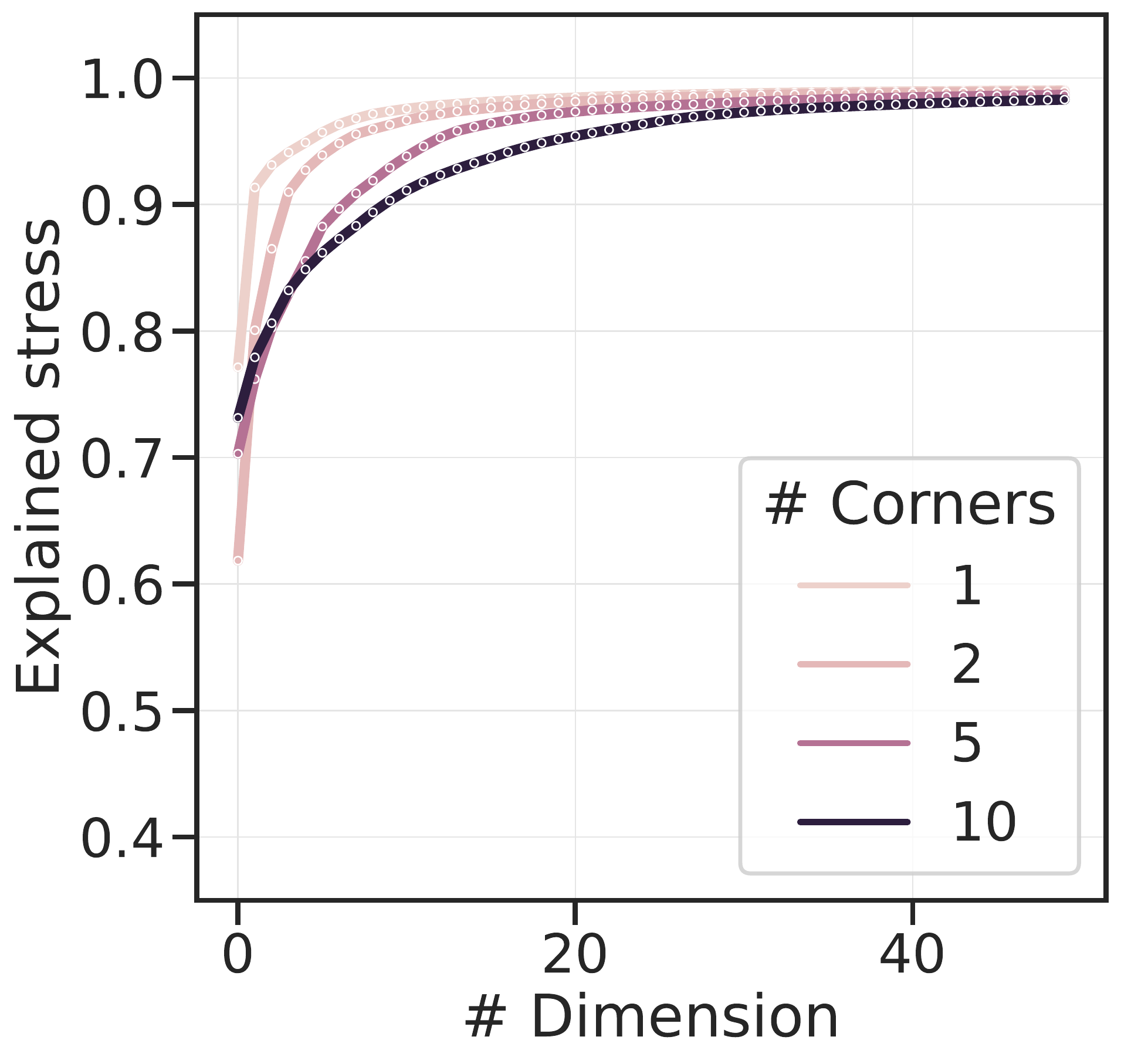}
\caption{Sloppy input data}
\label{fig:synthetic_corners_explained_stress_sloppy}
\end{subfigure}
\hfill
\begin{subfigure}[t]{0.45\linewidth}
\centering
\includegraphics[width=\linewidth]{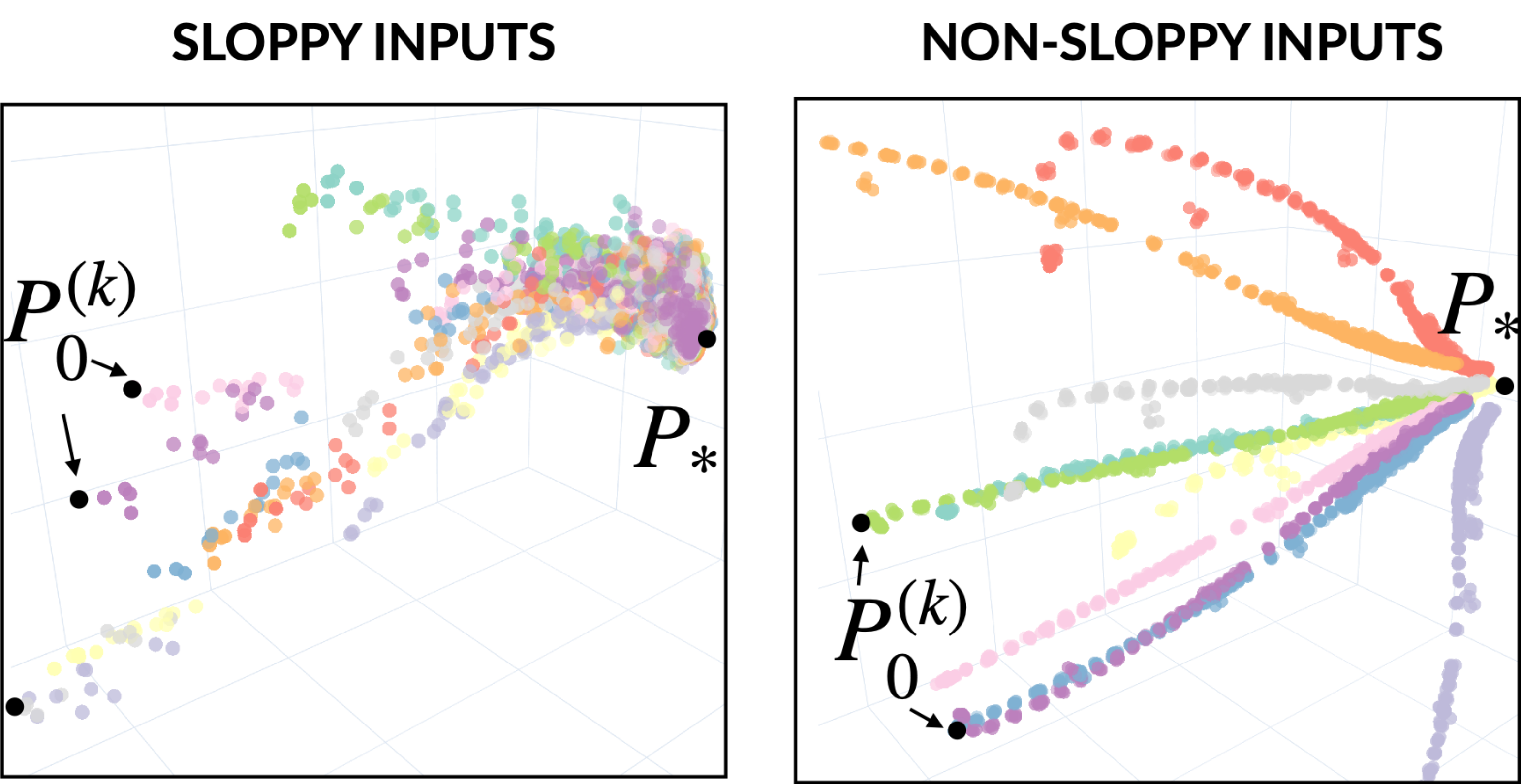}
\caption{}
\label{fig:synthetic_from_corners_panel}
\end{subfigure}
\caption{
\textbf{(a,b)}: the explained stress of an InPCA embedding of training trajectories that are initialized at different parts (``corners'') of the prediction space for non-sloppy and sloppy data respectively. For each setting we have chosen the same number of trajectories, i.e., 200 trajectories for 1 corner, 100 trajectories each from 2 corners etc. \textbf{(c)}: the top three dimensions of an InPCA embedding of models along train trajectories for sloppy and non-sloppy input data; colors indicate trajectories trained from different corners $P_0^{(k)}$. For sloppy input data, trajectories that begin at different corners quickly converge to the same manifold before heading to the truth $P_*$, but there is a larger spread in the points near the truth.}
\label{fig:synthetic_from_corners}
\end{figure}

We next investigated the effect of initialization. We sample weights of the fully-connected layers from a standard Gaussian distribution without scaling down the variance like that in the default PyTorch initialization. Due to this, the largest output probability of the network at initialization is close to 1 (as opposed to close to 0.2 for the standard initialization when there are 5 classes). Effectively, such models are near the corners of the probability simplex. We sampled 10 such corners and 50 weight initializations using the standard initialization for each corner; this gives 50 different probabilistic models (each, for two optimization algorithm: SGD and SGD with Nesterov's acceleration, and two values of weight-decay) near each of the 10 corners to begin training from. We only used a one hidden-layer fully-connected network for training from the corners. These networks were trained towards the truth $P_*$ with a fixed batch size (100) and two values of the weight decay coefficient (0 and $10^{-5}$). For both sloppy and non-sloppy data, this gives 200 trajectories from each of the 10 corners to be used for analysis. With the number of trajectories fixed to 200, we performed an InPCA embedding of models along trajectories starting from different corners, e.g., 200 trajectories from one corner, 100 trajectories each from two corners, 40 trajectories each from 5 corners, etc.

Again, as~\cref{fig:synthetic_corners_explained_stress_nonsloppy,fig:synthetic_corners_explained_stress_sloppy} show for non-sloppy and sloppy data respectively, the explained stress of an InPCA embedding of models along these trajectories in the top few dimensions is high. The explained stress captured by the top three dimensions for sloppy data is higher; this is because trajectories beginning from different corners look very similar in~\cref{fig:synthetic_from_corners_panel} for such data. For non-sloppy data, even if the explained stress is lower in the top three dimensions (the InPCA embedding in~\cref{fig:synthetic_from_corners_panel} shows a clearer separation between trajectories), the explained stress is much higher if the embedding has more dimensions. This is indicative of the difficulty in optimization for sloppy input data (one can also see a larger spread towards the end of training in~\cref{fig:synthetic_from_corners_panel} for sloppy data).

\begin{figure}[!t]
\centering
\begin{subfigure}[b]{0.4\linewidth}
\centering
\includegraphics[width=\linewidth]{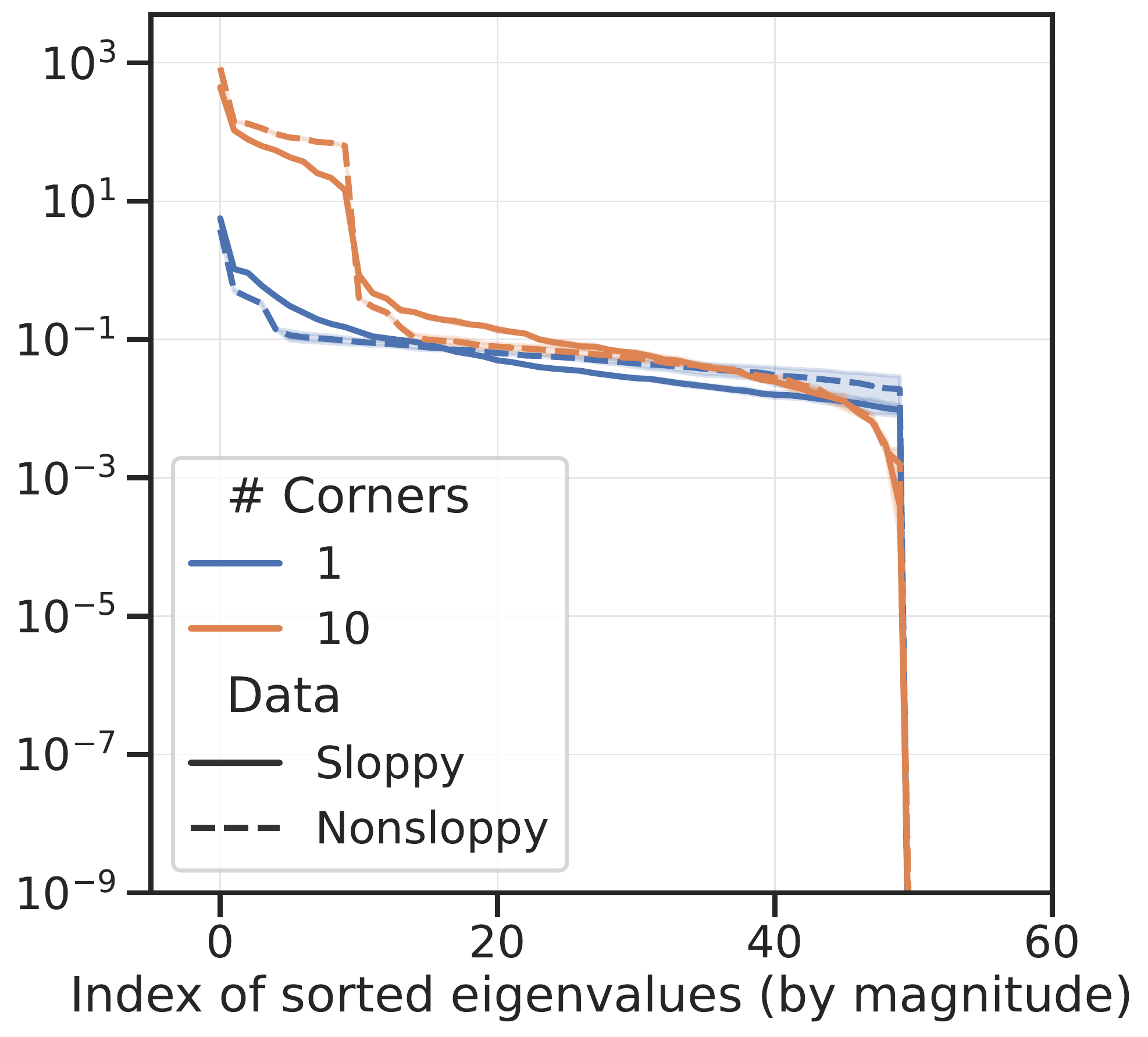}
\caption{}
\label{fig:synthetic_tube_shape_at_init}
\end{subfigure}
\hspace*{5ex}
\begin{subfigure}[b]{0.4\linewidth}
\centering
\includegraphics[width=\linewidth]{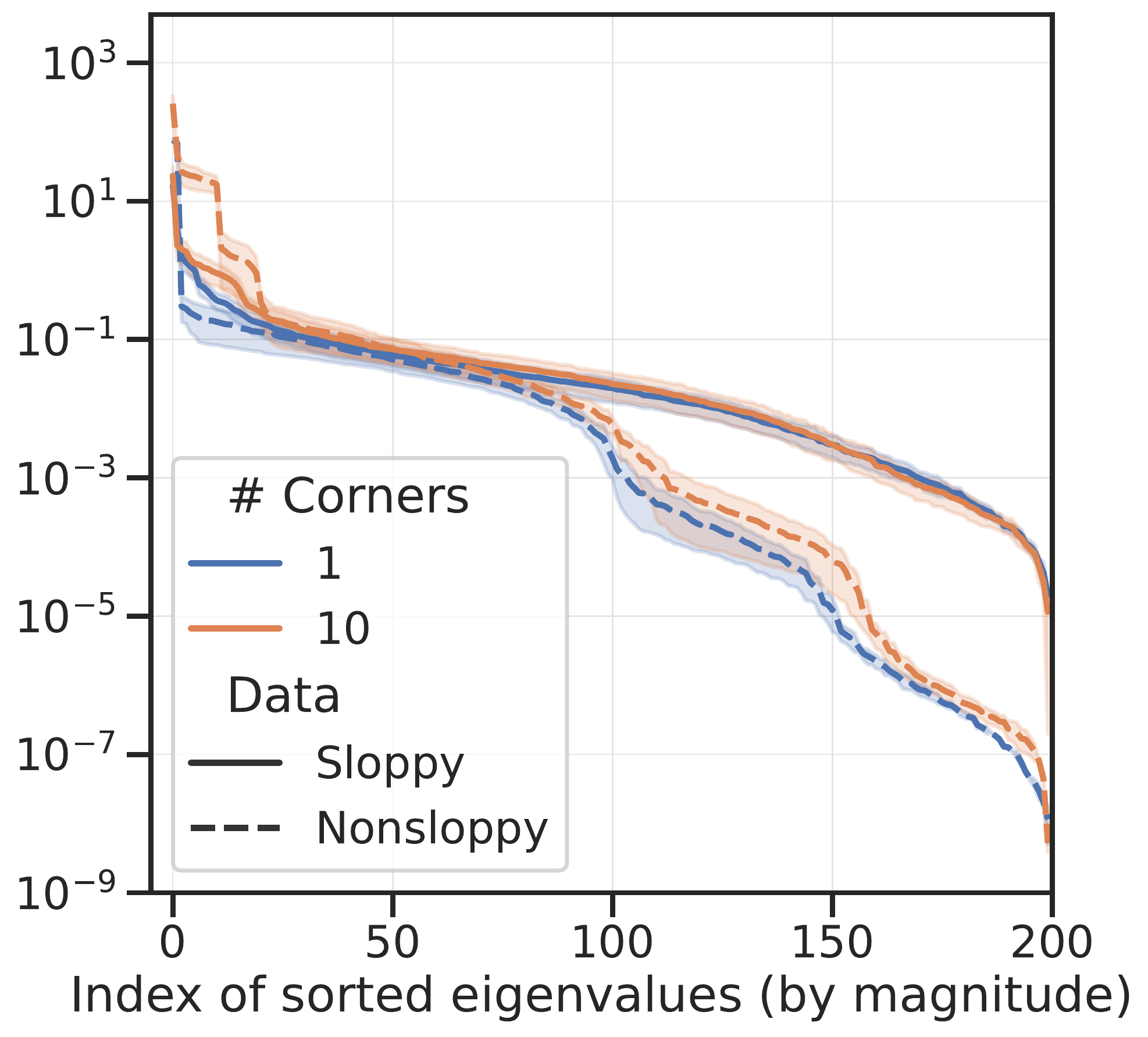}
\caption{}
\label{fig:synthetic_tube_shape_at_mid}
\end{subfigure}
\caption{
\textbf{(a,b)}: eigenvalues of the pairwise distance matrix $D$ (see~\cref{eq:w}) of InPCA of models trained from corners at the beginning of training and after 0.5\% of the training epochs, respectively. Our goal was to slice the tube of trajectories of networks with different weight initializations corresponding to the same configuration and investigate the dimensionality of the constituent models in this slice. In both cases, for both sloppy and non-sloppy input data, even if the slice is not low-dimensional the trajectories themselves in~\cref{fig:synthetic_from_corners} are effectively low-dimensional.}
\label{fig:synthetic_evals}
\end{figure}

The initial 200 probability distributions (corresponding to 50 weight initializations, 1 architecture, 2 different optimization algorithms and 2 different values of weight-decay) do not lie on a low-dimensional manifold, see~\cref{fig:synthetic_tube_shape_at_init}. In fact, the 200 probability distributions corresponding to models at an intermediate point of training (after 0.5\% of the total number of epochs) also do not lie on a low-dimensional manifold, see~\cref{fig:synthetic_tube_shape_at_mid}. So it is remarkable that in~\cref{fig:synthetic_from_corners}, the manifold formed by 200 trajectories across 4 training configurations that begin that these initializations can be embedded into a low-dimensional space faithfully (dynamics in the prediction space is clearly nonlinear). This is yet another evidence of the effectiveness of InPCA at plucking out structure in high-dimensional data.

These experiments on synthetic data suggest that, both initialization near ignorance in the prediction space and the spectral properties of the input data, could be the reason for the low-dimensionality of the train and test manifolds.

\end{appendix}